\documentclass{article}

\newif\iffancyversion
\fancyversionfalse

\PassOptionsToPackage{numbers, compress}{natbib}
\usepackage[preprint]{neurips_2023}

\usepackage[utf8]{inputenc} % allow utf-8 input
\usepackage[T1]{fontenc}    % use 8-bit T1 fonts
\usepackage{hyperref}       % hyperlinks
\usepackage{url}            % simple URL typesetting
\usepackage{booktabs}       % professional-quality tables
\usepackage{amsfonts}       % blackboard math symbols
\usepackage{amsthm}
\usepackage{amssymb}
\usepackage{mathtools}
\usepackage[inline]{enumitem}
\usepackage{nicefrac}       % compact symbols for 1/2, etc.
\usepackage{microtype}      % microtypography
\usepackage[dvipsnames]{xcolor}
\usepackage{xspace}
\usepackage{xparse}         % for \NewDocumentCommand
\usepackage{cleveref}       % \cref \Cref commands
\usepackage{stmaryrd}       % For \llbracket etc
\usepackage{graphicx}       % For graphicspath
\usepackage{subcaption}
\usepackage{bm}
\usepackage{commath} % for \dif,  \od[2]{f}{x},  \pd[2]{f}{x}
\usepackage{braket} 	% For \Set{ \ldots  | \ldots  }
\usepackage{tabularray}  % For author affiliations

\usepackage{titlesec}
\titlespacing{\paragraph}{%
  0pt}{%              left margin
  0.0\baselineskip}{% space before (vertical)
  1em}%               space after (horizontal)

\usepackage{etoolbox,siunitx}
\robustify\bfseries
\setitemize{noitemsep,topsep=2pt,parsep=2pt,partopsep=0pt}
\setenumerate{noitemsep,topsep=2pt,parsep=2pt,partopsep=0pt}

\newcommand{\nystrom}{Nystr{\"o}m}

\newcommand{\px}{P_X}
\newcommand{\py}{P_Y}
\newcommand\Pl{\hat{P}_{\lambda}}

\newcommand\dspace{\cX}

\newcommand\td{\pi}	% X data distribution
\newcommand\jd{\rho}	% Joint data distribution

\newcommand{\rkhs}{\cH}		% RKHS
\NewDocumentCommand\fmap{g}{\phi\IfNoValueF{#1}{(#1)}}
\DeclarePairedDelimiter{\iprkhs}{\langle}{\rangle_{\rkhs}}		% RKHS inner product
\DeclarePairedDelimiter{\nrkhs}{\lVert}{\rVert}		% RKHS norm
\DeclarePairedDelimiter{\noprkhs}{\lVert}{\rVert_{\text{\cB(\rkhs)}}}		% RKHS operator norm
\DeclarePairedDelimiter{\nop}{\lVert}{\rVert_{\textup{op}}}	% for concentration lemmas

\newcommand\hsH{\textup{HS}(\rkhs)}
\newcommand\HS{\textup{HS}}

\newcommand{\ltsp}[1][\td]{L^2_{#1}} 						% L₂(ρ)
\newcommand{\lt}{\ltsp}
\newcommand{\linf}{L^\infty_{\td}}
\DeclarePairedDelimiter{\nlt}{\lVert}{\rVert_{\ltsp}}		% L₂(ρ)
\DeclarePairedDelimiter{\nHlt}{\lVert}{\rVert_{\rkhs \rightarrow \ltsp}}% H→L₂(ρ)

\newcommand{\is}[1][\alpha]{[\cH]_{\td}^{#1}} 	% Interp. space

\newcommand{\pR}{\cR_{\HS}}
\newcommand{\eR}{\hat{\cR}_{\HS}}
\newcommand{\ER}{\cE_{\HS}}
\newcommand{\ERop}{\cE}

\newcommand{\kopnopi}{\cK}
\newcommand{\kop}{\cK_{\td}}			% Koopman operator L₂ → L₂
\newcommand\mftX{\Phi_X}	% Mean feature L_2 -> H (S^*)
\newcommand\amftX{\Phi_X^*} % inclusion H -> L_2  (S)
\newcommand\mftYcX{\Phi_{Y|X}}	% Mean feature L_2 -> H (S^*), conditional on X
\newcommand\amftYcX{\Phi_{Y|X}^*} % inclusion H -> L_2  (S), conditional on X
\newcommand{\emftX}{\hat{\Phi}_X}	 		% Mean feature over X (ℝ^n -> H)
\newcommand{\aemftX}{{\hat{\Phi}_X}^*}		% Adjoint / sampling operator (H -> ℝ^n)
\newcommand{\emftYcX}{\hat{\Phi}_{Y|X}}	 		% Mean feature over X (ℝ^n -> H)
\newcommand{\aemftYcX}{{\hat{\Phi}_{Y|X}}^*}		% Adjoint / sampling operator (H -> ℝ^n)
\newcommand{\nmftX}{\widetilde{\Phi}_X}		
\newcommand{\inmftX}{\nmftX^{\dagger}}
\newcommand{\anmftX}{\widetilde{\Phi}_X^*}
\newcommand{\ainmftX}{\nmftX^{*^\dagger}}

\newcommand{\nmftY}{\widetilde{\Phi}_Y}
\newcommand{\anmftY}{\widetilde{\Phi}_Y^*}

\newcommand{\Xcov}{C}  % GIAC: Changed this after main deadline
\newcommand{\rXcov}{C_{\lambda}}

\newcommand{\XYcov}{C_{XY}}
\newcommand{\YXcov}{C_{YX}}
\newcommand{\eXcov}{\hat{C}}  % VLAD: Changed this after main deadline
\newcommand{\reXcov}{\hat{C}_{\lambda}} % VLAD: Changed this after main deadline

\newcommand{\eYXcov}{\hat{C}_{YX}}
\newcommand{\rnXcov}{\tilde{C}_{\lambda}}

\newcommand{\nYXcov}{\tilde{C}_{YX}}

\DeclareDocumentCommand\cme{g}{μ_p\IfNoValueF{#1}{(#1)}}

\newcommand{\ptx}{\bm{x}}
\newcommand{\pty}{\bm{y}}

\DeclareDocumentCommand\ldm{g}{\tilde{\bm{x}}\IfNoValueF{#1}{_{#1}}} % (i-th) landmark
\DeclareDocumentCommand\ldmY{g}{\tilde{\bm{y}}\IfNoValueF{#1}{_{#1}}} % (i-th) landmark
\newcommand\ldmsX{\tilde{X}}
\newcommand\ldmsY{\tilde{Y}}

\NewDocumentCommand\kmat{O{n}}{K_{#1}}

\newcommand{\knmx}{K_{X, \ldmsX}}
\newcommand{\kmnx}{K_{\ldmsX, X}}
\newcommand{\kmmx}{K_{\ldmsX, \ldmsX}}
\newcommand{\kmmy}{K_{\ldmsY, \ldmsY}}
\newcommand{\knmy}{K_{Y, \ldmsY}}
\newcommand{\kmny}{K_{\ldmsY, Y}}
\newcommand{\kmmxy}{K_{\ldmsX, \ldmsY}}

\newcommand{\supfmap}{K}
\newcommand{\supk}{\supfmap^2}

\NewDocumentCommand\rest{O{\lambda}}{A_{#1}}	% Regularized estimator
\NewDocumentCommand\eest{O{\lambda}}{\hat{A}_{#1}}	% Empirical estimator
\NewDocumentCommand\nysest{O{\lambda}}{\hat{A}_{m,#1}^{\text{KRR}}}
\NewDocumentCommand\rrrest{O{\lambda}}{\hat{A}_{m,#1}^{\text{RRR}}}
\NewDocumentCommand\pcrest{O{\lambda}}{\hat{A}_{m}^{\text{PCR}}}
\NewDocumentCommand\fkrrest{O{\lambda}}{\hat{A}_{#1}^{\text{KRR}}}
\NewDocumentCommand\frrrest{O{\lambda}}{\hat{A}_{#1}^{\text{RRR}}}
\NewDocumentCommand\fpcrest{O{\lambda}}{\hat{A}_{}^{\text{PCR}}}

\NewDocumentCommand\krr{O{\lambda}}{A_{#1}}
\NewDocumentCommand\nkrr{O{\lambda}}{\tilde{A}_{#1}}
\NewDocumentCommand\ekrr{O{\lambda}}{\hat{A}_{#1}}
\NewDocumentCommand\rrr{O{\lambda}}{{A}_{#1}^{\text{RRR}}}
\NewDocumentCommand\pcr{O{\lambda}}{{A}_{#1}^{\text{PCR}}}
\NewDocumentCommand\npcr{O{\lambda}}{\hat{A}_{m,#1}^{\text{PCR}}}

\newcommand\tB{B}

\newcommand\nB{\tilde{B}}

\newcommand\ptB{P_{\tB}}
\newcommand\pnB{P_{\nB}}

\newcommand\ptC{P_{\rXcov}}
\newcommand\pnC{P_{\rnXcov}}

\DeclarePairedDelimiter{\nhsH}{\lVert}{\rVert_{\textup{HS}(\rkhs)}}

\newcommand\regikrr[1][\eXcov]{g_{\textup{KRR}}(#1)}
\newcommand\regipcr[1][\eXcov]{g_{\textup{PCR}}(#1)}

\newcommand\deff[1][\lambda]{d_{\textup{eff}}(#1)}

\newcommand\chalf{\theta_1}	%‖Chat^{-1/2}C^{1/2}‖
\newcommand\chalfsw{\theta_2}	% swapped signs
\newcommand\cone{\theta_3} %‖Chat^{-1}C‖

\DeclareDocumentCommand\cme{g}{\mu_p\IfNoValueF{#1}{(#1)}}

\newcommand\de{:=}
\newcommand\der{=:}
\newcommand\eg{e.g.\ }
\newcommand\ie{i.e.\ }

\newcommand\irange[1]{\llbracket #1 \rrbracket}

\newcommand{\tiid}{i.i.d.\ }

\newcommand\numberthis{\addtocounter{equation}{1}\tag{\theequation}}

\newcommand\restr[2]{{% we make the whole thing an ordinary symbol
  \left.\kern-\nulldelimiterspace % automatically resize the bar with \right
  #1 % the function
  \vphantom{\big|} % pretend it's a little taller at normal size
  \right|_{#2} % this is the delimiter
  }}

\DeclareMathOperator*{\cl}{cl}
\DeclareMathOperator*{\ran}{ran}

\DeclareMathOperator*{\argmin}{arg\,min}

\DeclareMathOperator*{\rk}{rk}
\DeclareMathOperator*{\erk}{r_{\text{eff}}}
\DeclareMathOperator*{\esssup}{ess\,sup}

\DeclareMathOperator*{\Tr}{tr}

\DeclareMathOperator*{\spa}{span}

\newcommand{\V}[1]{\symbf{#1}} % for pdflatex

\newcommand*\diff{\mathop{}\!\mathrm{d}}

\newcommand{\kron}{\otimes}

\newcommand{\bC}{\mathbb{C}}

\newcommand{\bN}{\mathbb{N}}

\newcommand{\bR}{\mathbb{R}}

\newcommand{\E}{\mathbf{E}}

\newcommand{\Prob}[1]{\mathbb{P}\brk*{#1}}

\DeclarePairedDelimiter{\prt}{(}{)}
\DeclarePairedDelimiter{\brk}{[}{]}
\DeclarePairedDelimiter{\cb}{\{}{\}}
\let\norm\relax
\DeclarePairedDelimiter{\norm}{\lVert}{\rVert}
\DeclarePairedDelimiter{\n}{\lVert}{\rVert}
\DeclarePairedDelimiter{\nhs}{\lVert}{\rVert_{\textup{HS}}}
\DeclarePairedDelimiter{\ip}{\langle}{\rangle}
\DeclarePairedDelimiter{\absv}{|}{|}

\DeclareFontFamily{U}{matha}{\hyphenchar\font45}
\DeclareFontShape{U}{matha}{m}{n}{
<-6> matha5 <6-7> matha6 <7-8> matha7
<8-9> matha8 <9-10> matha9
<10-12> matha10 <12-> matha12
}{}
\DeclareSymbolFont{matha}{U}{matha}{m}{n}

\DeclareFontFamily{U}{mathx}{\hyphenchar\font45}
\DeclareFontShape{U}{mathx}{m}{n}{
<-6> mathx5 <6-7> mathx6 <7-8> mathx7
<8-9> mathx8 <9-10> mathx9
<10-12> mathx10 <12-> mathx12
}{}
\DeclareSymbolFont{mathx}{U}{mathx}{m}{n}

\DeclareMathDelimiter{\vvvert} {0}{matha}{"7E}{mathx}{"17}%
\DeclarePairedDelimiterX{\normiii}[1]
{\vvvert}
{\vvvert}
{\ifblank{#1}{\:\cdot\:}{#1}}

\newcommand{\lcba}[1]{\left\{\begin{aligned}#1\end{aligned}\right.}

\usepackage{pgffor}
\foreach \x in {a,...,z}{
  \expandafter\xdef\csname V\x \endcsname{\noexpand\ensuremath{\noexpand\V{\x}}}
}
\foreach \x in {A,...,Z}{
  \expandafter\xdef\csname V\x \endcsname{\noexpand\ensuremath{\noexpand\V{\x}}}
  \expandafter\xdef\csname c\x \endcsname{\noexpand\ensuremath{\noexpand\mathcal{\x}}}
  \expandafter\xdef\csname f\x \endcsname{\noexpand\ensuremath{\noexpand\mathfrak{\x}}}
}

\Crefname{equation}{Eq.}{Eqs.}
\crefname{equation}{eq.}{eqs.}

\usepackage{import} 
\usepackage{etoolbox}
\usepackage[most]{tcolorbox}

\definecolor{customBlue}{RGB}{18,75,126}
\colorlet{titleCol}{customBlue}	% for lemmas, proofs etc.
\colorlet{titleThmCol}{titleCol} % for theorems
\colorlet{backCol}{titleCol!08!white}
\colorlet{backThmCol}{titleThmCol!08!white}

\tcbset{
  compactdefaultstyle/.style={
	breakable,
	theorem style=plain,
	enhanced,
	sharp corners,
	frame hidden,
	colback=backCol,
	opacityfill=0, 
	fontupper=\itshape,
	fonttitle=\bfseries\upshape,
	coltitle=black, 
	boxrule=0pt,
	before skip=1mm,
	after skip=1mm,
	right = 0mm, 
	left = -1mm, 
	top = 0mm,
	bottom = 0mm,
  },
  defaultthmstyle/.style={
	breakable,
	theorem style=plain,
	enhanced,
	sharp corners,
	frame hidden,
	colback=backCol,
	opacityfill=0, 
	fontupper=\itshape,
	fonttitle=\bfseries\upshape,
	coltitle=black, 
	boxrule=0pt,
	before skip=2mm,
	after skip=2mm,
	right = 0mm, 
	left = -1mm, 
	top = 1mm,
	bottom = 1mm,
  },
  defaultproofstyle/.style={
	breakable,
	theorem style=plain,
	terminator sign={.},  % End of title, before the content of the box (to separate because it's inlined)
	enhanced,
	sharp corners,
	frame hidden,
	colback=backCol,
	opacityfill=0, 
	fontupper=\upshape,
	fonttitle=\itshape,
	separator sign={:\ }, 
	coltitle=black, 
	boxrule=0pt,
	before skip=2mm,
	after skip=2mm,
	right = 0mm, 
	left = -1mm, 
	top = 1mm,
	bottom = 1mm,
  },
  thmstyle/.style={
	breakable,
	sharp corners,
	colbacktitle=titleThmCol,
	colback=backThmCol,
	colframe=titleThmCol, 
	fonttitle=\upshape\bfseries\hypersetup{linkcolor=white,citecolor=white},
	left = 2mm, 
	right = 2mm, 
	before skip=2mm,
	after skip=2mm,
  },
  lemmastyle/.style={
	breakable,
	enhanced,
	theorem style=plain,
	frame hidden,
	sharp corners,
	colback=backCol,
	boxrule=0pt,
	before skip=2mm,
	after skip=2mm,
	right = 2mm, 
	borderline={0.4mm}{0pt}{titleCol},
	fonttitle=\bfseries,
	coltitle=titleCol,
  },
  defstyle/.style={
	breakable,
	theorem style=plain,
	enhanced,
	sharp corners,
	frame hidden,
	colback=backCol,
	coltitle=titleCol,
	boxrule=0pt,
	borderline west={0.8mm}{0pt}{backCol},
	fonttitle=\upshape\bfseries\hypersetup{citecolor=titleCol},
	fontupper=\upshape,
	before skip=2mm,
	after skip=2mm,
	left = 3mm, 
  },
  proofstyle/.style={
	breakable,
	enhanced,
	theorem style=plain,
	frame hidden,
	sharp corners,
	colback=backCol,
	boxrule=0pt,
	before skip=2mm,
	after skip=2mm,
	right = 2mm, 
	borderline west={0.4mm}{0pt}{titleCol},
	fonttitle=\bfseries,
	coltitle=titleCol,
  }, 
  bluewestlinestyle/.style={
	breakable,
	theorem style=plain,
	enhanced,
	sharp corners,
	frame hidden,
	coltitle=customBlue,
	boxrule=0pt,
	borderline west={0.8mm}{0pt}{customBlue},
	fonttitle=\upshape\bfseries\hypersetup{citecolor=customBlue,linkcolor=customBlue},
	before skip=2mm,
	after skip=2mm,
	left = 3mm, 
	opacityfill=0, 
	bottom=0mm, 
	top=0mm, 
	toptitle=0mm,
	bottomtitle=0mm,
	right = 0mm, 
  },
  redvioletwestlinestyle/.style={
	breakable,
	theorem style=plain,
	enhanced,
	sharp corners,
	frame hidden,
	colback=black!03!white,
	coltitle=RedViolet,
	boxrule=0pt,
	borderline west={0.8mm}{0pt}{RedViolet},
	fonttitle=\upshape\bfseries\hypersetup{citecolor=RedViolet,linkcolor=RedViolet},
	before skip=2mm,
	after skip=2mm,
	left = 3mm, 
	opacityfill=0, 
	bottom=0mm, 
	top=0mm, 
	toptitle=0mm,
	bottomtitle=0mm,
	right = 0mm, 
  },
  greenwestlinestyle/.style={
	breakable,
	theorem style=plain,
	enhanced,
	sharp corners,
	frame hidden,
	coltitle=ForestGreen,
	boxrule=0pt,
	borderline west={0.8mm}{0pt}{ForestGreen},
	fonttitle=\upshape\bfseries\hypersetup{citecolor=ForestGreen,linkcolor=ForestGreen},
	before skip=2mm,
	after skip=2mm,
	left = 3mm, 
	opacityfill=0, 
	bottom=0mm, 
	top=0mm, 
	toptitle=0mm,
	bottomtitle=0mm,
	right = 0mm, 
  },
}

\tcbset{todobox/.style={
  title={Some work is needed here}, 
  colframe=red!70!black,
  colback=red!10,
  fonttitle=\bfseries, 
  coltext=red!70!black, 
  breakable,
  boxed title style={
    outer arc=0pt,
    arc=0pt,
    top=0pt,
    bottom=0pt,
    },
  }
}

\tcbset{warnbox/.style={
  colframe=black,
  colback=yellow!20,
  breakable,
  boxed title style={
    outer arc=0pt,
    arc=0pt,
    top=0pt,
    bottom=0pt,
    },
  }
}
\tcbset{infobox/.style={
  colframe=blue,
  colback=blue!05,
  breakable,
  boxed title style={
    outer arc=0pt,
    arc=0pt,
    top=0pt,
    bottom=0pt,
    },
  }
}

\makeatletter
\def\@LN@depthbox{%
  \ifdim\@tempdima = -1000pt
  \else
    \dp\@tempboxa=\@tempdima
    \nointerlineskip \kern-\@tempdima 
  \fi
  \box\@tempboxa
  } 
\makeatother

\iffancyversion

\newtcbtheorem[crefname={definition}{definitions},Crefname={Definition}{Definitions},number within=section]{tdefinition}{Definition}{bluewestlinestyle}{d}

\newtcbtheorem[crefname={theorem}{theorems},Crefname={Theorem}{Theorems},number within=section]{ttheorem}{Theorem}{bluewestlinestyle}{r}
\newtcbtheorem[crefname={lemma}{lemmas},Crefname={Lemma}{Lemmas},use counter from=ttheorem]{tlemma}{Lemma}{bluewestlinestyle}{r}
\newtcbtheorem[crefname={proposition}{propositions},Crefname={Proposition}{Propositions},use counter from=ttheorem]{tprop}{Proposition}{bluewestlinestyle}{r}
\newtcbtheorem[crefname={proof}{proofs},Crefname={Proof}{Proofs},number within=section]{tproof}{Proof}{redvioletwestlinestyle}{pf}
\newtcbtheorem[crefname={assumption}{assumptions},Crefname={Assumption}{Assumptions},use counter from=ttheorem]{tassumption}{Assumption}{redvioletwestlinestyle}{a}
\newtcbtheorem[crefname={remark}{remarks},Crefname={Remark}{Remarks},number within=section, ]{tremark}{Remark}{greenwestlinestyle}{a}

\else

\newtcbtheorem[crefname={theorem}{theorems},Crefname={Theorem}{Theorems},number within=section]{ttheorem}{Theorem}{defaultthmstyle}{r}
\newtcbtheorem[crefname={lemma}{lemmas},Crefname={Lemma}{Lemmas},use counter from=ttheorem]{tlemma}{Lemma}{defaultthmstyle}{r}
\newtcbtheorem[crefname={proposition}{propositions},Crefname={Proposition}{Propositions},use counter from=ttheorem]{tprop}{Proposition}{defaultthmstyle}{r}
\newtcbtheorem[crefname={proof}{proofs},Crefname={Proof}{Proofs},number within=section]{tproof}{Proof}{defaultproofstyle}{pf}
\newtcbtheorem[crefname={remark}{remarks},Crefname={Remark}{Remarks},number within=section, ]{tremark}{Remark}{defaultthmstyle}{a}
\newtcbtheorem[crefname={assumption}{assumptions},Crefname={Assumption}{Assumptions},use counter from=ttheorem, ]{tassumption}{Assumption}{compactdefaultstyle}{a}

\fi

\newenvironment{tproofof*}[2]{
\begin{tproof*}[title={Proof of \Cref{#1}:}]{}{#2}
}{
\null\hfill$\square$
\end{tproof*}
}

\graphicspath{{./figures/}}

\makeatletter
\patchcmd{\@setref}{\bfseries ??}{\bfseries\color{red} undefined Label}{}{}
\makeatother
\makeatletter
\patchcmd{\@@setcref}         {??}{\color{red} undefined Label}{}{}
\patchcmd{\@@setcref}         {??}{\color{red} undefined Label}{}{}
\patchcmd{\@@setcrefrange}    {??}{\color{red} undefined Label}{}{}
\patchcmd{\@@setcrefrange}    {??}{\color{red} undefined Label}{}{}
\patchcmd{\@@setcrefrange}    {??}{\color{red} undefined Label}{}{}
\patchcmd{\@@setcrefrange}    {??}{\color{red} undefined Label}{}{}
\patchcmd{\@@setcrefrange}    {??}{\color{red} undefined Label}{}{}
\patchcmd{\@@setcrefrange}    {??}{\color{red} undefined Label}{}{}
\patchcmd{\@@setnamecref}     {??}{\color{red} undefined Label}{}{}
\patchcmd{\@@setnamecref}     {??}{\color{red} undefined Label}{}{}
\patchcmd{\@@setcpageref}     {??}{\color{red} undefined Label}{}{}
\patchcmd{\@@setcpageref}     {??}{\color{red} undefined Label}{}{}
\patchcmd{\@@setcpagerefrange}{??}{\color{red} undefined Label}{}{}
\patchcmd{\@@setcpagerefrange}{??}{\color{red} undefined Label}{}{}
\patchcmd{\@@setcpagerefrange}{??}{\color{red} undefined Label}{}{}
\patchcmd{\@@setcpagerefrange}{??}{\color{red} undefined Label}{}{}
\patchcmd{\@@setcpagerefrange}{??}{\color{red} undefined Label}{}{}
\patchcmd{\@@cref}            {??}{\color{red} undefined Label}{}{}
\makeatother
\title{Estimating Koopman operators with sketching to provably learn large scale dynamical systems}

\author{
    Giacomo Meanti\textsuperscript{1, *} \\ {\tt \small giacomo.meanti@iit.it} \And 
    Antoine Chatalic\textsuperscript{2, *} \\ {\tt \small antoine.chatalic@dibris.unige.it} \And 
    Vladimir R. Kostic\textsuperscript{1,3} \\ {\tt \small vladimir.kostic@iit.it} \And
    Pietro Novelli\textsuperscript{1} \\ {\tt \small pietro.novelli@iit.it} \And 
    Massimiliano Pontil\textsuperscript{1,5} \\ {\tt \small massimiliano.pontil@iit.it} \And 
    Lorenzo Rosasco\textsuperscript{1,2,4} \\ {\tt \small lrosasco@mit.edu} 
}

\begin{document}

\maketitle
\renewcommand*{\thefootnote}{\fnsymbol{footnote}}
\footnotetext[1]{
Equal contribution \\
\begin{tblr}{hspan=even, colspec={XX}}%{*{2}{>{$\relax}X<{$}}*{4}{X[si]}},}
    \textsuperscript{1}Istituto Italiano di Tecnologia & \textsuperscript{2}MaLGa -- DIBRIS, Università di Genova
\end{tblr}\\
\begin{tblr}{hspan=even, colspec={XXX}}%{*{2}{>{$\relax}X<{$}}*{4}{X[si]}},}
    \textsuperscript{3}University of Novi Sad &
    \textsuperscript{4}MIT, CBMM &
    \textsuperscript{5}University College London
\end{tblr}%
}
\renewcommand*{\thefootnote}{\arabic{footnote}}
\setcounter{footnote}{0}
% \footnotetext[1]{Istituto Italiano di Tecnologia}
% \footnotetext[2]{MaLGa -- DIBRIS, Università degli Studi di Genova}
% \footnotetext[3]{University of Novi Sad}
% \footnotetext[4]{MIT, CBMM}
% \footnotetext[5]{University College London}
\begin{abstract}
The theory of Koopman operators allows to deploy non-parametric machine learning algorithms to predict and analyze complex dynamical systems.
Estimators such as principal component regression (PCR) or reduced rank regression (RRR) in kernel spaces can be shown to provably learn Koopman operators from finite empirical observations of the system's time evolution. 
Scaling these approaches to very long trajectories is a challenge and requires introducing suitable approximations to make computations feasible. 
In this paper, we boost the efficiency of 
different kernel-based Koopman operator estimators using random projections (sketching).
We derive, implement and test the new ``sketched'' estimators with extensive experiments on synthetic and large-scale molecular dynamics datasets. 
Further, we establish non asymptotic error bounds giving a sharp characterization of the trade-offs between statistical learning rates and computational efficiency.
Our empirical and theoretical analysis shows that the proposed estimators provide a sound and efficient way to learn large scale dynamical systems.
In particular our experiments indicate that the proposed estimators retain the same accuracy of PCR or RRR, while being much faster. Code is available at \url{https://github.com/Giodiro/NystromKoopman}.
\end{abstract}

%%%%%%%%%%%%%%%%%%%%%%%%%%%%%%%%%%%%%%%%%%%%%%%%%%%%%%%%%%%%%%%%%%%%%%%
%                         Intro & estimators                          %
%%%%%%%%%%%%%%%%%%%%%%%%%%%%%%%%%%%%%%%%%%%%%%%%%%%%%%%%%%%%%%%%%%%%%%%

\section{Introduction}\label{s:intro}

In the physical world, temporally varying phenomena are everywhere, from biological processes in the cell to fluid dynamics to electrical fields. Correspondingly, they generate large amounts of data both through experiments and simulations.
This data is often analyzed in the framework of dynamical systems, where the state of a system $\ptx$ is observed at a certain time $t$, and the dynamics is described by a function $f$ which captures its evolution in time
\begin{equation*}
    \ptx_{t + 1} = f(\ptx_{t}).
\end{equation*}
The function $f$ must capture the whole dynamics, and as such it may be non-linear and even stochastic for instance when modeling stochastic differential equations, or simply noisy processes. 
Applications of this general formulation arise in fields ranging from robotics, atomistic simulations, epidemiology, and many more.
Along with a recent increase in the availability of simulated data, data-driven techniques for learning the dynamics underlying physical systems have become commonplace.
The typical approach of such techniques is to acquire a dataset of training pairs $(\ptx_t, \pty_t=\ptx_{t+1})$ sampled in time, and use them to learn a model for $f$ which minimizes a forecasting error.
Since dynamical systems stem from real physical processes, forecasting is not the only goal and the ability to interpret the dynamics is paramount.
One particularly important dimension for interpretation is the separation of dynamics into multiple temporal scales: fast fluctuations can \eg be due to thermodynamical noise or electrical components in the system, while slow dynamics describe important conformational changes in molecules or mechanical effects.

Koopman operator theory~\cite{koopman31,koopman32} provides an elegant framework in which the potentially non-linear 
%function describing time evolution $f$, can be linearized 
%via the choice of a (typically infinite-dimensional) transformation of the state $\psi$. 
%Formally, the Koopman operator $\kopnopi$ is defined as
%by an unknown and possibly infinite-dimensional transformation of the state $\psi$. Formally, the Koopman operator $\kopnopi$ is defined as
dynamics of the system
can be studied via the Koopman operator
\begin{equation}\label{e:koopman}
    (\kopnopi \psi)(\ptx) = \E\brk*{\psi(f(\ptx))}, 
\end{equation}
which has the main advantage of being linear but is defined on a typically infinite-dimensional set of observable functions.  
The expectation in \eqref{e:koopman} is taken with respect to the potential stochasticity of $f$. 
%Note that this operator is linear, hence for example
Thanks to its linearity, the operator $\kopnopi$ 
can \eg be applied twice to get two-steps-ahead forecasts, and one can compute its spectrum (beware however that $\kopnopi$ is not self-adjoint, unless the dynamical process is time-reversible). 
Accurately approximating the Koopman operator and its spectral properties is of high interest for the practical analysis of dynamical systems. However doing so efficiently for long temporal trajectories remains challenging.
In this paper we are interested in designing estimators which are both theoretically accurate and computationally efficient.

\paragraph{Related works}
Learning the spectral properties of the Koopman operator directly from data has been considered for at least 3 decades~\cite{mezic_phd}, resulting in a large body of previous work. Among the different approaches proposed over time (see~\citet{mezic_koop_21} for a recent review) it is most common to search for finite dimensional approximations to the operator.
DMD~\cite{dmdSchmid2010,dmdTu2014}, tICA~\cite{ticaMolgedey1994,ticaNoe2013} and many subsequent extensions~\cite{kutz_book_16} for example can be seen as minimizers of the forecasting error when $\psi$ is restricted to be a linear function of the states~\cite{rowley_09}. eDMD~\cite{edmd, edmdKlus2016} and VAC~\cite{VACNoe2013,VACNuske2014} instead allow for a (potentially learnable, as in recent deep learning algorithms~\cite{li_edmdlearned_17, lusch_deepkoop_18, enoch_deep_19, naoya_deep_17}) dictionary of non-linear functions $\psi$. 
KernelDMD~\cite{kerneldmd,kerneleidec_klus_20} and kernel tICA~\cite{kerneltica_pande_15} are further generalizations which again approximate the Koopman operator but using an infinite dimensional space of features $\psi$, encoded by the feature map of a reproducing kernel. 
While often slow from a computational point of view, kernel methods are highly expressive and can be analyzed theoretically, to prove convergence and derive learning rates of the resulting estimators~\cite{kostic2022learning}. 
Approximate kernel methods which are much faster to run have been recently used for Koopman operator learning by~\citet{baddooSparseKernel2022} where an iterative procedure is used to identify the best approximation to the full kernel, but no formal learning rates are demonstrated, and by~\citet{ahmad2023SketchSketchOut} who derive learning rates in Hilbert-Schmidt norm (while we consider operator norm) for the Nyström KRR estimator (one of the three considered in this paper).
\paragraph{Contributions}
In this paper we adopt the kernel learning approach. Starting from the problem of approximating the Koopman operator in a reproducing kernel Hilbert space, we derive three different estimators based on different inductive biases: 
%(regularization and optimization constraints).
kernel ridge regression (KRR) which comes from Tikhonov regularization, principal component regression (PCR) which is equivalent to dynamic mode decompositin (DMD) and its extensions, and reduced rank regression (RRR) which comes from a constraint on the maximum rank of the estimator~\cite{rrrIzenman1975}.
We show how to overcome the computational scalability problems inherent in full kernel methods using an approximation based on random projections which is known as the \nystrom{} method~\cite{smola_sparse_2000,williams_using_2001}. The approximate learning algorithms scale very easily to the largest datasets, with a computational complexity which goes from $O(n^3)$ for the exact algorithm to $O(n^2)$ for the approximate one. We can further show that the \nystrom{} KRR, PCR and RRR estimators 
%possess the same theoretical guarantees as their exact, slow counterparts -- which are known to be optimal. 
have the same convergence rates as theirs exact, slow counterparts -- which are known to be optimal under our assumptions. 
We provide learning bounds in operator norm, 
which are known to translate to bounds for dynamic mode decomposition and are thus of paramount importance for applications.
%By proving learning bounds in operator norm, we
%are able to leverage existing bounds~\cite{kostic2023KoopmanOperatorLearning} on the approximation of the Koopman operator's spectrum \todo{keep it or not? MP: Mention in the conclusion or in passing in the paper somewhere, but I'd not do it now because 1: the other paper  is unpublished, 2: we have enough material here, 3: we do one more paper!}, which is particularly important for downstream analyses.
Finally, we thoroughly validate the approximate PCR and RRR estimators on synthetic dynamical systems, comparing efficiency and accuracy against their exact counterparts~\cite{kostic2022learning}, as well as recently proposed fast Koopman estimator streaming KAF~\cite{giannakis2021LearningForecastDynamical}. To showcase a realistic scenario, we train on a molecular dynamics simulation of the fast-folding Trp-cage protein~\cite{fastfolding11}. %, and show that the leading Koopman eigenfunction accurately distinguishes between folded and unfolded states. 

\paragraph{Structure of the paper}
We introduce the setting in \Cref{s:setting_notations}, and define our three estimators in \Cref{s:estimators}. In \Cref{s:theory} we provide bounds on the excess risk of our estimators, and extensive experiments on synthetic as well as large-scale molecular dynamics datasets in \Cref{s:experiments}.

\section{Background and related work}\label{s:background}
\paragraph{Notation}\label{s:setting_notations}
We consider a measurable space $(\dspace,\cB)$ where $\dspace$ corresponds to the state space, and 
denote $\ltsp[\td]\de L^2(\dspace,\cB,\td)$ the $L^2$ space of functions on $\dspace$ w.r.t. to a probability measure $\td$,
and $\linf$ the space of measurable functions bounded almost everywhere. 
We denote $\hsH$ the space of Hilbert-Schmidt operators on a space $\rkhs$.

\paragraph{Setting}
The setting we will consider is that of Markovian, time-homogeneous stochastic process $\{X_t\}_{t\in\bN}$ on $\cX$. 
By definition of a Markov process, $X_t$ only depends on $X_{t-1}$ and not on any previous states.
Time-homogeneity ensures that the transition probability $\Prob{X_{t+1}\in B \vert X_t = \ptx}$ for any measurable set $B$ does not depend on $t$, and can be denoted with $p(\ptx, B)$.
This implies in particular that the distribution of $(X_t,X_{t+1})$ does not depend on $t$, and we denote it $\jd$ in the following.
We further assume the existence of the \emph{invariant} density $\pi$ which satisfies $\pi(B) = \int_{\cX} \pi(\ptx) p(\ptx, B) \diff \ptx$. 
This classical assumption allows one to study large class of stochastic dynamical systems, but also deterministic systems on the attractor, see e.g.~\cite{Prato1996}.
The Koopman operator $\kop: L^2_\pi(\cX)\to L^2_\pi(\cX)$ is a bounded linear operator, defined by
\begin{equation}\label{e:koopman2}
    (\kop g)(\ptx) = \int_\cX p(\ptx, \pty) g(\pty) \diff \pty = \E\brk*{g(X_{t+1}) \vert X_t = \ptx}, \quad g\in L^2_\pi(\cX), \ptx\in\cX.
\end{equation}
We are in particular interested in the eigenpairs $(\lambda _i,\varphi_i)\in \bC \times \ltsp$, that satisfy
\begin{equation}\label{eq:spectral_eq}
    \kop\varphi_i = \lambda_i \varphi_i.
\end{equation}
%The linearity of $\kop$ ensures the existence of its eigenvalues $\lambda_i\in\bC$ and eigenfunctions $\varphi_i: \cX\to\bC$ which satisfy
%\begin{equation}
    %\kop\varphi_i = \lambda_i \varphi_i.
%\end{equation}
Through this decomposition it is possible to interpret the system by separating fast and slow processes, or projecting the states onto fewer dimensions~\cite{dellnitz99,froyland14,brunton2021ModernKoopmanTheory}.
In particular, the Koopman mode decomposition (KMD) allows to propagate the system state in time. Given an observable $g:\cX\to\bR^d$ such that $g\in\spa\cb{\varphi_i \vert i \in \bN}$, the modes allow to reconstruct $g(\ptx)$ with a Koopman eigenfunction basis. The modes $\bm{\eta}_i^g\in\bC^d$ are the coefficients of this basis expansion:
\begin{equation}\label{eq:kmd}
    (\kop g)(\ptx) = \E\brk*{g(X_t) \vert X_0 = \ptx} = \sum_{i} \lambda_i \varphi_i(\ptx) \bm{\eta}_i^g.
\end{equation}
This decomposition describes the system's dynamics in terms of a stationary component (the Koopman modes), a temporal component (the eigenvalues $\lambda_i$) and a spatial component (eigenfunctions $\varphi_i$). 

\paragraph{Kernel-based learning}
In this paper we approximate $\kop$ with kernel-based algorithms, using operators in reproducing kernel Hilbert spaces (RKHS) $\rkhs$ associated with kernel $k: \cX\times\cX\to\bR$ and feature map $\fmap: \cX\to\rkhs$.
We wish to find an operator $A: \rkhs\to\rkhs$ which minimizes the risk
\begin{equation}
    \pR(A) = \E_{\jd} \brk*{\ell(A,(\ptx, \pty)) }
	\quad\text{ where }\quad
	\ell(A,(\ptx,\pty)) \de \nrkhs{\fmap(\pty)-A\fmap(\ptx)}^2.  
    %\hat{\cR}_\HS(A) = \frac{1}{n}\nhs{\emftYcX - A \emftX}^2.
	\label{e:population_risk}
\end{equation}
The operator $A^*$ should thus be understood as an estimator of the Koopman operator $\kop$ in $\rkhs$ as will be clarified in \eqref{e:population_risk_opnorm}.
In practice $\td$ and $\jd$ are unknown, and one typically has access to a dataset $\{(\ptx_i, \pty_{i})\}_{i = 1}^n$ sampled from $\jd$, where each pair $(\ptx_i, \pty_i = f(\ptx_i))$ may equivalently come from a single long trajectory or multiple shorter ones concatenated together. We thus use the empirical risk
\begin{equation}
    \eR(A) = \frac{1}{n} \sum_{i=1}^n \ell(A,(\ptx_i, \pty_i))
	\label{e:emprsk}
\end{equation}
as a proxy for \eqref{e:population_risk}. In practice, minimizing \cref{e:emprsk} may require finding the solution to a very badly conditioned linear system. To avoid this potential pitfall, different regularization methods (such as Tikhonov or truncated SVD) can be applied on top of the empirical risk.

\begin{tremark}{Connections to other learning problems}{}
    The problem of minimizing \cref{e:emprsk,e:population_risk} has strong connections to learning conditional mean embeddings~\cite{lecme09,muandet2017KernelMeanEmbedding,li2022OptimalRatesRegularized} where the predictors and targets are embedded in different RKHSs, and to structured prediction~\cite{ciliberto2016ConsistentRegularizationApproacha,ciliberto2022GeneralFrameworkConsistent} which is an even more general framework. On the other hand, the most substantial difference from the usual kernel regression setting~\cite{caponnetto07} is the embedding of both targets and predictors into a RKHS, instead of just targets.
    % {\color{blue} Remark that In general, the RKHS in which we embed targets may be different form the one in which we embed predictors, which is the case known as conditional mean embedding (CME) learning~\cite{lecme09,li2022OptimalRatesRegularized} important in different ML problems such as structure prediction~\cite{ciliberto2022GeneralFrameworkConsistent}.\todo{add citations CME and structure prediction Ciliberto \& Loz}. In this paper we address the setting of dynamical systems, hence you use the same RKHS for both embedding, but all the results can be easily applied to general CME context.}
\end{tremark}

We denote the input and cross covariance
$\Xcov=\E_{\td}[\fmap(\ptx)\kron \fmap(\ptx)]$ and
$\YXcov=\E_{\jd}[\fmap{\pty}\kron\fmap{\ptx}]$,
and their empirical counterparts as 
$\eXcov=\tfrac{1}{n}\sum_{i=1}^n [\fmap(\ptx_i)\kron \fmap(\ptx_i)]$ and
$\eYXcov=\tfrac{1}{n} \sum_{i=1}^n \fmap{\pty_i}\kron\fmap{\ptx_i}]$.
We also use the abbreviation $\rXcov\de\Xcov+\lambda I$.
%The solution to the minimization problem~\eqref{e:emprsk} can be obtained for example by using principal component regression as regularizer and the kernel trick, to get the following~\cite{kerneldmd, kostic2022learning}:
%\begin{equation}\label{e:emp_pcr}
    %\hat{A} = \emftYcX \irange{\emftX^* \emftX}_r^\dagger \aemftX
%\end{equation}
%\todo{Do we want to keep that? Maybe just add a ref. to estimators in the next section.}
%where $K = \emftX^* \emftX \in \bR^{n\times n}$ is the kernel matrix (such that $K_{i,j} = k(\bm{x}_i, \bm{x}_j)$), $\irange{\cdot}_r$ indicates the truncation to the first $r$ principal components, and $^\dagger$ the Moore-Penrose pseudoinverse. 
%
%The first full estimator can be obtained by adding Tikhonov regularization~\cite{caponnetto07} to the objective of \cref{e:emprsk}.
Minimizing the empirical risk \eqref{e:emprsk} with Tikhonov regularization~\cite{caponnetto07} yields the following KRR estimator
\begin{equation}
    \eest
	%= \argmin_{A: \rkhs\to\rkhs} \frac{1}{n} \nhs{\emftYcX - A \emftX}^2 + \lambda\nhs{A}^2 
	= \argmin_{A\in \hsH} \eR(A) + \lambda \nhs{A}^2
	= \eYXcov (\eXcov + \lambda I)^{-1}.
	\label{e:krr}
\end{equation}

\Cref{e:krr} can be computed by transforming its expression with the kernel trick~\cite{hofmann_kernels_08}, to arrive at a form where one must invert the kernel matrix -- a $n\times n$ matrix whose $i,j$-th entry is $k(\ptx_i, \ptx_j)$. This operation requires $O(n^3)$ time and $O(n^2)$ memory, severely limiting the scalability of KRR to $n \lesssim \num{100 000}$ points.
Improving the scalability of kernel methods is a well-researched topic, with the most important solutions being random features~\cite{rahimi_random_2008,rahimi09,yang12,gittens16} and random projections~\cite{smola_sparse_2000,williams_using_2001,gittens16}. In this paper we use the latter approach, whereby the kernel matrix is assumed to be approximately low-rank and is \emph{sketched} to a lower dimensionality. 
In particular we will use the \nystrom{} method to approximate the kernel matrix projecting it onto a small set of inducing points, chosen among the training set.
The sketched estimators are much more efficient than the exact ones, increasingly so as the training trajectories become longer. 
For example, the state of the art complexity for solving (non vector valued) approximate kernel ridge regression is $O(n\sqrt{n})$ time instead of $O(n^3)$~\cite{falkonlibrary2020,belkinlarge23}. Furthermore, when enough inducing points are used (typically on the order of $\sqrt{n}$), the learning rates of the exact and approximate estimators are the same, and optimal~\cite{bachnystrom_13, rudi2016LessMoreNystr}. Hence it is possible -- and in this paper we show it for learning the Koopman operator -- to obtain large efficiency gains, without losing anything in terms of theoretical guarantees of convergence.

\section{Nyström estimators for Koopman operator regression}\label{s:estimators}

%In this section, we introduce three approximate estimators based on KRR, PCR and RRR, combined with the Nyström approximation: a random projection onto the low-dimensional subspace of $\rkhs$, spanned by the feature-embedding of a subset of the data.
In this section, we introduce three efficient approximations of the KRR, PCR and RRR estimators of the Koopman operator.
Our estimators rely on the Nyström approximation, \ie on random projections onto low-dimensional subspaces of $\rkhs$ spanned by the feature-embeddings of subsets of the data.
We thus consider two sets of $m\ll n$ inducing points $\cb{\ldm{j}}_{j=1}^{m}\subset \{\ptx_t \}_{t=1}^n$ and $\cb{\ldmY{j}}_{j=1}^{m}\subset \{\pty_t \}_{t=1}^n$ sampled respectively from the input and output data.
 %are known as Nyström centers or inducing points, and 
The choice of these inducing points (also sometimes called Nyström centers) is important to obtain a good approximation.
Common choices include uniform sampling, leverage score sampling~\cite{drineas_levscore12,rudi_levscore18}, and iterative procedures such as the one used in~\cite{baddooSparseKernel2022} to identify the most relevant centers.
In this paper we focus on uniform sampling for simplicity, but we stress that our theoretical results in \Cref{s:theory} can easily be extended to leverage scores sampling by means of \cite[Lemma 7]{rudi2016LessMoreNystr}.
To formalize the \nystrom{} estimators, we define operators $\nmftX,\nmftY:\bR^m\to\rkhs$ as
$\nmftX w = \sum_{j=1}^m w_j \fmap{\ldm{j}}$ and $\nmftY w = \sum_{j=1}^m w_j \fmap{\ldmY{j}}$,
and denote $\px$ and $\py$ the orthogonal projections onto $\spa \nmftX$ and $\spa \nmftY$ respectively.

In the following paragraphs we apply the projection operators to three estimators corresponding to different choices of regularization. 
For each of them a specific proposition (proven in \Cref{s:estimators_app}) states an efficient way of computing it based on the kernel trick. 
For this purpose we introduce the kernel matrices $\kmnx, \kmny \in \bR^{m \times n}$ between training set and inducing points with entries 
$(\kmnx)_{ji} = k(\ldm{j},\ptx_i)$,
$(\kmny)_{ji} = k(\ldmY{j},\pty_i)$, 
and the kernel matrices of the inducing points $\kmmx, \kmmy \in \bR^{m \times m}$  with entries 
$(\kmnx)_{jk} = k(\ldm{j},\ldm{k})$ and
$(\kmnx)_{jk} = k(\ldmY{j},\ldmY{k})$.

\paragraph{Kernel Ridge Regression (KRR)}
The cost of computing $\eest$ defined in \Cref{e:krr} is $O(n^3)$~\cite{kostic2022learning} which is prohibitive for datasets containing long trajectories. 
However, applying the projection operators to each side of the empirical covariance operators, we obtain an estimator which additionally depends on the $m$ inducing points:
\begin{equation}\label{e:nysest}
     \nysest \de \py \eYXcov \px (\px \eXcov \px + \lambda I)^{-1} : \rkhs\to\rkhs.
\end{equation}
If $\rkhs$ is infinite dimensional, \Cref{e:nysest} cannot be computed directly. 
\Cref{r:nkrrest} (proven in \Cref{s:estimators_app}) provides a computable version of the estimator.
\begin{tprop}{\nystrom{} KRR}{nkrrest}
	The \nystrom{} KRR estimator  \eqref{e:nysest} can be expressed as
    \begin{equation}\label{e:nys_krr_k}
        \nysest = \nmftY \kmmy^\dagger \kmny\knmx (\kmnx\knmx + n\lambda\kmmx)^\dagger\anmftX.
    \end{equation}
    The computational bottlenecks are the inversion of an $m\times m$ matrix and a large matrix multiplication, which overall need $O(2m^3 + 2m^2 n)$ operations. In particular, in \Cref{s:bounds} we will show that $m \asymp \sqrt{n}$ is sufficient to guarantee optimal rates even with minimal assumptions, leading to a final cost of $O(n^2)$. Note that a similar estimator was derived in~\cite{ahmad2023SketchSketchOut}.
\end{tprop}
Please note that the $O(n^2)$ cost is for a straightforward implementation, and can indeed be reduced via iterative linear solvers (possibly preconditioned, to further reduce the practical running time), and randomized linear algebra techniques. In particular, we could leverage results from \citet{rudi_falkon17} to reduce the computational cost to $O(n\sqrt{n})$.

\paragraph{Principal Component Regression (PCR)}
Typical settings in which Koopman operator theory is used focus on the decomposition of a dynamical system into a small set of components, obtained from the eigendecomposition of the operator itself. For this reason, a good prior on the Koopman estimator is for it to be low rank. 
The kernel PCR estimator $\hat{A}^\mathrm{PCR} = \eYXcov \irange{\eXcov}_r^\dagger$ formalizes this concept~\cite{kostic2022learning, kerneldmd}, where here $\irange{\cdot}_r$ denotes the truncation to the first $r$ components of the spectrum.
%Kernel PCR formalizes this concept, and the full estimator is given by~\cite{kostic2022learning, kerneldmd} $\hat{A}^\mathrm{PCR} = \eYXcov \irange{\eXcov}_r^\dagger$ where $\irange{\cdot}_r$ indicates truncating to the first $r$ components of the spectrum.
Again this is expensive to compute when $n$ is large, but the estimator can be sketched as follows:
\begin{align}
	\pcrest &= P_Y \eYXcov \irange{P_X \eXcov P_X}_r^\dagger.
	\label{e:nys_pcr}
\end{align}
The next proposition provides an efficiently implementable version of this estimator.
\begin{tprop}{\nystrom{} PCR}{npcrest}
    The sketched PCR estimator \eqref{e:nys_pcr} satisfies
    \begin{equation}\label{e:nys_pcr_k}
        \pcrest = \nmftY \kmmy^\dagger \kmny \knmx \irange{\kmmx^\dagger \kmnx \knmx}_r \anmftX
    \end{equation}
    requiring $O(2m^3 + 2m^2n)$ operations, \ie optimal rates can again be obtained at a cost of at most $O(n^2)$ operations.
	%when $m\asymp \sqrt{m}$ to get optimal rates.
\end{tprop}
Note that with $m = n$, $\pcrest$ is equivalent to the kernel DMD estimator~\cite{kerneldmd}, also known as kernel analog forecasting (KAF)~\cite{kaf}. The sketched estimator of \Cref{r:npcrest} was also recently derived in~\cite{baddooSparseKernel2022}, albeit without providing theoretical guarantees. 
 
\paragraph{Reduced Rank Regression (RRR)}
%Instead of rank $r$ truncation as a means of regularization, as is done in PCR, we can add a constraint on the optimization problem to find the optimal rank $r$ Koopman estimator. For the problem to be well-conditioned, Tikhonov regularization must also be employed. The resulting optimization problem, formalized in \Cref{e:emp_rrr}, falls under the reduced rank regression framework~\cite{rrrIzenman1975,kostic2022learning}.
Another way to promote low-rank estimators is to add an explicit rank constraint when minimizing the empirical risk. 
Combining such a constraint with Tikhonov regularization
corresponds to the reduced rank regression~\cite{rrrIzenman1975,kostic2022learning} estimator:
\begin{equation}\label{e:emp_rrr}
    A_\lambda^{\mathrm{RRR}} = \argmin_{A\in \HS:\rk(A)\leq r} \eR(A) + \lambda \nhs{A}^2.
\end{equation}
Minimizing \Cref{e:emp_rrr} requires solving a $n\times n$ generalized eigenvalue problem. The following proposition introduces the sketched version of this estimator, along with a procedure to compute it which instead requires the solution of a $m\times m$ eigenvalue problem. For $m \asymp \sqrt{n}$, which is enough to guarantee optimal learning rates with minimal assumptions (see \Cref{s:bounds}), this represents a reduction from $O(n^3)$ to $O(n\sqrt{n})$ time.

\begin{tprop}{\nystrom{} RRR}{nrrrest}
    The \nystrom{} RRR estimator can be written as
    \begin{equation}\label{e:RRR_practical}
        \rrrest = \irange{ \py\eYXcov\px (\px\eXcov\px + \lambda I)^{-1/2}}_r (\px\eXcov\px + \lambda I)^{-1/2} .
    \end{equation}
    To compute it, solve the $m\times m$ eigenvalue problem
    \begin{equation*}
        (\kmnx\knmx + n\lambda\kmmx)^\dagger \kmnx\knmy\kmmy^\dagger\kmny\knmx w_i = \sigma_i^2 w_i
    \end{equation*}
    for the first $r$ eigenvectors $W_r = [w_1, \dots, w_r]$, appropriately normalized. %normalized such that
    %\(
    %    W_r^* \kmnx\knmy\kmmy^\dagger\kmny\knmx W_r = I.
    %\)
    Then denoting $D_r\de\kmmy^\dagger \kmny\knmx W_r$ and $E_r\de(\kmnx\knmx + n\lambda\kmmx)^\dagger\kmnx\knmy D_r$ it holds
    \begin{equation}
        \rrrest = \nmftY D_r E_r^* \anmftX.
    \end{equation}
\end{tprop}

%%%%%%%%%%%%%%%%%%%%%%%%%%%%%%%%%%%%%%%%%%%%%%%%%%%%%%%%%%%%%%%%%%%%%%%
%                               Bounds                                %
%%%%%%%%%%%%%%%%%%%%%%%%%%%%%%%%%%%%%%%%%%%%%%%%%%%%%%%%%%%%%%%%%%%%%%%

\color{black}
\section{Learning bounds in operator norm for the sketched estimators}\label{s:bounds}\label{s:theory}

In this section, we state the main theoretical results showing that optimal rates for operator learning with KRR, PCR and RRR can be reached with Nyström estimators. 
\paragraph{Assumptions}
We first make two assumptions on the space $\rkhs$ used for the approximation, via its reproducing kernel~$k$.

\begin{tassumption}{Bounded kernel}{bounded_fmap}
There exists $\supfmap<\infty$ such that $\esssup_{\ptx\sim \td}\nrkhs{\fmap(\ptx)}\leq \supfmap$. 
\end{tassumption}
\Cref{a:bounded_fmap} ensures that $\rkhs$ is compactly embedded in $\ltsp$ \cite[Lemma 2.3]{steinwart2012MercerTheoremGeneral}, and we denote $\amftX: \rkhs \rightarrow \ltsp$ the embedding operator which maps any function in $\rkhs$ to its equivalence class $\td$-almost everywhere in~$\ltsp$.
\begin{tassumption}{Universal kernel}{universal_kernel}
The kernel $k$ is universal, i.e. $\cl(\ran(\amftX))=\lt$. 
\end{tassumption}
We refer the reader to \cite[Definition 4.52]{steinwart2008SupportVectorMachines} for a definition of a universal kernel.
%
%We work in the well-specified setting, as covered by the following assumption.
%\begin{tassumption}{Well-specified setting}{regularity}
	%We assume that there exists $A_*\in \hsH$ such that $\kop \amftX = \amftX A_*$.
%\end{tassumption}
%
The third assumption on the RKHS is related to the embedding property from \citet{fischer2020SobolevNormLearning}, connected to the embedding of interpolation spaces. For a detailed discussion see~\Cref{s:interpolating_spaces}.
\begin{tassumption}{Embedding property}{embedding_property}
	There exists $\tau \in ]0,1]$ and $c_\tau>0$ such that $\esssup_{\ptx\sim \td}\nrkhs{\rXcov^{-1/2}\fmap{\ptx}}^2\leq c_{\tau}\lambda^{-\tau}$.
\end{tassumption}
%This assumption is related to the properties of the embedding operator of some interpolating spaces between $\rkhs$ and $\ltsp$ into $\linf$~\cite{fischer2020SobolevNormLearning}, however we postpone technical details to \Cref{s:interpolating_spaces} and work with this weaker property which is sufficient to derive our results.

Next, we make an assumption on the decay of the spectrum of the covariance operator that is of paramount importance for derivation of optimal learning bounds.
In the following, $\lambda_i(A)$ and $\sigma_i(A)$ always denote the eigenvalues and singular values of an operator $A$ (in decreasing order).
\begin{tassumption}{Spectral decay}{spectral_decay}
	There exists $\beta \in ]0,\tau]$ and $c>0$ such that $\lambda _i(\Xcov)\leq c i^{-1/\beta }$. 
\end{tassumption}
This assumption is common in the literature, and we will see that the optimal learning rates depend on $\beta$.
It implies the bound $\deff\de \Tr(\rXcov^{-1}\Xcov)\lesssim \lambda ^{-\beta}$ on the effective dimension, which is a key quantity in the analysis (both statements are actually equivalent, see \Cref{s:proof_krr_rates_a1}).
%and is actually equivalent to assuming $\deff \lesssim \lambda ^{-\beta}$ as discussed in \Cref{s:proof_krr_rates_a1}.
%Note that $\n{\rXcov^{-1/2}\fmap(x)}\leq \supfmap \lambda ^{-1}$ always hold which is why $\beta\leq 1$, and as
Note that $\deff=\E_{\ptx\sim \td} \n{\rXcov^{-1/2}\fmap{\ptx}}\leq \esssup_{\ptx\sim \td}\n{\rXcov^{-1/2}\fmap{\ptx}}$, and thus it necessarily holds $\beta \leq \tau$.
For a Gaussian kernel, both $\beta $ and $\tau $ can be chosen arbitrarily close to zero.

Finally, we make an assumption about the regularity of the problem itself. A common assumption occurring in the literature is that $\E\brk{f(X_1)\,\vert\, X_0=\cdot}\in\rkhs$ for every $f\in\rkhs$, 
%for every $\E[f(X_1)\,\vert\, X_0=\cdot]\in\rkhs$ for all $f\in\rkhs$, 
meaning that one can define the Koopman operator directly on the space $\rkhs$, i.e.~the learning problem is \emph{well-specified}. 
However, this assumption is often too strong. Following~\cite[D.1]{kostic2023KoopmanOperatorLearning} we make a different assumption on the cross-covariance remarking that, irrespectively of the choice of RKHS, it holds true whenever the Koopman operator is self-adjoint (i.e.~the dynamics is time-reversible).

\begin{tassumption}{Regularity of $\kop$}{regularity}
	There exists $a>0$ such that $\XYcov\XYcov^* \preccurlyeq  a^2 \Xcov^{2}$.
\end{tassumption}
%
%
%This assumption essentially says that $\rkhs$ is stable by $\kop$, in the sense that $\kop \rkhs\subseteq \rkhs$.
%Although we consider this formulation for simplicity, note that this assumption can be restrictive in some settings and our results hold under a weaker assumption introduced by \cite{kostic2023KoopmanOperatorLearning} as discussed in \Cref{s:relaxed_assumptions}.
%
%We eventually make two assumptions allowing us to control the quantities $\deff\de \E_{x\sim \td}\n{\rXcov^{-1/2}\fmap{x}}=\Tr(\rXcov^{-1}\Xcov)$ and $\esssup_{x\sim \td}\n{\rXcov^{-1/2}\fmap{x}}$, which play a key role in the analysis.
%
%
%
%\textbf{Rates}~~
\paragraph{Rates}
The risk can be decomposed as $\pR(A) = \ER(A)+ \cR_{\HS, 0}$ where $\cR_{\HS,0}$ is a constant and $\ER(A) \de \nhs{ \kop \amftX - \amftX A^* }^2$ corresponds to the excess risk (more details in \Cref{s:risk_decomposition}). 
Optimal learning bounds for the KRR estimator in the context of CME (\ie in Hilbert-Schmidt norm) have been developed in \cite{li2022OptimalRatesRegularized} under \Cref{a:bounded_fmap,a:embedding_property,a:universal_kernel,a:spectral_decay} in well-specified and misspecified settings.
On the other hand, in the context of dynamical systems, \citet[Theorem 1]{kostic2022learning} report the importance of \emph{reduced rank estimators} that have a 
small excess risk in operator norm
\begin{align}
	\ERop(A) &\de \nHlt{ \kop \amftX - \amftX A^* }^2.
	\label{e:population_risk_opnorm}
\end{align}
%The reasoning is that when using the an estimator of the Koopman operator in \cref{eq:spectral_eq,eq:kmd}, one induces the error mainly determined by\GM{Previous should be rephrased but I don't understand it.}
The rationale behind considering the operator norm is that it allows to control the error of the eigenvalues approximation and thus of the KMD \eqref{eq:spectral_eq}, \eqref{eq:kmd} as discussed below.
%As shown in \todo{ref appendix}, replacing the operator norm by the Hilbert-Schmidt norm in \eqref{e:population_risk_opnorm} yields exactly the risk \eqref{e:population_risk}.
Optimal learning bounds in operator norm for KRR, PCR and RRR are established in~\cite{kostic2023KoopmanOperatorLearning}. 
%where also the eigenvalue/eigenfunction learning bounds for time-reversal invariant dynamics have been developed. 
%With this in mind, 
In this work we show that the same optimal rates remain valid for the \emph{\nystrom{}} KRR, PCR and RRR estimators. According to \cite{kostic2022learning} and \cite{kostic2023KoopmanOperatorLearning} these operator norm bounds lead to reliable approximation of the Koompan mode decomposition of \Cref{eq:kmd}.

We now provide our main result.
\begin{ttheorem}{Operator norm error for KRR, \tiid data}{bound_ER_KRR_a1}
	Let \cref{a:bounded_fmap,a:embedding_property,a:universal_kernel,a:regularity,a:spectral_decay} hold.
	Let $(\ptx_i,\pty_i)_{1\leq i\leq n}$ be i.i.d. samples, 
	and let $\py=\px$ be the projection induced by $m$ Nyström landmarks drawn uniformly from $(\ptx_i)_{1\leq i\leq n}$ without replacement. 
	Let $\lambda =c_\lambda n^{-1/(1+\beta )}$ where $c_\lambda $ is a constant given in the proof, and assume
		$n \geq  (c_\lambda /\supfmap^2 )^{1+\beta }$.
	%\begin{align*}
		%n &\geq  (c_\lambda /\supfmap^2 )^{1+\beta }\\
		%m &\geq \max(67, 5c_\tau c_\lambda ^{-\tau }n^{\tau /(1+\beta )})\log\frac{20\supk}{\lambda \delta }.
	%\end{align*}
	Then it holds with probability at least $1-\delta $
	\begin{align*}
		\ERop(\nysest)^{1/2}
		&\lesssim n^{-\frac{1}{2(1+\beta)}}
		\quad\quad\text{provided}\quad\quad
		m \gtrsim \max(1, n^{\tau /(1+\beta )})\log(\nicefrac{n}{\delta}).
	\end{align*}
\end{ttheorem}

The proof is provided in \Cref{s:proof_krr_rates_a1}, but essentially relies on a decomposition involving the terms $\n{\rXcov^{-1/2}(\YXcov-\eYXcov)}$, $\n{\rXcov^{-1/2}(\Xcov-\eXcov)}$, $\n{\rXcov^{-1/2}(\Xcov-\eXcov)\rXcov^{-1/2}}$, as well as bounding the quantity $\nrkhs{\px^\perp\Xcov^{1/2}}$ where $\px^\perp$ denotes the projection on the orthogonal of $\ran(\px)$.
All these terms are bounded using two variants of the Bernstein inequality.
Note that our results can easily be extended to leverage score sampling of the landmarks by bounding term $\nrkhs{\px^\perp\Xcov^{1/2}}$ by means of \cite[Lemma 7]{rudi2016LessMoreNystr}; the same rate could then be obtained using a smaller number $m$ of Nyström points.

The rate $n^{-1/(2(1+\beta))}$ is known to be optimal (up to the log factor) in this setting by assuming an additional lower bound on the decay of the covariance's eigenvalues of the kind $\lambda_i(\Xcov)\gtrsim i^{-1/\beta}$, see \cite[Theorem 7 in D.4]{kostic2023KoopmanOperatorLearning}.
One can see that without particular assumptions ($\beta =\tau =1$), we only need the number $m$ of inducing points to be of the order of $\Omega(\sqrt{n})$ in order to get an optimal rates. For $\tau$ fixed, this number increases when $\beta$ decreases (faster decay of the covariance's spectrum), however note that the optimal rate depends on $\beta$ and also improves in this case. The dependence in $\tau$ is particularly interesting, as for instance with a Gaussian kernel it is known that $\tau$ can be chosen arbitrarily closed to zero~\cite{li2022OptimalRatesRegularized,fischer2020SobolevNormLearning}. In that case, the number $m$ of inducing points can be taken on the order of $\Omega(\log{n})$.

Note that a bound for the Nyström KRR estimator has been derived in Hilbert-Schmidt norm by \citet{ahmad2023SketchSketchOut}. Using the operator norm however allows to derive bounds on the eigenvalues (see discussion below), which is of paramount importance for practical applications.
Moreover, we now provide a bound on the error of PCR and RRR estimators, which are not covered in~\cite{ahmad2023SketchSketchOut}.

\begin{tlemma}{Operator norm error for PCR and RRR, \tiid data}{bound_ER_RRR_a1}
	%Let \Cref{a:bounded_fmap,a:embedding_property,a:universal_kernel,a:regularity,a:spectral_decay} hold.
	%Let $(\ptx_i,\pty_i)_{1\leq i\leq n}$ be \tiid samples, 
	%and let $\py=\px$ be the projection induced by $m$ Nyström landmarks drawn uniformly from $(\ptx_i)_{1\leq i\leq n}$ without replacement.
	Under the assumptions of \Cref{r:bound_ER_KRR_a1}, 
	taking $\lambda =c_\lambda n^{-1/(1+\beta )}$ with $c_\lambda $ as in \Cref{r:bound_ER_KRR_a1}, 
		$n \geq  (c_\lambda /\supfmap^2 )^{1+\beta }$,
	and provided
	\begin{align*}
		%n &\geq  (c_\lambda /\supfmap^2 )^{1+\beta }\\
		%m &\geq \tfrac{1}{1+\beta}\max(67, 5c_\tau c_\lambda ^{-\tau }n^{\tau /(1+\beta )})\log\frac{20\supk n}{c_\lambda\delta }.
		m &\gtrsim \max(1, n^{\tau /(1+\beta )})\log(\nicefrac{n}{\delta}),
	\end{align*}
	it holds with probability at least $1-\delta$
	\begin{align*}
		\ERop(\rrrest)^{1/2} & \lesssim c_{\rm RRR}\,n^{-\frac{1}{2(1+\beta)}},\;\text{ for } r \text{ s.t. } \sigma_{r+1}(\mftYcX) < \min(\sigma_{r}(\mftYcX), n^{-\frac{1}{2(1+\beta)}})  \\
	%\end{align*}
 %and 
     %\begin{align*}
	 	\text{and}\quad
		\ERop(\pcrest)^{1/2} & \lesssim c_{\rm PCR}\,n^{-\frac{1}{2(1+\beta)}},\;\text{ for } r> n^{\frac{1}{\beta(1+\beta)}},
	\end{align*}
 where  $c_{\rm RRR} = (\sigma_{r}^2(\mftYcX) - \sigma_{r+1}^2(\mftYcX))^{-1}$ and $c_{\rm PCR} = (\sigma_{r}(\mftX) - \sigma_{r+1}(\mftX))^{-1}$ are the problem dependant constants.
\end{tlemma}

Note that when rank of $\kop$ is $r$, then there is no restriction on $r$ for the RRR estimator, while for PCR the choice of $r$ depends on the spectral decay property of the kernel. In general, if $r> n^{\frac{1}{\beta(1+\beta)}}$, then
%\[
$
\sigma_{r+1}(\mftYcX) \leq  \sigma_{r+1}(\mftX) \lesssim n^{-1/(2(1+\beta))},
$
%\]
which implies that RRR estimator can achieve the same rate of PCR but with smaller rank. 
Again the rate is sharp (up to the log factor) in this setting~\cite{kostic2023KoopmanOperatorLearning}. 
% (We say it above already).

\paragraph{Koopman mode decomposition}
%According to \cite[Theorem 1]{kostic2022learning}, 
%working in operator norm allows us to bound the error of our estimators for dynamic mode decomposition, as well as to quantify how close the Koopman eigenpairs $(\lambda_i, \mftX \varphi_i)$ are to being eigenpairs of the considered estimator. \GM{In \Cref{s:experiments}, we empirically show that the proposed estimators accurately learn the Koopman spectrum.}
%\todo{VK: The paragraph above maybe we wish to say something like this:
According to \cite[Theorem 1]{kostic2022learning}, 
working in operator norm allows us to bound the error of our estimators for dynamic mode decomposition, as well as to quantify how close the eigenpairs $(\hat{\lambda}_i, \hat{\varphi}_i)$  of an estimator $\hat{A}^*$ are to being eigenpairs of the Koopman operator. Namely, recalling that for function $\hat{\varphi}_i$, the corresponding candidate for Koopman eigenfunction in $\lt$ space is $\amftX \hat{\varphi}_i$, one has $\n{\kop(\amftX \hat{\varphi}_i) - \hat{\lambda}_i (\amftX \hat{\varphi}_i) } / \n{\amftX \hat{\varphi}_i}\leq \ERop(\hat{A})^{1/2} \n{\hat{\varphi}_i}/\n{\amftX\hat{\varphi}_i}$. While eigenvalue and eigenfunction learning rates were studied, under additional assumptions, in \cite{kostic2023KoopmanOperatorLearning}, where the operator norm error rates were determinant, here, in \Cref{s:experiments}, we empirically show that the proposed estimators accurately learn the Koopman spectrum. We refer the reader to \Cref{s:kmd_app} for the details on computation of eigenvalues, eigenfunctions and KMD of an estimator in practice.
%}

\paragraph{Dealing with non-\tiid data}
The previous results hold for \tiid data, which is not a very realistic assumption when learning from sampled trajectories. 
Our results can however easily be extended to $\beta$-mixing processes by considering random variables $Z_i = \sum_{j=1}^k X_{i+j}$ (thus representing portions of the trajectory) sufficiently separated in time to be nearly independent.
We now consider a trajectory $\ptx_1,\dots,\ptx_{n+1}$ with $\ptx_1\sim \td$ and $\ptx_{t+1}\sim p(\ptx_t,\cdot)$ for $t\in[1,n]$, and use \Cref{r:mixing} (re-stated from \cite{kostic2022learning})
which allows to translate concentration results on the $Z_i$ to concentration on the $X_i$ by means of the $\beta$-mixing coefficients defined as
	$\beta_X(k) %&
	\de \sup_{B\in \cB \otimes \cB} \absv*{\jd_k(B)-(\td \times \td)(B)}$
where $\jd_k$ denotes the joint probability of $(X_t, X_{t+k})$.
%it is known that \tiid translation can be transfered to $\beta$-mixing processes.
%The following lemma can then be leveraged, where the probability on the rhs can be controlled using our concentration results in the iid setting.
Using this result the concentration results provided in appendix can thus be generalied to the $\beta$-mixing setting, and apart from logarithmic dependencies we essentially obtain similar results to the \tiid setting except that the sample size $n$ is replaced by $p\approx n/(2k)$.

%%%%%%%%%%%%%%%%%%%%%%%%%%%%%%%%%%%%%%%%%%%%%%%%%%%%%%%%%%%%%%%%%%%%%%%
%                             Experiments                             %
%%%%%%%%%%%%%%%%%%%%%%%%%%%%%%%%%%%%%%%%%%%%%%%%%%%%%%%%%%%%%%%%%%%%%%%

\section{Experimental validation}\label{s:experiments}
In this section we show how the estimators proposed in \cref{s:estimators} perform in various scenarios, ranging from synthetic low dimensional ODEs to large-scale molecular dynamics simulations. The code for reproducing all experiments is available online.
Our initial aim is to demonstrate the speed of NysPCR and NysRRR, compared to the recently proposed alternative Streaming KAF (sKAF)~\cite{giannakis2021LearningForecastDynamical}.
Then we show that their favorable scaling properties make it possible to train on large molecular dynamics datasets without any subsampling.
In particular we run a metastability analysis of the alanine dipeptide and the Trp-cage protein, showcasing the accuracy of our models' eigenvalue and eigenfunction estimates, as well as their efficiency on massive datasets ($> \num{500000}$ points)

\begin{figure}
	\centering
    \begin{minipage}[t]{.48\textwidth}
        % \vspace{0pt}
        \includegraphics[width=0.98\textwidth]{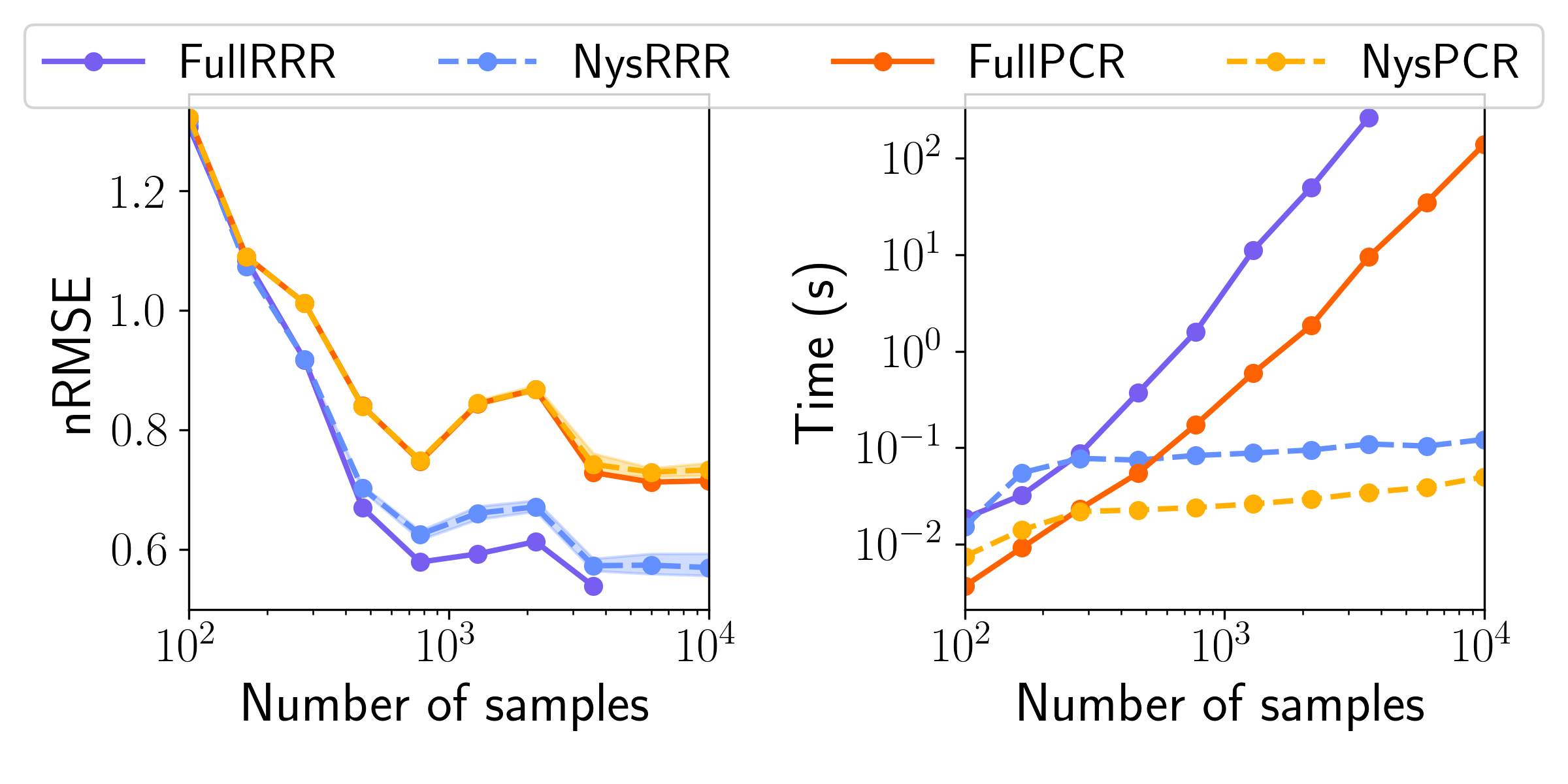}
        \caption{Full and \nystrom{} estimators trained on L63 with increasing $n$. Error (\emph{left}) and running time (\emph{right}) are plotted to show efficiency gains without accuracy loss with the \nystrom{} approximation. RBF($\sigma=3.5$) kernel, $r=25$ principal components and $m=250$ inducing points.}
        \label{fig:l63-full}
    \end{minipage}
    \hfill
    \begin{minipage}[t]{.48\textwidth}
        % \vspace{0pt}
        \includegraphics[width=0.98\textwidth]{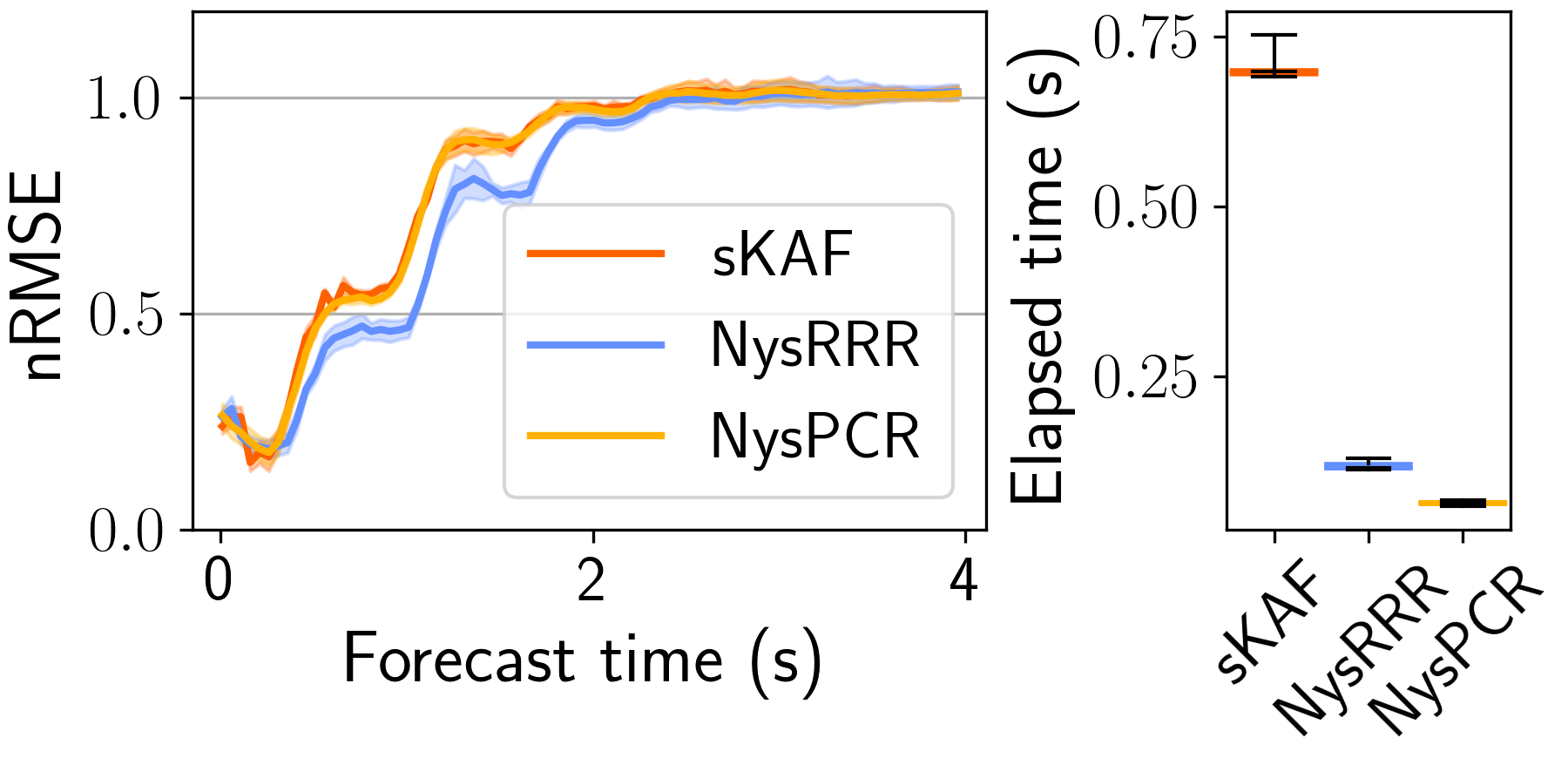}
        \caption{\nystrom{} and sKAF estimators trained on L63 for increasing forecast horizons; the error (\emph{left}) and overall running times (\emph{right)} are shown. We used a RBF kernel with $\sigma = 3.5$, $r=50$, $m=250$ (for \nystrom{} methods) and $\sqrt{n}\log n$ random features (for sKAF).}
        \label{fig:l63-incq}
    \end{minipage}
\end{figure}

\textbf{Efficiency Benchmarks on Lorenz '63}~~
The chaotic Lorenz '63 system~\cite{lorenz63} consists of 3 ODEs with no measurement noise. With this toy dynamical system we can easily compare the \nystrom{} estimators to two alternatives: \begin{enumerate*}
    \item the corresponding \emph{exact} estimators and
    \item the sKAF algorithm which also uses randomized linear algebra to improve the efficiency of PCR.
\end{enumerate*}
In this setting we sample long trajectories from the system, keeping the first points for training (the number of training points varies for the first experiment, and is fixed to \num{10000} for the second, see \cref{fig:l63-incq}), and the subsequent ones for testing.
In \Cref{fig:l63-full} we compare the run-time and accuracy with of NysPCR and NysRRR versus their full counterparts. To demonstrate the different scaling regimes we fix the number of inducing points to 250 and increase the number of data points $n$. The accuracy of the two solvers (as measured with the normalized RMSE metric (nRMSE)~\cite{giannakis2021LearningForecastDynamical} on the first variable) is identical for PCR and close for RRR, but the running time of the approximate solvers increases much slower with $n$ than that of the exact solvers. Each experiment is repeated 20 times to display error bars over the choice of \nystrom{} centers.
In the second experiment, shown in \cref{fig:l63-incq}, we reproduce the setting of \cite{giannakis2021LearningForecastDynamical} by training at increasingly long forecast horizons. Plotting the nRMSE we verify that sKAF and NysPCR converge to very similar accuracy values, although NysPCR is approximately $10$ times faster. NysRRR instead offers slightly better accuracy, at the expense of a higher running time compared to NysPCR. Error bars are the standard deviation of nRMSE over 5 successive test sets with \num{10000} points each.

% \subsection{Timescale analysis}
% We show how our models can be used to analyse the implied time-scales of a 2D hidden Markov model (HMM), which is not linearly separable.

% \begin{figure}
%     \centering
%     \includegraphics[width=0.7\textwidth]{figures/msm/full_annotated.jpg}
%     \caption{implied time-scale analysis of a 2D HMM. a) depicts the model along with transition probabilities. Both states draw data from a non-isotropic Gaussian, which is then transformed by $\bm{x}_1 = \bm{x}_1 + \sqrt{\bm{x}_0}$. In b) we show the first non-trivial eigenfunction of a RRR model, overlayed on top of some the data. c), d) and e) show how the learned models change as a function of the lag time. In a) the correlation between eigenfunction and hidden state of the Markov model, in c) the forecast error, and in e) the implied time-scale (ITS) which measures error in the eigenvalue estimates.}
%     \label{fig:msm}
% \end{figure}

\textbf{Molecular dynamics datasets}~~
An important application of Koopman operator theory is in the analysis of molecular dynamics (MD) datasets, where the evolution of a molecule's atomic positions as they evolve over time is modelled. 
Interesting systems are very high dimensional, with hundreds or thousands of atoms. Furthermore, trajectories are generated at very short time intervals ($< \SI{1}{\nano\second}$) but interesting events (e.g.~protein folding/unfolding) occur at timescales on the order of at least \SI{10}{\micro\second}, so that huge datasets are needed to have a few samples of the rare events.
The top eigenfunctions of the Koopman operator learned on such trajectories can be used to project the high-dimensional state space onto low-dimensional coordinates which capture the long term, slow dynamics.

\begin{figure}
    \centering
	\vspace{-0.2cm}
    \includegraphics[width=0.8\textwidth]{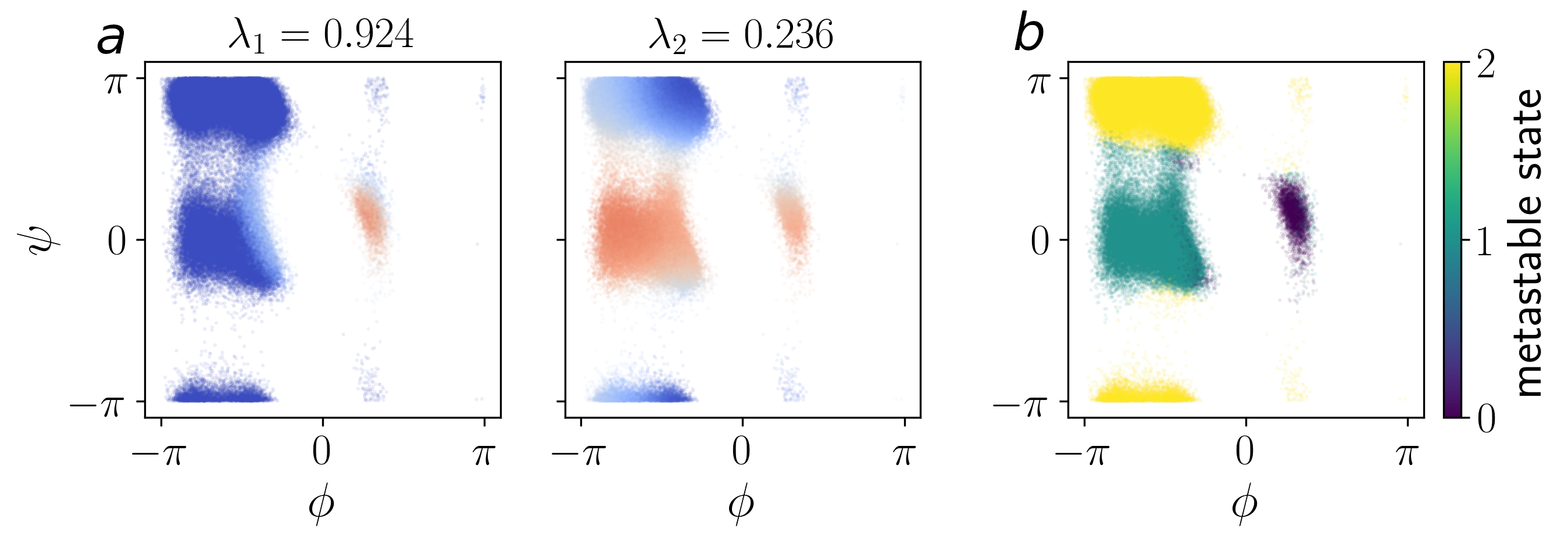}
	\vspace{-0.2cm}
    \caption[Alanine dipeptide]{Dynamics of the alanine dipeptide (lag-time 100), Nystr{\"o}m RRR model. On the left the first two non-constant eigenfunctions, overlaid in color on the Ramachandran plot which fully describes the metastable states. On the right the three states of a PCCA+ model trained on the eigenfunctions.}
    \label{fig:alanine}
\end{figure}

We take three \SI{250}{\nano\second} long simulations sampled at \SI{1}{\pico\second} of the alanine dipeptide~\cite{wehmeyer_ticaautoencoder18}, which is often taken as a model system for molecular dynamics~\cite{nuske17,VACNuske2014}.
We use the pairwise distances between heavy atoms as features, yielding a 45-dimensional space.
We train a NysRRR model with \num{10000} centers on top of the full dataset (\num{449940} points are used for training, the rest for validation and testing) with lag time $\SI{100}{\pico\second}$, and recover a 2-dimensional representation which correlates well with the $\phi, \psi$ backbone dihedral angles of the molecule, known to capture all relevant long-term dynamics. \Cref{fig:alanine}a shows the top two eigenfunctions overlaid onto $\phi, \psi$, the first separates the slowest transition between low and high $\phi$; the second separates low and high $\psi$. 
The implied time-scales from the first two non-trivial eigenvalues are \SI{1262}{\pico\second} and \SI{69}{\pico\second}, which are close to the values reported by~\citet{nuske17} (\SI{1400}{\pico\second} and \SI{70}{\pico\second}) who used a more complex post-processing procedure to identify time-scales.
We then train a PCCA+~\cite{deuflhard_pcca} model on the first three eigenfunctions to obtain three states, as shown in \cref{fig:alanine}b. PCCA+ acts on top of a fine clustering (in our case obtained with k-means, $k=50$), to find the set of maximally stable states by analyzing transitions between the fine clusters. The coarse clusters clearly correspond to the two transitions described above.

\begin{figure}
    \centering
    \includegraphics[width=0.7\textwidth]{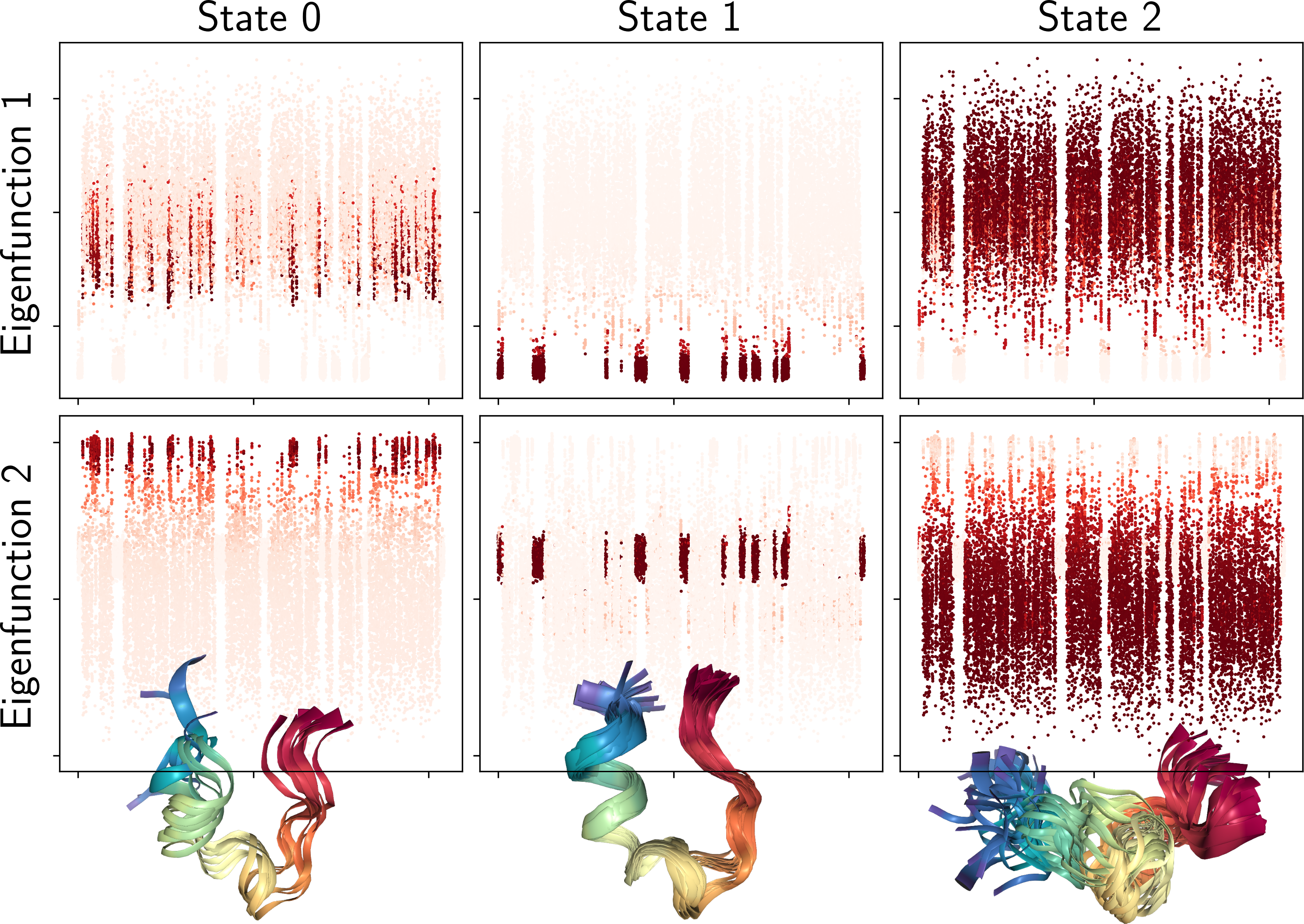}
    \caption{First eigenfunctions for Trp-cage dynamics, colored according to the membership probability for each state in a PCCA+ model. The bottom insets show a few overlaid structures from each state. The first eigenfunction exhibits a strong linear separation between state 1 (folded) and the other states. The second separates between state 0 (partially folded) ant the rest. NysRRR model trained with $m=5000$, $r=10$, RBF($\sigma=0.02$) kernel, $\lambda=10^{-10}$.}
    \label{fig:trp_cage}
\end{figure}

Finally we take a \SI{208}{\micro\second} long simulation of the fast-folding Trp-cage protein~\cite{fastfolding11}, sampled every \SI{0.2}{\nano\second}. Again, the states are the pairwise distances between non-hydrogen atoms belonging to the protein, in \num{10296} dimensions. 
A NysRRR model is trained on \num{626370} points, using \num{5000} centers in approximately 10 minutes. Note that without sketching this would be a completely intractable problem. 
Using a lag-time of \SI{10}{\nano\second} we observe a spectral gap between the third and fourth eigenvalues, hence we train a PCCA+ model on the first 3 eigenfunctions to obtain the states shown in \cref{fig:trp_cage}. 
The first non-trivial Koopman eigenvector effectively distinguishes between the folded (state 1) and unfolded states as is evident from the first row of \cref{fig:trp_cage}. The second one instead can be used to identify a partially folded state of the protein (state 0), as can be seen from the insets in \cref{fig:trp_cage}.

\section{Conclusions}\label{s:conclusions}

We introduced three efficient kernel-based estimators of the Koopman operator relying on random projections, and provided a bound on their excess risk in operator norm -- which is of paramount importance to control the accuracy of Koopman mode decomposition.
Random projections allow to process efficiently even the longest trajectories, 
%while they retain minimal-optimal rates.
%Furthermore, these gains come for free, as they are not offset by any drop in the theoretical learning rates.
and these gains come for free as our estimators still enjoy optimal theoretical learning rates.
We leave for future work the refinement our analysis under \eg an additional source condition assumption or in the misspecified setting. Another future research direction shall be to devise ways to further reduce the computational complexity of the estimators.

\section{Acknowledgements}\label{s:ack}
This paper is part of a project that has received funding from the European Research Council (ERC) under the European Union's Horizon 2020 research and innovation programme (grant agreement No. 819789). 
L. R. acknowledges the financial support of the European Research Council (grant SLING 819789), the AFOSR projects FA9550-18-1-7009, FA9550-17-1-0390 and BAA-AFRL-AFOSR-2016-0007 (European Office of Aerospace Research and Development), the EU H2020-MSCA-RISE project NoMADS - DLV-777826, and the Center for Brains, Minds and Machines (CBMM), funded by NSF STC award CCF-1231216.
M. P., V. K. and P. N. acknowledge financial support from PNRR MUR project PE0000013-FAIR and the European Union (Projects 951847 and 101070617).
\bibliography{biblio}

\begin{thebibliography}{66}
\providecommand{\natexlab}[1]{#1}
\providecommand{\url}[1]{\texttt{#1}}
\expandafter\ifx\csname urlstyle\endcsname\relax
  \providecommand{\doi}[1]{doi: #1}\else
  \providecommand{\doi}{doi: \begingroup \urlstyle{rm}\Url}\fi

\bibitem[Abedsoltan et~al.(2023)Abedsoltan, Belkin, and Pandit]{belkinlarge23}
Amirhesam Abedsoltan, Mikhail Belkin, and Parthe Pandit.
\newblock Toward large kernel models, 2023.
\newblock \href{https://arxiv.org/abs/2302.02605}{{\ttfamily arXiv:2302.02605
  [cs.LG]}}.

\bibitem[Adi Ben-Israel(2003)]{adiben-israel2003GeneralizedInversesTheory}
Thomas N. E. Greville~(auth.) Adi Ben-Israel.
\newblock \emph{Generalized Inverses: {{Theory}} and Applications}.
\newblock {{CMS}} Books in Mathematics. {Springer}, 2 edition, 2003.

\bibitem[Ahmad et~al.(2023)Ahmad, Brogat-Motte, Laforgue, and
  Florence]{ahmad2023SketchSketchOut}
Tamim~El Ahmad, Luc Brogat-Motte, Pierre Laforgue, and d'Alché-Buc Florence.
\newblock Sketch {{In}}, {{Sketch Out}}: {{Accelerating}} both {{Learning}} and
  {{Inference}} for {{Structured Prediction}} with {{Kernels}}, 2023.
\newblock \href{https://arxiv.org/abs/2302.10128}{{\ttfamily
  arxiv:2302.10128}}.

\bibitem[Alexander and Giannakis(2020)]{kaf}
Romeo Alexander and Dimitrios Giannakis.
\newblock Operator-theoretic framework for forecasting nonlinear time series
  with kernel analog techniques.
\newblock \emph{Physica D: Nonlinear Phenomena}, 409, 2020.
\newblock \doi{https://doi.org/10.1016/j.physd.2020.132520}.

\bibitem[Bach(2013)]{bachnystrom_13}
Francis Bach.
\newblock Sharp analysis of low-rank kernel matrix approximations.
\newblock \emph{Journal of Machine Learning Research}, 30, 2013.

\bibitem[Baddoo et~al.(2022)Baddoo, Herrmann, McKeon, and
  Brunton]{baddooSparseKernel2022}
Peter~J. Baddoo, Benjamin Herrmann, Beverley~J. McKeon, and Steven~L. Brunton.
\newblock Kernel learning for robust dynamic mode decomposition: linear and
  nonlinear disambiguation optimization.
\newblock \emph{Proceedings of the Royal Society A}, 2022.
\newblock \doi{http://doi.org/10.1098/rspa.2021.0830}.

\bibitem[Brunton et~al.(2021)Brunton, Budišić, Kaiser, and
  Kutz]{brunton2021ModernKoopmanTheory}
Steven~L. Brunton, Marko Budišić, Eurika Kaiser, and J.~Nathan Kutz.
\newblock Modern {{Koopman}} theory for dynamical systems, 2021.
\newblock \href{https://arxiv.org/abs/2102.12086}{{\ttfamily
  arxiv:2102.12086}}.

\bibitem[Caponnetto and Vito(2007)]{caponnetto07}
Andrea Caponnetto and Ernesto~De Vito.
\newblock Optimal rates for the regularized least-squares algorithm.
\newblock \emph{Foundations of Computational Mathematics}, 7\penalty0
  (3):\penalty0 331 -- 368, 2007.
\newblock \doi{10.1007/s10208-006-0196-8}.

\bibitem[Chatalic et~al.(2022)Chatalic, Schreuder, Rosasco, and
  Rudi]{chatalic2022NystromKernelMean}
Antoine Chatalic, Nicolas Schreuder, Lorenzo Rosasco, and Alessandro Rudi.
\newblock Nyström {{Kernel Mean Embeddings}}.
\newblock In \emph{Proceedings of the 39th {{International Conference}} on
  {{Machine Learning}}}, volume 162 of \emph{Proceedings of {{Machine Learning
  Research}}}, pages 3006--3024. {PMLR}, 2022.

\bibitem[Ciliberto et~al.(2016)Ciliberto, Rosasco, and
  Rudi]{ciliberto2016ConsistentRegularizationApproacha}
Carlo Ciliberto, Lorenzo Rosasco, and Alessandro Rudi.
\newblock A {{Consistent Regularization Approach}} for {{Structured
  Prediction}}.
\newblock In \emph{Advances in {{Neural Information Processing Systems}}},
  volume~29. {Curran Associates, Inc.}, 2016.

\bibitem[Ciliberto et~al.(2022)Ciliberto, Rosasco, and
  Rudi]{ciliberto2022GeneralFrameworkConsistent}
Carlo Ciliberto, Lorenzo Rosasco, and Alessandro Rudi.
\newblock A general framework for consistent structured prediction with
  implicit loss embeddings.
\newblock \emph{Journal of Machine Learning Research}, 21\penalty0
  (1):\penalty0 98:3852--98:3918, 2022.

\bibitem[Da~Prato and Zabczyk(1996)]{Prato1996}
G.~Da~Prato and J.~Zabczyk.
\newblock \emph{Ergodicity for Infinite Dimensional Systems}.
\newblock London Mathematical Society Lecture Note Series. Cambridge University
  Press, 1996.
\newblock \doi{10.1017/CBO9780511662829}.

\bibitem[Dellnitz and Junge(1999)]{dellnitz99}
Michael Dellnitz and Oliver Junge.
\newblock On the approximation of complicated dynamical behavior.
\newblock \emph{SIAM Journal on Numerical Analysis}, 36\penalty0 (2):\penalty0
  491--515, 1999.
\newblock \doi{10.1137/S0036142996313002}.

\bibitem[Deuflhard and Weber(2005)]{deuflhard_pcca}
Peter Deuflhard and Marcus Weber.
\newblock Robust perron cluster analysis in conformation dynamics.
\newblock \emph{Linear Algebra and its Applications}, 398:\penalty0 161--184,
  2005.
\newblock \doi{https://doi.org/10.1016/j.laa.2004.10.026}.

\bibitem[Drineas et~al.(2012)Drineas, Magdon-Ismail, Mahoney, and
  Woodruff]{drineas_levscore12}
Petros Drineas, Malik Magdon-Ismail, Michael~W. Mahoney, and David~P. Woodruff.
\newblock Fast approximation of matrix coherence and statistical leverage.
\newblock \emph{Journal of Machine Learning Research}, 13\penalty0 (1), 2012.

\bibitem[Fischer and Steinwart(2020)]{fischer2020SobolevNormLearning}
Simon Fischer and Ingo Steinwart.
\newblock Sobolev norm learning rates for regularized least-squares algorithms.
\newblock \emph{Journal of Machine Learning Research}, 21\penalty0
  (1):\penalty0 8464--8501, 2020.

\bibitem[Froyland et~al.(2014)Froyland, Gottwald, and Hammerlindl]{froyland14}
Gary Froyland, Georg~A. Gottwald, and Andy Hammerlindl.
\newblock A computational method to extract macroscopic variables and their
  dynamics in multiscale systems.
\newblock \emph{SIAM Journal on Applied Dynamical Systems}, 13\penalty0
  (4):\penalty0 1816--1846, 2014.
\newblock \doi{10.1137/130943637}.

\bibitem[Giannakis et~al.(2021)Giannakis, Henriksen, Tropp, and
  Ward]{giannakis2021LearningForecastDynamical}
Dimitris Giannakis, Amelia Henriksen, Joel~A. Tropp, and Rachel Ward.
\newblock Learning to {{Forecast Dynamical Systems}} from {{Streaming Data}},
  2021.
\newblock \href{https://arxiv.org/abs/2109.09703}{{\ttfamily
  arxiv:2109.09703}}.

\bibitem[Gittens and Mahoney(2016)]{gittens16}
Alex Gittens and Michael~W. Mahoney.
\newblock Revisiting the nystr\"{o}m method for improved large-scale machine
  learning.
\newblock \emph{Journal of Machine Learning Research}, 17:\penalty0 3977--4041,
  2016.

\bibitem[Hofmann et~al.(2008)Hofmann, Sch{\"o}lkopf, and
  Smola]{hofmann_kernels_08}
Thomas Hofmann, Bernhard Sch{\"o}lkopf, and Alexander~J. Smola.
\newblock Kernel methods in machine learning.
\newblock \emph{The Annals of Statistics}, 36\penalty0 (3):\penalty0 1171 --
  1220, 2008.
\newblock \doi{10.1214/009053607000000677}.

\bibitem[Izenman(1975)]{rrrIzenman1975}
Alan~Julian Izenman.
\newblock Reduced-rank regression for the multivariate linear model.
\newblock \emph{Journal of Multivariate Analysis}, 5\penalty0 (2):\penalty0
  248--264, 1975.
\newblock \doi{https://doi.org/10.1016/0047-259X(75)90042-1}.

\bibitem[Klus et~al.(2016)Klus, Koltai, and Schütte]{edmdKlus2016}
Stefan Klus, Péter Koltai, and Christof Schütte.
\newblock On the numerical approximation of the {Perron-Frobenius} and
  {Koopman} operator.
\newblock \emph{Journal of Computational Dynamics}, 3\penalty0 (1):\penalty0
  51--79, 2016.
\newblock ISSN 2158-2491.
\newblock \doi{10.3934/jcd.2016003}.

\bibitem[Klus et~al.(2020)Klus, Schuster, and Muandet]{kerneleidec_klus_20}
Stefan Klus, Ingmar Schuster, and Krikamol Muandet.
\newblock Eigendecompositions of transfer operators in reproducing kernel
  hilbert spaces.
\newblock \emph{Journal of Nonlinear Science}, 30\penalty0 (1):\penalty0
  283--315, 2020.
\newblock \doi{10.1007/s00332-019-09574-z}.

\bibitem[Koopman(1931)]{koopman31}
B.~O. Koopman.
\newblock Hamiltonian systems and transformation in hilbert space.
\newblock \emph{Proceedings of the National Academy of Sciences}, 17\penalty0
  (5):\penalty0 315--318, 1931.
\newblock \doi{10.1073/pnas.17.5.315}.

\bibitem[Koopman and v.~Neumann(1932)]{koopman32}
B.~O. Koopman and J.~v.~Neumann.
\newblock Dynamical systems of continuous spectra.
\newblock \emph{Proceedings of the National Academy of Sciences}, 18\penalty0
  (3):\penalty0 255--263, 1932.
\newblock \doi{10.1073/pnas.18.3.255}.

\bibitem[Kostic et~al.(2022)Kostic, Novelli, Maurer, Ciliberto, Rosasco, and
  Pontil]{kostic2022learning}
Vladimir Kostic, Pietro Novelli, Andreas Maurer, Carlo Ciliberto, Lorenzo
  Rosasco, and Massimiliano Pontil.
\newblock Learning {{Dynamical Systems}} via {{Koopman Operator Regression}} in
  {{Reproducing Kernel Hilbert Spaces}}, 2022.
\newblock \href{https://arxiv.org/abs/2205.14027}{{\ttfamily arXiv:2205.14027
  [cs.LG]}}.

\bibitem[Kostic et~al.(2023)Kostic, Lounici, Novelli, and
  Pontil]{kostic2023KoopmanOperatorLearning}
Vladimir Kostic, Karim Lounici, Pietro Novelli, and Massimiliano Pontil.
\newblock Koopman operator learning: Sharp spectral rates and spurious
  eigenvalues, 2023.
\newblock \href{https://arxiv.org/abs/2302.02004}{{\ttfamily arXiv:2302.02004
  [cs.LG]}}.

\bibitem[Kutz et~al.(2016)Kutz, Brunton, Brunton, and Proctor]{kutz_book_16}
{J. Nathan} Kutz, {Steven L.} Brunton, {Binghi W.} Brunton, and Joshua Proctor.
\newblock \emph{Dynamic Mode Decomposition: Data-Driven Modeling of Complex
  Systems}.
\newblock SIAM, 2016.

\bibitem[Li et~al.(2017)Li, Dietrich, Bollt, and Kevrekidis]{li_edmdlearned_17}
Qianxiao Li, Felix Dietrich, Erik~M. Bollt, and Ioannis~G. Kevrekidis.
\newblock Extended dynamic mode decomposition with dictionary learning: A
  data-driven adaptive spectral decomposition of the koopman operator.
\newblock \emph{Chaos: An Interdisciplinary Journal of Nonlinear Science},
  27\penalty0 (10), 2017.
\newblock \doi{10.1063/1.4993854}.

\bibitem[Li et~al.(2022)Li, Meunier, Mollenhauer, and
  Gretton]{li2022OptimalRatesRegularized}
Zhu Li, Dimitri Meunier, Mattes Mollenhauer, and Arthur Gretton.
\newblock Optimal {{Rates}} for {{Regularized Conditional Mean Embedding
  Learning}}, 2022.
\newblock \href{https://arxiv.org/abs/2208.01711}{{\ttfamily
  arXiv:2208.01711}}.

\bibitem[Lin and Cevher(2020)]{lin2020OptimalConvergenceDistributed}
Junhong Lin and Volkan Cevher.
\newblock Optimal convergence for distributed learning with stochastic gradient
  methods and spectral algorithms.
\newblock \emph{Journal of Machine Learning Research}, 21\penalty0
  (147):\penalty0 1--63, 2020.

\bibitem[Lindorff-Larsen et~al.(2011)Lindorff-Larsen, Piana, Dror, and
  Shaw]{fastfolding11}
Kresten Lindorff-Larsen, Stefano Piana, Ron~O. Dror, and David~E. Shaw.
\newblock How fast-folding proteins fold.
\newblock \emph{Science}, 334\penalty0 (6055):\penalty0 517--520, 2011.
\newblock \doi{10.1126/science.1208351}.

\bibitem[Lorenz(1963)]{lorenz63}
Edward~N. Lorenz.
\newblock Deterministic nonperiodic flow.
\newblock \emph{Journal of Atmospheric Sciences}, 20\penalty0 (2):\penalty0 130
  -- 141, 1963.
\newblock \doi{https://doi.org/10.1175/1520-0469(1963)020<0130:DNF>2.0.CO;2}.

\bibitem[Lusch et~al.(2018)Lusch, Kutz, and Brunton]{lusch_deepkoop_18}
Bethany Lusch, J.~Nathan Kutz, and Steven~L. Brunton.
\newblock Deep learning for universal linear embeddings of nonlinear dynamics.
\newblock \emph{Nature Communications}, 9\penalty0 (1), 2018.
\newblock \doi{10.1038/s41467-018-07210-0}.

\bibitem[Meanti et~al.(2020)Meanti, Carratino, Rosasco, and
  Rudi]{falkonlibrary2020}
Giacomo Meanti, Luigi Carratino, Lorenzo Rosasco, and Alessandro Rudi.
\newblock Kernel methods through the roof: handling billions of points
  efficiently.
\newblock In \emph{Advances in Neural Information Processing Systems 32}, 2020.

\bibitem[Mezi\`c(1994)]{mezic_phd}
Igor Mezi\`c.
\newblock \emph{On the geometrical and statistical properties of dynamical
  systems: Theory and applications}.
\newblock PhD thesis, ProQuest LLC, California Institute of Technology, 1994.

\bibitem[Mezi{\'c}(2021)]{mezic_koop_21}
Igor Mezi{\'c}.
\newblock Koopman operator, geometry, and learning of dynamical systems.
\newblock \emph{Notices of the American Mathematical Society}, 68\penalty0
  (7):\penalty0 1087--1105, 2021.

\bibitem[Molgedey and Schuster(1994)]{ticaMolgedey1994}
L.~Molgedey and H.~G. Schuster.
\newblock Separation of a mixture of independent signals using time delayed
  correlations.
\newblock \emph{Phys. Rev. Lett.}, 72:\penalty0 3634--3637, 1994.
\newblock \doi{10.1103/PhysRevLett.72.3634}.

\bibitem[Mollenhauer et~al.(2020)Mollenhauer, Schuster, Klus, and
  Sch{\"u}tte]{mollenhauer_svd20}
Mattes Mollenhauer, Ingmar Schuster, Stefan Klus, and Christof Sch{\"u}tte.
\newblock Singular value decomposition of operators on reproducing kernel
  hilbert spaces.
\newblock In \emph{Advances in Dynamics, Optimization and Computation}, pages
  109--131, 2020.

\bibitem[Muandet et~al.(2017)Muandet, Fukumizu, Sriperumbudur, and
  Schölkopf]{muandet2017KernelMeanEmbedding}
Krikamol Muandet, Kenji Fukumizu, Bharath Sriperumbudur, and Bernhard
  Schölkopf.
\newblock Kernel mean embedding of distributions: A review and beyond.
\newblock \emph{Foundations and Trends in Machine Learning}, 10\penalty0
  (1-2):\penalty0 1--141, 2017.
\newblock \doi{10.1561/2200000060}.

\bibitem[No\'{e} and N\"{u}ske(2013)]{VACNoe2013}
Frank No\'{e} and Feliks N\"{u}ske.
\newblock A variational approach to modeling slow processes in stochastic
  dynamical systems.
\newblock \emph{Multiscale Modeling \& Simulation}, 11\penalty0 (2):\penalty0
  635--655, 2013.
\newblock \doi{10.1137/110858616}.

\bibitem[Nüske et~al.(2014)Nüske, Keller, Pérez-Hernández, Mey, and
  Noé]{VACNuske2014}
Feliks Nüske, Bettina~G. Keller, Guillermo Pérez-Hernández, Antonia S. J.~S.
  Mey, and Frank Noé.
\newblock Variational approach to molecular kinetics.
\newblock \emph{Journal of Chemical Theory and Computation}, 10\penalty0
  (4):\penalty0 1739 -- 1752, 2014.
\newblock \doi{10.1021/ct4009156}.

\bibitem[Nüske et~al.(2017)Nüske, Wu, Prinz, Wehmeyer, Clementi, and
  Noé]{nuske17}
Feliks Nüske, Hao Wu, Jan-Hendrik Prinz, Christoph Wehmeyer, Cecilia Clementi,
  and Frank Noé.
\newblock {M}arkov state models from short non-equilibrium simulations —
  analysis and correction of estimation bias.
\newblock \emph{The Journal of Chemical Physics}, 146\penalty0 (9), 2017.
\newblock \doi{10.1063/1.4976518}.

\bibitem[Pinelis and Sakhanenko(1986)]{pinelis1986RemarksInequalitiesLarge}
I.~F. Pinelis and A.~I. Sakhanenko.
\newblock Remarks on inequalities for large deviation probabilities.
\newblock \emph{Theory of Probability \& Its Applications}, 30\penalty0
  (1):\penalty0 143--148, 1986.
\newblock \doi{10.1137/1130013}.

\bibitem[Pérez-Hernández et~al.(2013)Pérez-Hernández, Paul, Giorgino,
  De~Fabritiis, and Noé]{ticaNoe2013}
Guillermo Pérez-Hernández, Fabian Paul, Toni Giorgino, Gianni De~Fabritiis,
  and Frank Noé.
\newblock Identification of slow molecular order parameters for markov model
  construction.
\newblock \emph{The Journal of Chemical Physics}, 139\penalty0 (1), 07 2013.
\newblock \doi{10.1063/1.4811489}.

\bibitem[Rahimi and Recht(2008)]{rahimi_random_2008}
Ali Rahimi and Benjamin Recht.
\newblock Random features for large-scale kernel machines.
\newblock In \emph{{NeurIPS} 20}, 2008.

\bibitem[Rahimi and Recht(2009)]{rahimi09}
Ali Rahimi and Benjamin Recht.
\newblock Weighted sums of random kitchen sinks: Replacing minimization with
  randomization in learning.
\newblock In \emph{Advances in Neural Information Processing Systems 21}, 2009.

\bibitem[Rowley et~al.(2009)Rowley, Mezi\`c, Bagheri, Schlatter, and
  Henningson]{rowley_09}
{Clarence W.} Rowley, Igor Mezi\`c, Shervin Bagheri, Philipp Schlatter, and
  {Dan S.} Henningson.
\newblock Spectral analysis of nonlinear flows.
\newblock \emph{Journal of Fluid Mechanics}, pages 115--127, 2009.
\newblock \doi{10.1017/S0022112009992059}.

\bibitem[Rudi et~al.(2015)Rudi, Camoriano, and Rosasco]{rudi2016LessMoreNystr}
Alessandro Rudi, Raffaello Camoriano, and Lorenzo Rosasco.
\newblock Less is more: {Nystr{\"o}m} computational regularization.
\newblock In \emph{Proceedings of the 28th International Conference on Neural
  Information Processing Systems}, {{NIPS}}'15, pages 1657--1665, 2015.

\bibitem[Rudi et~al.(2017)Rudi, Carratino, and Rosasco]{rudi_falkon17}
Alessandro Rudi, Luigi Carratino, and Lorenzo Rosasco.
\newblock {FALKON}: An optimal large scale kernel method.
\newblock In I.~Guyon, U.~Von Luxburg, S.~Bengio, H.~Wallach, R.~Fergus,
  S.~Vishwanathan, and R.~Garnett, editors, \emph{Advances in Neural
  Information Processing Systems}, volume~30, 2017.

\bibitem[Rudi et~al.(2018)Rudi, Calandriello, Carratino, and
  Rosasco]{rudi_levscore18}
Alessandro Rudi, Daniele Calandriello, Luigi Carratino, and Lorenzo Rosasco.
\newblock On fast leverage score sampling and optimal learning.
\newblock In \emph{Advances in Neural Information Processing Systems},
  volume~31, 2018.

\bibitem[Schmid(2010)]{dmdSchmid2010}
Peter~J. Schmid.
\newblock Dynamic mode decomposition of numerical and experimental data.
\newblock \emph{Journal of Fluid Mechanics}, 656:\penalty0 5–28, 2010.
\newblock \doi{10.1017/S0022112010001217}.

\bibitem[Schwantes and Pande(2015)]{kerneltica_pande_15}
Christian~R. Schwantes and Vijay~S. Pande.
\newblock Modeling molecular kinetics with {tICA} and the kernel trick.
\newblock \emph{Journal of Chemical Theory and Computation}, 11\penalty0
  (2):\penalty0 600--608, 2015.
\newblock \doi{10.1021/ct5007357}.

\bibitem[Smola and Schölkopf(2000)]{smola_sparse_2000}
Alex~J. Smola and Bernhard Schölkopf.
\newblock Sparse greedy matrix approximation for machine learning.
\newblock In \emph{{ICML} 17}, 2000.

\bibitem[Song et~al.(2009)Song, Huang, Smola, and Fukumizu]{lecme09}
Le~Song, Jonathan Huang, Alex Smola, and Kenji Fukumizu.
\newblock Hilbert space embeddings of conditional distributions with
  applications to dynamical systems.
\newblock In \emph{Proceedings of the 26th Annual International Conference on
  Machine Learning}, page 961–968, 2009.
\newblock \doi{10.1145/1553374.1553497}.

\bibitem[Steinwart and Christmann(2008)]{steinwart2008SupportVectorMachines}
Ingo Steinwart and Andreas Christmann.
\newblock \emph{Support Vector Machines}.
\newblock {Springer Science \& Business Media}, 2008.
\newblock URL \url{https://link.springer.com/book/10.1007/978-0-387-77242-4}.

\bibitem[Steinwart and Scovel(2012)]{steinwart2012MercerTheoremGeneral}
Ingo Steinwart and Clint Scovel.
\newblock Mercer’s theorem on general domains: {{On}} the interaction between
  measures, kernels, and {{RKHSs}}.
\newblock \emph{Constructive Approximation}, 35\penalty0 (3):\penalty0
  363--417, 2012.

\bibitem[Takeishi et~al.(2017)Takeishi, Kawahara, and Yairi]{naoya_deep_17}
Naoya Takeishi, Yoshinobu Kawahara, and Takehisa Yairi.
\newblock Learning koopman invariant subspaces for dynamic mode decomposition.
\newblock In \emph{Advances in {{Neural Information Processing Systems}}}, page
  1130–1140, 2017.

\bibitem[Tu et~al.(2014)Tu, Rowley, Luchtenburg, Brunton, and Kutz]{dmdTu2014}
Jonathan~H. Tu, Clarence~W. Rowley, Dirk~M. Luchtenburg, Steven~L. Brunton, and
  J.~Nathan Kutz.
\newblock On dynamic mode decomposition: Theory and applications.
\newblock \emph{Journal of Computational Dynamics}, 1\penalty0 (2):\penalty0
  391--421, 2014.
\newblock \doi{10.3934/jcd.2014.1.391}.

\bibitem[Wehmeyer and Noé(2018)]{wehmeyer_ticaautoencoder18}
Christoph Wehmeyer and Frank Noé.
\newblock Time-lagged autoencoders: Deep learning of slow collective variables
  for molecular kinetics.
\newblock \emph{The Journal of Chemical Physics}, 148\penalty0 (24), 2018.
\newblock \doi{10.1063/1.5011399}.

\bibitem[Williams and Seeger(2001)]{williams_using_2001}
Christopher K.~I. Williams and Matthias Seeger.
\newblock Using the {Nyström} method to speed up kernel machines.
\newblock In \emph{{NeurIPS} 13}, 2001.

\bibitem[Williams et~al.(2015{\natexlab{a}})Williams, Kevrekidis, and
  Rowley]{edmd}
Mattew~O. Williams, Ioannis~G. Kevrekidis, and Clarence~W. Rowley.
\newblock A data–driven approximation of the {K}oopman operator: Extending
  dynamic mode decomposition.
\newblock \emph{Journal of Nonlinear Science}, 25\penalty0 (6):\penalty0 1307
  -- 1346, 2015{\natexlab{a}}.
\newblock \doi{10.1007/s00332-015-9258-5}.

\bibitem[Williams et~al.(2015{\natexlab{b}})Williams, Rowley, and
  Kevrekidis]{kerneldmd}
Matthew~O. Williams, Clarence~W. Rowley, and Ioannis~G. Kevrekidis.
\newblock A kernel-based method for data-driven {K}oopman spectral analysis.
\newblock \emph{Journal of Computational Dynamics}, 2\penalty0 (2):\penalty0
  247--265, 2015{\natexlab{b}}.
\newblock ISSN 2158-2491.
\newblock \doi{10.3934/jcd.2015005}.

\bibitem[Yang et~al.(2012)Yang, Li, Mahdavi, Jin, and Zhou]{yang12}
Tianbao Yang, Yu-Feng Li, Mehrdad Mahdavi, Rong Jin, and Zhi-Hua Zhou.
\newblock Nystr\"{o}m method vs {R}andom {F}ourier {F}eatures: A theoretical
  and empirical comparison.
\newblock In \emph{Advances in Neural Information Processing Systems 24}, 2012.

\bibitem[Yeung et~al.(2019)Yeung, Kundu, and Hodas]{enoch_deep_19}
Enoch Yeung, Soumya Kundu, and Nathan Hodas.
\newblock Learning deep neural network representations for koopman operators of
  nonlinear dynamical systems.
\newblock In \emph{2019 American Control Conference (ACC)}, 2019.
\newblock \doi{10.23919/ACC.2019.8815339}.

\bibitem[Yurinsky(1995)]{yurinsky1995SumsGaussianVectors}
Vadim Yurinsky.
\newblock \emph{Sums and {G}aussian Vectors}.
\newblock Lecture {{Notes}} in {{Mathematics}} 1617. {Springer-Verlag Berlin
  Heidelberg}, 1 edition, 1995.

\end{thebibliography}

%%%%%%%%%%%%%%%%%%%%%%%%%%%%%%%%%%%%%%%%%%%%%%%%%%%%%%%%%%%%%%%%%%%%%%%
%                              Appendix                               %
%%%%%%%%%%%%%%%%%%%%%%%%%%%%%%%%%%%%%%%%%%%%%%%%%%%%%%%%%%%%%%%%%%%%%%%

\newpage
\appendix

%%%%%%%%%%%%%%%%%%%%%%%%%%%%%%%%%%%%%%%%%%%%%%%%%%%%
\section{Setting and notations}

\subsection{Operators and notations}

We define the following operators:
\begin{itemize}
	\item $\mftX: \ltsp\rightarrow  \rkhs$, defined by $\mftX f = \int_{\dspace} f(x)\fmap(x) \dif\td(x)$ for any $f\in \ltsp$.
	\item $\amftX: \rkhs \rightarrow  \ltsp$, defined by $\amftX h = \iprkhs{h, \fmap{\cdot }}$ for any $h\in \rkhs$ (i.e. the embedding operator mapping a function to its $\td$-equivalence class in $\ltsp$).
	%\item $\mftYcX: \ltsp \rightarrow  \rkhs$, defined by $\mftYcX  f = \int_{\dspace} f(x)\cme{x} \dif\td(x)$ for any $f\in \ltsp$.
	\item $\mftYcX: \ltsp \rightarrow  \rkhs$, defined by 
		$\mftYcX=\mftX\kop^*$.
		%$\mftYcX  f = \int_{\dspace} f(x)\cme{x} \dif\td(x)$ for any $f\in \ltsp$.
	\item $\amftYcX: \rkhs \rightarrow  \ltsp$, defined by
		$\amftYcX=\kop\amftX$.
		%$\amftYcX  h = \iprkhs{h, \cme{\cdot }}$ for any $h\in \rkhs$.
	\item $\Xcov: \rkhs\rightarrow  \rkhs$ defined as $\Xcov = \E_{x\sim \td} \fmap{x}\kron \fmap{x}= \mftX \amftX$, 
		satisfying $\Tr(C) \leq \supfmap^2$. 
		Note that under our assumptions, this also corresponds to the covariance of $Y$.
	\item $\XYcov \de \E_{(x,y)\sim \jd} \fmap{x} \kron \fmap{y} = \mftX\amftYcX $.
\end{itemize}

As well as the following discretized variants:
\begin{itemize}
	\item $\emftX: \bR^n \rightarrow  \rkhs$, defined by $\emftX v = \sum_{i=1}^n v_i  \fmap{x_i }$ for any $v=[v_1 ,\ldots ,v_n]\in \bR^n$
	\item $\aemftX: \rkhs \rightarrow  \bR^n$, defined by $\aemftX h = [ \iprkhs{\fmap(x_1 ),h}, \ldots , \iprkhs{\fmap(x_n),h} ]^T $ for any $h\in \rkhs$
	\item $\emftYcX: \bR^n \rightarrow  \rkhs$, defined by $\emftYcX v = \sum_{i=1}^n v_i  \fmap{y_i }$ for any $v=[v_1 ,\ldots ,v_n]\in \bR^n$.
	\item $\aemftYcX: \rkhs \rightarrow  \bR^n$, defined by $\aemftYcX h = [ \iprkhs{\fmap(y_1 ),h}, \ldots , \iprkhs{\fmap(y_n),h} ]^T $ for any $h\in \rkhs$.
	\item $\eXcov = \tfrac{1}{n} \emftX \aemftX = \frac{1}{n} \sum_{i=1}^n \fmap(x_i ) \kron \fmap(x_i ) \in \cL(\rkhs)$ is the empirical covariance.
	%\item $\eXYcov = \tfrac{1}{n} \sum_{i=1}^n \fmap(x)\fmap(y)^*$
		%is the empirical cross-covariance.
\end{itemize}

The \nystrom{} discretized operators are obtained by applying the kernel map to $m \ll n$ inducing points $\cb{\ldm{j}}_{j=1}^{m}\subset \{\ptx_j \}_{j=1}^n$ and $\cb{\ldmY{j}}_{j=1}^{m}\subset \{\pty_j \}_{j=1}^n$:
\begin{itemize}
    \item $\nmftX:\bR^m\to\rkhs$ such that $\nmftX w = \sum_{j=1}^m w_j \fmap{\ldm{j}}$.
    \item $\nmftY:\bR^m\to\rkhs$ such that $\nmftY w = \sum_{j=1}^m w_j \fmap{\ldmY{j}}$.
\end{itemize}
Furthermore denote by $\px$ and $\py$ the orthogonal projections onto $\spa \nmftX$ and $\spa \nmftY$ respectively.

One important quantity to derive the rates is the so-called effective dimension, defined as 
\begin{align*}
	\deff 
	&\de \Tr(\rXcov^{-1}\Xcov).
\end{align*}
where $\rXcov\de\Xcov + \lambda I$.
\color{black}

\subsection{Conditional mean embedding}

For any $x\in\dspace$, we denote $\cme{x}$ the conditional mean embedding associated to the transition kernel defined as
\begin{align*}
	\cme{x} 
	&\de \E\brk*{\fmap(X_{t+1})| X_t=x}
	 = \int \fmap(y) p(x,\dif y) 
\end{align*}

The following lemma provides a characterization of $\amftYcX$ in terms of the conditional mean embedding.

\begin{tlemma}{}{props_ops}
	We have the following relations:
	\begin{align}
		\mftYcX  f 
			&= \int_{\dspace} f(x)\cme{x} \dif\td(x), \quad f\in \ltsp \\
		(\amftYcX f)(x)
			&= \ip{f, \cme(x)}, 
		\quad f \in \rkhs
		 \label{e:cme_mftYcX}\\
		\mftYcX \amftYcX
			&= \E_{x\sim \td} \cme{x} \kron \cme{x}
		\label{e:cov_cme} 
	\end{align}
\end{tlemma}
\begin{tproofof*}{r:props_ops}{}
	For the first property:
	\begin{align}
		(\amftYcX f)(x)
		&= (\kop (\amftX f))(x)\\
		&= \int (\amftX f)(y) p(x,\dif y)\\
		&= \int f(y) p(x,\dif y)\\
		&= \ip{f, \int \fmap(y) p(x,\dif y)} 
		 = \ip{f, \cme(x)}
	\end{align}
	where we used that $f$ and $\amftX f$ coincide $\td$-almost everywhere.
	The second property is a direct consequence of the definition of the adjoint.
	For \eqref{e:cov_cme}, we simply use \eqref{e:cme_mftYcX} and the definition of $\mftYcX$ to get
	\begin{align*}
		\mftYcX(\amftYcX f) 
		&= \int \ip{f, \cme(z)} \cme{z} \dif \td (z) 
		 = \prt*{\int \cme{z} \cme{z}^* \dif \td (z)} f.
	\end{align*}
\end{tproofof*}

\subsection{Power spaces}
\label{s:interpolating_spaces}

We now define the $\alpha$-power space $\is$ in order to provide some intuition regarding \Cref{a:embedding_property}.

By \Cref{a:bounded_fmap}, $\Tr(\Xcov)=\int \Tr(\fmap(x)\kron \fmap(x))\dif \td(x)\leq \supfmap^2 $ and thus $\Xcov$ is trace-class (and compact).
By \cite{fischer2020SobolevNormLearning}, there exists a non-increasing summable sequence $(\mu _i)_{i\in I}$ for an at most countable index set $I$,
a family $(e_i)_{i\in I}\in \rkhs$ s.t.
$(\amftX e_i)_{i\in I}$ is an orthonormal basis of $\overline{\spa \amftX}\subseteq \lt$ 
and $(\mu _i^{1/2}e_i)_{i\in I}$ is an orthonormal basis of $(\ker \amftX)^\perp\subseteq \rkhs$ such that
\begin{align*}
	%\itop &= \sum_{i\in I} \mu _i \ipltsp{\cdot ,\amftX e_i} \amftX e_i  \\
	\Xcov &= \sum_{i\in I} \mu _i \iprkhs{\cdot ,\mu _i^{1/2} e_i} \mu _i^{1/2} e_i.
\end{align*}

%\todo{Here $\mu _i=\sigma _i^2 $ if $\sigma _i$ corresponds to Vlad's notations}
%\todo{Use steinwart/scover 2012 lemma 2.12}
%Note in particular that the eigenfunctions of $\itop$ and $\Xcov$ coincide up to scaling on the support of $\td$.
For $\alpha\geq  0$, we now define the $\alpha$-power space as 
%\todo{so we need $\sum \mu _i^\alpha  e_i^2 (x)<\infty $ for $x\in X$?}
\begin{align*}
	\is[\alpha ] \de \Set{\sum_{i\in I} a_i \mu _i^{\alpha /2} \amftX e_i| (a_i)_{i\in I}\in \ell_2 (I)}\subseteq \lt
\end{align*}
equipped with norm
\begin{align*}
	\norm*{\sum_{i\in I} a_i \mu _i^{\alpha /2} \amftX e_i}_{\is[\alpha ]} &\de \norm{(a_i)_{i\in I}}_{\ell_2 (I)}.
\end{align*}

We can now make the following assumption regarding the embedding of the power spaces into $\linf$.
\begin{tassumption}{Embedding}{embedding_property_strong}
	There exists $\tau \in [\beta ,1]$ such that
		$c_\tau 
		\de \n{\is[\tau ] \hookrightarrow \linf}^2 
		<\infty$.
\end{tassumption}
We stress that \Cref{a:embedding_property_strong} implies in particular \Cref{a:embedding_property}, and is a common assumption in the literature, see for instance \cite{fischer2020SobolevNormLearning}.

%%%%%%%%%%%%%%%%%%%%%%%%%%%%%%%%%%%%%
\section{Expression of the risk}
\label{s:risk_decomposition}

We have the following risk decomposition.
\begin{tlemma}{}{risk_as_regression}
	The risk can alternatively be written
	\begin{align*}
		\pR(A)
		&= \E_{(x,y)\sim \jd } \nrkhs{\fmap(y) - A \fmap(x)}^2  \\
		&= \cR_{\HS, 0} + \cE_\HS(A)\\
		\text{where}\quad
		\cR_{\HS, 0} 
		&\de \nhs{\mftX}^2 -\nhs{\mftYcX }^2  \\
		&= \int \nrkhs{\cme{x}-\fmap{y}}^2  \dif\rho (x,y) \\
		\text{and}\quad
		\ER(A) 
		&\de \nhs{\mftYcX  - A\mftX}^2  \\
		%&\de \nhs{\kop \amftX - \amftX A^*}^2  \\
		&= \int \nrkhs{\cme{x} - A\fmap{x}}^2  \dif \pi (x).
	\end{align*}
	where $\inf_{A\in \hsH} \ER(A)=0$, and thus we interpret $\ER$ as the excess risk.
\end{tlemma}
\begin{tproofof*}{r:risk_as_regression}{}
Let $(h_i)_{i\in \bN}$ be an orthonormal basis of $\rkhs$. Then
\begin{align*}
	\cE_\HS(A) 
	&\de \nhs{\mftYcX  - A\mftX}^2  \\
	&= \sum_{i\in \bN} \nlt{\amftYcX  h_i - \amftX A^* h_i}^2  \\
	&= \sum_{i\in \bN} \int ((\amftYcX  h_i)(x) - \iprkhs{ A^* h_i, \fmap{x}} )^2  \dif \pi (x) \\
\text{(by \eqref{e:cme_mftYcX})}\quad
	&= \sum_{i\in \bN} \int (\iprkhs{h_i, \cme{x}} - \iprkhs{ h_i, A\fmap{x}} )^2  \dif \pi (x) \\
	&= \int \nrkhs{\cme{x} - A\fmap{x}}^2  \dif \pi (x).
\end{align*}
It holds
\begin{align*}
	\cR_{\HS, 0}
	&= \int \nrkhs{\cme{x}-\fmap{y}}^2 
		\dif \rho (x,y) \\
	&= \int 
		\prt*{\nrkhs{\cme{x}}^2 
		-2\iprkhs*{\cme{x},\fmap{y}}
		+\Tr{\fmap{y}\fmap{y}^*}}
		\dif \rho (x,y) \\
	&= \int 
		\nrkhs{\cme{x}}^2 
	%\Tr{\cme{x}\cme{x}^*}
		\dif \td(x)
		-2\int\iprkhs*{\cme{x},\int\fmap{y} p(x,\dif y)}\dif\td(x)
		+\int\int\Tr{\fmap{y}\fmap{y}^*} p(x,\dif y)\dif\td(x)
		 \\
	&\stackrel{(i)}{=} -\int\nrkhs{\cme{x}}^2 \dif\td(x)
		+\int\Tr\prt*{\fmap{y}\fmap{y}^*}\dif\td(y)
		 \\
	&= -\Tr\prt*{\int\cme{x}\cme{x}^*\dif\td(x)}
		+\Tr\prt*{ \Xcov } \\
	&= -\Tr\prt*{\mftYcX\amftYcX}
		+\Tr\prt*{\mftX\amftX}
\end{align*}
where we used the invariance property of $\td$ in $(i)$ and \Cref{r:props_ops} for the last inequality.
Then one can easily check that the sum of both corresponds to the full risk defined in \eqref{e:population_risk}:
	\begin{align*}
	\cR_{\HS, 0} + \cE_\HS(A)
		&= \int \nrkhs{\cme{x}-\fmap{y}}^2  \dif\jd (x,y) 
		   + \int \nrkhs{\cme{x} - A\fmap{x}}^2  \dif \pi (x) \\
		&= \int\prt*{ \nrkhs{\cme{x}}^2 
				-2\ip{\cme{x},\int\fmap{y}p(x,\dif y)}
				+\int\nrkhs{\fmap{y}}^2 p(x,\dif y) 
			}\dif \td(x) \\
		 &\quad + \int\prt*{ \nrkhs{\cme{x}}^2 -2\ip{\cme{x},A\fmap{x}}+\n{ A\fmap{x}}^2 } \dif \pi (x)\\
		&= \int \prt*{ \int \nrkhs{\fmap(y)}^2 p(x,\dif y) -2\ip{\int \fmap(y) p(x,\dif y),  A \fmap(x)}+\n{A \fmap(x)}^2 }\dif\td(x) \\
		&= \int \prt*{ \nrkhs{\fmap(y)}^2 -2\ip{\fmap(y),  A \fmap(x)}+\n{ A \fmap(x)}^2 } \dif \jd(x,y) \\
		&= \int \nrkhs{\fmap(y) - A \fmap(x)}^2 \dif \jd(x,y) 
		= \pR(A).
	\end{align*}
\end{tproofof*}
%%%%%%%%%%%%%%%%%%%%%%%%%%%%%%%%%%%%%%%%%%%%%%%%%%%%
\section{Expression of the estimators}\label{s:estimators_app}
In this section we give proofs of \cref{r:nkrrest,r:npcrest,r:nrrrest} on how to efficiently compute the \nystrom{} estimators.

For all three -- KRR, PCR and RRR -- estimators, the starting point is their respective \emph{full} estimator which can be derived by following the first-order optimality criterion for the following minimization problems
\begin{align}
    \text{\textbf{Full KRR: }} \qquad &\fkrrest = \argmin_{A\in\rkhs\to\rkhs} \nhs{\emftYcX - A \emftX}^2 + \lambda \nhs{A}^2 \label{e:fkrr_app} \\
    \text{\textbf{Full PCR: }} \qquad &\fpcrest = \argmin_{A\in\rkhs\to\rkhs} \nhs{\emftYcX - A \Pi_r \emftX}^2 \label{e:fpcr_app} \\
    \text{\textbf{Full RRR: }} \qquad &\frrrest = \argmin_{A\in\rkhs\to\rkhs:\rk(A)\leq r} \nhs{\emftYcX - A \emftX}^2 + \lambda \nhs{A}^2 \label{e:frrr_app}
\end{align}
where $\Pi_r$ is the orthogonal projection onto the top-r eigenvectors of $\eXcov$.

To derive the \nystrom{} estimators, we project the embedded data $\emftX$, $\emftYcX$ onto the span of the embedded inducing points -- $\px\emftX$, $\py\emftYcX$ -- and then express the resulting estimators as $\nmftY W \anmftX$ with $W\in\bR^{m\times m}$. This form is particularly useful for later computing forecasts, eigenfunctions and Koopman modes with the estimator.
In particular the following equalities for the projection (shown here for $\px$ but equivalently exist for $\py$)
\begin{equation*}
    \px = \px \px = \nmftX (\anmftX\nmftX)^\dagger \anmftX = \ainmftX \anmftX = \nmftX \inmftX,
\end{equation*}
and the characterization of $\px$ through the SVD of $\nmftX = U\Sigma V^*$, such that $\px = UU^*$.

\subsection{\nystrom{} KRR}
We begin with the \nystrom{} KRR estimator, providing an alternative but equivalent description in \cref{r:nystrom_estimator_onb}.
\begin{tlemma}{Expression of the KRR regularization}{nystrom_estimator_onb}
	Let $U$ be such that $\px=UU^*$, $U^*U=I$. 
	Then it holds 
	\begin{align}
		\regikrr
		\de P_X (\px\eXcov \px + \lambda I)^{-1}
		&= U (U^* \eXcov U + \lambda  I)^{-1} U^*.
		\label{e:nystrom_estimator_onb}
	\end{align}
\end{tlemma}
\begin{tproofof*}{r:nystrom_estimator_onb}{}
Using $U^*U=I$, it holds $(U^*\eXcov U+\lambda I)U^* = U^*(UU^*\eXcov UU^*+\lambda UU^*)$ and thus $U^*(UU^*\eXcov UU^*+\lambda UU^*)^{-1} = (U^*\eXcov U+\lambda I)^{-1}U^*$.
As a consequence, 
\begin{align*}
	\regikrr
	&= \px (\px \eXcov \px + \lambda  I)^{-1} \\
	&= UU^* (UU^* \eXcov UU^* + \lambda  I)^{-1}  \\
	&= U (U^* \eXcov U + \lambda  I)^{-1} U^*.
\end{align*}
\end{tproofof*}

Then we can provide the computatable formulas for \nystrom{} KRR
\begin{tprop}{\nystrom{} KRR}{nkrrest_app}
	The \nystrom{} KRR estimator, obtained by projection of \cref{e:fkrr_app} is
    \begin{align*}
        \nysest &= \py \eYXcov \px (\px \eXcov \px + \lambda I)^{-1} \\
                &= \nmftY \kmmy^\dagger \kmny\knmx (\kmnx\knmx + n\lambda\kmmx)^\dagger\anmftX.
    \end{align*}
\end{tprop}
\begin{tproofof*}{r:nkrrest_app}{}
    Using the definition in \cref{e:nystrom_estimator_onb}, and \cref{r:nystrom_estimator_onb}, we have
    \begin{align*}
    	\nysest 
    	&= \py  \eYXcov \regikrr  \\
    	&= \py  \eYXcov U (U^* \eXcov U + \lambda I)^{-1} U^*  \\
    	&= \py  \eYXcov U \Sigma V^*V\Sigma^{-1} (U^* \eXcov U + \lambda I)^{-1} \Sigma^{-1}V^*V\Sigma U^*
    \end{align*}
    Now using the fact that $\Sigma,V,V^*$ and $U^* \eXcov U + \lambda I$ are full-rank, it holds~\cite[eq. (20)]{adiben-israel2003GeneralizedInversesTheory}
    \begin{align*}
    	\nysest 
    	&= \py  \eYXcov \nmftX (V^*)^\dagger (\Sigma U^* \eXcov U\Sigma + \lambda \Sigma^2)^\dagger V^\dagger\anmftX  \\
    	&= \py  \eYXcov \nmftX (V\Sigma U^* \eXcov U\Sigma V^*  + \lambda V\Sigma^2 V^* )^\dagger\anmftX.
    \end{align*}
    Finally, by definition of $\py$, $\eYXcov$ and $\eXcov$,
    \begin{align*}
    	\nysest 
    	&= \nmftY (\anmftY \nmftY)^\dagger \nmftY^* \emftYcX\aemftX \nmftX (\anmftX \emftX\aemftX \nmftX  + n\lambda \anmftX\nmftX )^\dagger \anmftX  \\
    	&= \nmftY \kmmy^\dagger \kmny \knmx (\knmx\knmx + n\lambda \kmmx )^\dagger \anmftX.
    \end{align*}
\end{tproofof*}

\begin{tremark}{Alternative derivation of the Nyström KRR estimator}{}
    Note that the \nystrom{} KRR estimator can equivalently be derived as the solution to a variational problem similar to \cref{e:fkrr_app}, where the operator $A$ is restricted to operate between spaces $\rkhs_{\widetilde{X}} \de \spa{\nmftX}$ and $\rkhs_{\widetilde{Y}} \de \spa{\nmftY}$.
\end{tremark}

\subsection{\nystrom{} PCR}
Define the following filter on the spectrum of $\px\eXcov\px$: $\regipcr = \irange{\px\eXcov\px}_r^\dagger$, which truncates it to the first $r$ components before taking the pseudo-inverse.
The \nystrom{} PCR estimator, obtained by projection of \cref{e:fpcr_app} is
\begin{equation}\label{e:nys_pcr_app}
    \pcrest = \py\eYXcov \regipcr.
\end{equation}

The next proposition provides an efficiently implementable version of the PCR estimator.
\begin{tprop}{\nystrom{} PCR}{npcrest_app}
    The sketched PCR estimator \cref{e:nys_pcr_app} satisfies
    \begin{equation}
        \pcrest = \nmftY \kmmy^\dagger \kmny \knmx \irange{\kmmx^\dagger \kmnx \knmx}_r \anmftX
    \end{equation}
\end{tprop}
\begin{tproofof*}{r:npcrest_app}{}
    We begin by computing the decomposition of $\px\eXcov\px$ which is necessary to obtain $\regipcr$. The following expressions are equivalent~\cite[Proposition 3]{mollenhauer_svd20} for determining its eigenvectors $\tilde{h}$ and eigenvalues $\lambda$:
	\begin{align*}
		UU^* \eXcov UU^* \tilde{h} &= \lambda \tilde{h} \\
		U^* \eXcov U h &= \lambda h, \qquad \tilde{h} = U h.
	\end{align*}
	Let the truncated eigenvalues be $\Lambda_r = \mathrm{diag}\brk*{\lambda_1, \dots, \lambda_r}$ and the eigenvectors be $H_r = [h_1, \dots, h_r]$. Then $\tilde{H}_r = UH_r$ must be normalized such that $\tilde{H}_r^* \tilde{H}_r = H_r^* U^* U H_r = I$.
    The rank-r truncation $\irange{\px\eXcov\px}_r$ is a projection onto $\tilde{H}_r\tilde{H}_r^*$:
	\begin{align*}
		\irange{\px  \eXcov \px }_r^\dagger = (UU^* \eXcov UU^* (UH_r) (UH_r)^*)^\dagger &= (UH\Lambda H^* H_r H_r^* U^*)^\dagger = U H_r \Lambda_r^{-1} H_r^* U^*
	\end{align*}
	where we used that $U^*\eXcov U = H\Lambda H^*$.
 
	Now substitute $U = \nmftX V \Sigma^{-1}$ to simplify the eigendecomposition of $U^* \eXcov U$:
	\begin{align}
		\Sigma^{-1} V^* \kmnx \knmx V \Sigma^{-1} h &= \lambda h \nonumber \\
		V\Sigma^{-2}V^* \kmnx\knmx d &= \lambda d, \qquad h = \Sigma^{-1} V^* \kmnx\knmx d. \label{e:pcr-edec-2}
	\end{align}
	where $V\Sigma^{-2}V^* = \kmmx^\dagger$. Denote by $D_r = [d_1, \dots, d_r]$ the truncated eigenvectors such that $H_r = \Sigma^{-1} V^* \kmnx\knmx D_r$, normalized such that $H^* H = D^* \kmnx\knmx\kmmx^\dagger\kmnx\knmx D = I$,
	\begin{align*}
		U H_r \Lambda_r^{-1} H_r^* U^* &= \nmftX \kmmx^\dagger \kmnx \knmx D_r \Lambda_r^{-1} D_r^* \kmnx\knmx \kmmx^\dagger \anmftX \\
		&= \nmftX D_r \Lambda_r D_r^* \anmftX \\
		&= \nmftX \irange{\kmmx^\dagger \kmnx \knmx}_r \anmftX.
	\end{align*}	
	Finally, we can plug the pieces together to get
	\begin{align*}
		\py  \eYXcov \irange{\px  \eXcov \px }_r^\dagger = \nmftY \kmmy^\dagger \kmny \knmx \irange{\kmmx^\dagger \kmnx \knmx}_r \anmftX.
	\end{align*}
\end{tproofof*}

\begin{tremark}{Variational problem for \nystrom{} PCR}{}
    Note that, unlike the NysKRR estimator, the variational problem for NysPCR where the operator is restricted to $A: \rkhs_{\widetilde{X}}\to\rkhs_{\widetilde{Y}}$ is not equivalent to the one obtained in \cref{r:npcrest_app} by projecting the covariance operator. In fact, the former does not take the full covariance into account when computing the low-rank projection, but just the \nystrom{} points.
\end{tremark}

\subsection{\nystrom{} RRR}
The \nystrom{} RRR estimator does not correspond to a specific spectral filter. We can nonetheless compute it starting from the expression of the exact empirical estimator~\cite{kostic2022learning}, projecting the covariance operators, and rearranging the expression to result in a finite-dimensional procedure.

\begin{tprop}{\nystrom{} RRR}{nrrrest_app}
    The sketched RRR estimator can be written as
    \begin{equation}\label{eq:RRR}
        \rrrest = \irange{ \py\eYXcov\px (\px\eXcov\px + \lambda I)^{-1/2}}_r (\px\eXcov\px + \lambda I)^{-1/2} .
    \end{equation}
    To compute it, solve the $m\times m$ eigenvalue problem
    \begin{equation*}
        (\kmnx\knmx + n\lambda\kmmx)^\dagger \kmnx\knmy\kmmy^\dagger\kmny\knmx w_i = \sigma_i^2 w_i
    \end{equation*}
    for the first $r$ eigenvectors $W_r = [w_1, \dots, w_r]$, normalized such that
    \(
        W_r^* \kmnx\knmy\kmmy^\dagger\kmny\knmx W_r = I.
    \)
    Then let $D_r\de\kmmy^\dagger \kmny\knmx W_r$ and $E_r\de(\kmnx\knmx + n\lambda\kmmx)^\dagger\kmnx\knmy U_r$, such that the following holds
    \begin{equation}
        \rrrest = \nmftY D_r E_r^* \anmftX.
    \end{equation}
\end{tprop}
\begin{tproofof*}{r:nrrrest_app}{}
    Let $B := n^{1/2} \py  \eYXcov \px  (\px  \eXcov \px  + \lambda I)^{-1/2}$. 
    The computationally intensive part for this estimator is in evaluating the rank-r truncation $\irange{B}_r$. Its singular values and left singular vectors can be obtained by solving the symmetric eigenvalue problem $BB^* q_i = \sigma_i^2 q_i$. We rewrite $BB^*$
    \begin{align*}
        BB^* &= \py \emftYcX\aemftX \px  (\px  \emftX\aemftX \px + n\lambda I)^{-1} \px \emftX\aemftYcX \py \\ 
        &= \py \emftYcX \aemftX \px \emftX (\aemftX \px \emftX + n\lambda I)^{-1} \aemftYcX \py \\
        &= \py \emftYcX \knmx\kmmx^\dagger\kmnx (\knmx\kmmx^\dagger\kmnx + n\lambda I)^{-1}\aemftYcX \py \\
        &= \py \emftYcX \knmx\kmmx^\dagger (\kmnx\knmx\kmmx + n\lambda I)^{-1}\kmnx\aemftYcX\py \\
        &= \py \emftYcX \knmx (\kmnx\knmx + n\lambda \kmmx)^\dagger \kmnx\aemftYcX \py
    \end{align*}
    where the second and fourth equalities are applications of the push-through identity, the third by definition of projections and kernel matrices, and the last by collecting $\kmmx$. By construction, the non-trivial eigenfunctions of $BB^*$ are in the range of $\py\emftYcX\knmx$, therefore we can set $q_i = \py\emftYcX\knmx w_i$ for some $w_i\in\bR^{m}$, and solve the following eigenvalue problem instead
    \begin{align*}
    	\py  \emftYcX \knmx (\kmnx\knmx + n\lambda\kmmx)^{\dagger} \kmnx \aemftYcX \py  \emftYcX \knmx w_i &= \sigma_i^2 \py  \emftYcX \knmx w_i \\
    	(\kmnx\knmx + n\lambda\kmmx)^{\dagger} \kmnx \knmy\kmmy^\dagger\kmny \knmx w_i &= \sigma_i^2 w_i
    \end{align*}
    where we have simplified the left term of both sides of the equation.
    
    The eigenfunctions of $BB^*$ are therefore $q_{i} = \py \emftYcX \knmx w_{i}$, which must be normalized as
    \begin{align*}
    	\nrkhs{q_{i}}^2 & = w_{i}^{\top} \kmnx \knmy \kmmy^\dagger \kmny \knmx w_{i} = 1.
    \end{align*}
    Thanks to this normalization, the projector onto the $r$ leading left singular vectors of $B$ is $Q_r Q_r^*$, where $Q_r = \brk*{q_1, \dots, q_r}$. Then the NysRRR estimator can be written as 
    \begin{align*}
    	Q_r Q_r^* B (\px  \eXcov \px  + \lambda I)^{-1/2}
    \end{align*}
    where
    \begin{align*}
    	B (\px \eXcov \px + \lambda I)^{-1/2} &= \py \eYXcov \px  (\px  \eXcov \px + \lambda I)^{-1} \\
    	&= \py \emftYcX \knmx (\kmnx\knmx + n\lambda\kmmx)^{-1} \anmftX.
    \end{align*}
    with the same techniques we used for rewriting $BB^*$.
    Finally, let $D_r$ and $E_r$ as in the statement. We can apply the projection to obtain
    \begin{align*}
    	Q_r Q_r^* B (\px  \emftX\aemftX \px  + n \lambda I)^{-1/2} = \nmftY D_r E_r^* \anmftX.
    \end{align*}
\end{tproofof*}

%%%%%%%%%%%%%%%%%%%%%%%%%%%%%%%%%%%%%%%%%%%%%%%%%%%%
\section{Forecasting \& Koopman Modes}\label{s:kmd_app}
The three estimators considered in \Cref{s:estimators_app} are all of the form
\begin{equation*}
    \eest = \nmftY W \anmftX, \qquad W\in\bR^{m\times m}.
\end{equation*}
We will use this generic form to provide expressions for the following operations:
\begin{enumerate}
    \item producing forecasts of the dynamical system at a future time,
    \item computing the approximate eigenvalues and eigenfunctions of the Koopman operator,
    \item computing the Koopman modes.
\end{enumerate}

\subsection{Forecasting}
Given a new data-point $x\in\dspace$ and an observable function $g\in\rkhs$ (note that this can simply be the identity function), we can approximate the one-step-ahead expectation $\E\brk*{g(X_{t+1}) | X_t = \ptx} = (\kop g)(\ptx)$ by using the obtained estimators $\eest[]^*$. 
Note that by the reproducing property $\anmftY g = [g(\pty_i), \dots, g(\pty_m)]^\top \der g_m$, then
\begin{equation*}
	(\eest[]^* g)(x) = (\nmftX W^{\top} \anmftY g)(\ptx) = (\nmftX W^{\top} g_m)(\ptx) = \sum_{i=1}^m (W^\top g_m)_i k(\ldm_i, \ptx).
\end{equation*}

\subsection{Eigenfunctions and eigenvalues}

We wish to compute the eigenfunctions $\xi, \psi \in \rkhs$, as well as the eigenvalues $\lambda_i$ of $\hat{A}$. 
The left eigenfunctions satisfy $\eest[]^*\xi_i = \bar{\lambda_i} \xi_i$ and the right eigenfunctions satisfy $\eest[]\psi_i = \lambda_i \psi_i$.
In the following we will use \citet[Proposition 3]{mollenhauer_svd20} to manipulate the eigendecomposition of operators in $\rkhs$.
% \begin{tprop}{}{edec}
% 	Given operator $S: \rkhs\to\rkhs$ with $S = \Upsilon B \Phi^*$ and $B\in\bR^{m\times m}$, then
% 	\begin{enumerate}
% 		\item if $\lambda$ is an eigenvalue of $B\Phi^*\Upsilon\in\bR^{m\times m}$ with corresponding eigenvector $w\in\bR^{m}$, then $\Upsilon w \in \rkhs$ is an eigenfunction of $S$ corresponding to $\lambda$.
% 		\item If $\lambda \ne 0$ is an eigenvalue of $S$ corresponding two $v\in\rkhs$, then $B\Phi^* v\in\bR^m$ is an eigenvector of $B\Phi^*\Upsilon\in\bR^{m\times m}$ corresponding to the eigenvalue $\lambda$.
% 	\end{enumerate}
% \end{tprop}

Consider the decomposition $W = U_r V_r^*$ with $U_r, V_r \in \bC^{m\times r}$, which is available for all considered estimators with $r \le m$. For example, in the \nystrom{} RRR estimator of \cref{r:nrrrest_app}, we can simply take $U_r = D_r$ and $V_r = E_r$. For the \nystrom{} KRR estimator instead, $r = m$ and we can take the whole of $W$ as our $U_r$ and $V_r = I$.

To compute the \textbf{right eigenfunctions} $\psi_i$, such that $(\nmftY U_r V_r^* \anmftX)\psi_i = \lambda_i \psi_i$, 
consider the following equivalent eigendecomposition
\begin{equation*}
    V_r^* \anmftX\nmftY U_r \tilde{g}_i = \lambda_i \tilde{g}_i, \quad \text{where}~\psi_i = \nmftY U_r \tilde{g}_i.
\end{equation*}
Note that $\anmftX\nmftY = \kmmxy$ is a finite-dimensional object which can easily be computed.
The eigenfunctions $\psi_i$ must be normalized such that $\psi_i^*\psi_i = 1$ for every $i$, so we must have
\begin{equation*}
    \tilde{g}_i^* U_r^* \anmftY \nmftY U_r \tilde{g}_i = 1.
\end{equation*}

A very similar process can be followed to obtain the \textbf{left eigenfunctions} $\xi_i$, such that $\nmftX V_r U_r^* \anmftY \xi_i = \bar{\lambda}_i \xi_i$. Here we consider instead
\begin{equation*}
    U_r^* \anmftY \nmftX V_r \tilde{h}_i = \bar{\lambda}_i \tilde{h}_i, \quad \text{where}~\xi_i = \nmftX V_r \tilde{h}_i.
\end{equation*}
where once again, $\anmftY\nmftX = \kmmxy^\top$ and the eigenfunctions must be normalized such that $\tilde{h}_i^* V_r^* \anmftX\nmftX V_r \tilde{h}_i = 1$ for every $i$.
Finally, $\psi$ and $\xi$ must be orthogonal to each other: we must have for $i, j \in [r]$ that $\langle\psi_i, \bar{\xi}_j\rangle_\rkhs = \delta_{ij}$ (where $\delta_{ij}$ is a Dirac delta equals to 1 when $i = j$ and 0 otherwise).
We can compute
\begin{equation*}
    \langle\psi_i, \bar{\xi}_j\rangle_\rkhs = \tilde{h}_i^* V_r^* \kmmxy U_r \tilde{g}_i = \lambda_j \tilde{h}_i^* \tilde{g}_j,
\end{equation*}
and note that $\tilde{h}_i^* \tilde{g}_j = \delta_{ij}$, but we must normalize $\xi$ such that 
\begin{equation*}
    \xi_i = \nmftX V_r \tilde{h}_i / \bar{\lambda}_i.
\end{equation*}

\subsection{Koopman modes}
Given the eigendecomposition of any estimator $\eest[]$ as $\eest[r] = \sum_{i=1}^r \lambda_i \psi_i \otimes \bar{\xi}_i$, for an observable $g$ we have the following
\begin{equation*}
	\eest[r]^* g = \sum_{i=1}^r \lambda_i \xi_i \langle g, \bar{\psi}_i \rangle_{\rkhs}
\end{equation*}
where $\langle g, \bar{\psi}_i \rangle_{\rkhs} = \gamma_i^g$ are the Koopman modes. Expanding the definition of $\psi_i$ we get
\begin{equation*}
	\gamma_i^g = \langle g, \bar{\psi}_i \rangle_{\rkhs} = \tilde{g}_i^* U_r^* \anmftY g = \tilde{g}_i^* U_r^* g_m \in \bC^m
\end{equation*}
which we can efficiently compute.

%%%%%%%%%%%%%%%%%%%%%%%%%%%%%%%%%%%%%%%%%%%%%%%%%%%%
\section{Excess risk of the Nyström KRR estimator}

\subsection{Almost-sure decomposition of the KRR excess risk}

\begin{tlemma}{Excess risk decomposition in operator norm for KRR}{bound_ER_det_KRR_a1}
	Let \Cref{a:universal_kernel,a:bounded_fmap,a:regularity,a:embedding_property} hold.
	Then the Nyström KRR estimator \eqref{e:nysest} satisfies almost surely
	\begin{align*}
		\cE(\nysest)^{1/2}
		&\leq  a\lambda ^{1 /2} 
			 + a \theta _1^2 \noprkhs{(\reXcov - \rXcov)\rXcov^{-1/2}} 
			 + \chalf^2  \noprkhs{(\YXcov-\eYXcov)\rXcov^{-1/2}}\\
			&\quad + a\chalf \chalfsw\cone \noprkhs{\px^\perp\rXcov^{1/2}}
			 + \chalf^2  \noprkhs{\py^\perp \rXcov^{1/2}}
	\end{align*}
	where 
	$\chalf\de\n{\reXcov^{-1/2}\rXcov^{1/2}}$,  
	$\chalfsw\de\n{\reXcov^{1/2}\rXcov^{-1/2}}$, 
	$\cone\de\n{\reXcov^{-1}\rXcov}$, and $a$ is the constant of \Cref{a:regularity}.
\end{tlemma}
\begin{tproofof*}{r:bound_ER_det_KRR_a1}{}
	Let 
	$\chalf\de\n{\reXcov^{-1/2}\rXcov^{1/2}}$, 
	$\cone\de\n{\reXcov^{-1}\rXcov}$. As in \Cref{r:nystrom_estimator_onb} define $\regikrr \de U(U^* \eXcov U+\lambda I)^{-1}U^*$.
	We have
	\begin{align*}
		\cE(\nysest)^{1/2}
		&= \n{\mftYcX  - \nysest\mftX}_{\cB(\lt,\rkhs)} \\
		&\leq  
		\n{
			\mftYcX  - \rest\mftX
		}_{\cB(\lt,\rkhs)} + \n{
			(\rest
			- \YXcov \regikrr)\mftX
		}_{\cB(\lt,\rkhs)}\\& \quad+ \n{
			(\YXcov \regikrr
			-\nysest )\mftX 
		}_{\cB(\lt,\rkhs)} \\
		&\leq  
		\underbrace{\n{
			\mftYcX  - \rest\mftX
		}_{\cB(\lt,\rkhs)}}_{A} + \underbrace{\n{
			(\rest - \YXcov \regikrr)\Xcov^{1/2}
		}}_{B} \\ &\quad + \underbrace{\n{
			(\YXcov \regikrr -\nysest)\Xcov^{1/2}
		}}_{C}
		\numberthis\label{e:error_decomposition}
	\end{align*}
	where we used the polar decomposition $\amftX = W \Xcov^{1/2}$ for some partial isometry $W:\rkhs \rightarrow  \ltsp$.
	
	\textbf{The first term} is
	\begin{align*}
		\n{\mftYcX  - \rest\mftX}
		&= \n{\mftX \kop^*  - \YXcov\rXcov^{-1}\mftX}\\
		%&= \n{\mftYcX (I - \amftX(\mftX\amftX+\lambda I)^{-1}\mftX)} \\
		%&= \n{\mftYcX (I - \ritop^{-1}\itop)} \\
		%&= \lambda \n{\mftYcX \ritop^{-1}} \\
		&\leq  a\lambda ^{1/2} + \n{(I-P_{\rkhs}) \mftYcX}
	\end{align*}
	where we used the definition of $\mftYcX$ and applied \Cref{r:bound_rest}.
	%and the second term is zero under \Cref{a:universal_kernel}.

	\bigskip
	\textbf{The second term} of our decomposition \eqref{e:error_decomposition} can be bounded as follows:

It holds
\begin{align*}
B
	&= \n{\YXcov(\rXcov^{-1} - \regikrr)\Xcov^{1/2}} \\
\text{(by \Cref{r:bound_rest_opnorm_A_a1}:)}\quad\quad
	&\leq  a\n{\Xcov(\rXcov^{-1} - \regikrr)\Xcov^{1/2}} \\
	&\leq  a\prt*{
		\underbrace{\n{\Xcov(\rXcov^{-1} - \reXcov^{-1})\Xcov^{1/2}}}_{B_1 }
	  + \underbrace{ \n{\Xcov(\reXcov^{-1} - \regikrr)\Xcov^{1/2}} }_{B_2 } } 
\end{align*}
We now bound the terms $B_1 $ and $B_2 $ separately.
\begin{align*}
B_1  &= 
	\n{\Xcov(\rXcov^{-1} - \reXcov^{-1})\Xcov^{1/2}} \\
	&= \n{\Xcov\rXcov^{-1}(\reXcov - \rXcov)\reXcov^{-1}\Xcov^{1/2}} \\
	&\leq  \n{\Xcov\rXcov^{-1}}\n{(\reXcov - \rXcov)\rXcov^{-1/2}}\n{\rXcov^{1/2}\reXcov^{-1/2}}\n{\reXcov^{-1/2}\Xcov^{1/2}} \\
	&\leq  \theta _1^2 \n{(\reXcov - \rXcov)\rXcov^{-1/2}} 
\end{align*}
Let $\Pl\de \reXcov^{1/2} \regikrr \reXcov^{1/2}$.
We recall that $\regikrr=\px\regikrr$, so that
\begin{align*}
	\Pl^2 
	&= \reXcov^{1/2} (\regikrr \reXcov \px) \regikrr \reXcov^{1/2} \\
	&= \reXcov^{1/2} \px \regikrr \reXcov^{1/2} \\
	&= \Pl.
\end{align*}
This implies $\Pl^2 =\Pl=\Pl^*$. Hence $\Pl$ is an orthogonal projection, and defining $\Pl^\perp=I-\Pl$ it holds $\noprkhs{\Pl^\perp}\leq 1$. We can thus bound $B_2 $ as follows:
\begin{align*}
	B_2  &= 
	\n{\Xcov(\reXcov^{-1} - \regikrr)\Xcov^{1/2}} \\
	&=\n{\Xcov\reXcov^{-1/2}(I-\Pl)\reXcov^{-1/2}\Xcov^{1/2}} \\
\text{(by \Cref{r:rkrrls_onehalf})}\quad\quad
	&=\n{\Xcov\reXcov^{-1}\px^\perp\reXcov^{1/2}(I-\Pl)\reXcov^{-1/2}\Xcov^{1/2}} \\
	&=\noprkhs{\Xcov\reXcov^{-1}}\noprkhs{\px^\perp\rXcov^{1/2}}\noprkhs{\rXcov^{-1/2}\reXcov^{1/2}}\n{I-\Pl}\noprkhs{\reXcov^{-1/2}\Xcov^{1/2}} \\
	&\leq  \chalf \chalfsw \cone \n{\px^\perp\rXcov^{1/2}} 
\end{align*}

	\textbf{For the third term,} due to \Cref{r:nystrom_estimator_onb}:
	\begin{align}
		C
		&= \n{ (\YXcov - \py \eYXcov ) \regikrr\Xcov^{1/2}} \\
		&\leq  \n{ (\YXcov - \py \eYXcov )\rXcov^{-1/2}}\noprkhs{\rXcov^{1/2}\reXcov^{-1/2}}\noprkhs{ \Pl}\noprkhs{ \reXcov^{-1/2}\Xcov^{1/2}} \\
		&\leq  \chalf^2  \noprkhs{ (\YXcov - \py \eYXcov )\rXcov^{-1/2}} \\
		&\leq  \chalf^2  \prt*{
			\noprkhs{\py^\perp \rXcov^{1/2}}
			+ \noprkhs{(\YXcov-\eYXcov)\rXcov^{-1/2}} }
	\end{align}
	where we used \Cref{r:bound_proj_crosscov_onehalf} for the last inequality.

	Starting again from \eqref{e:error_decomposition} and putting everything together, we get
	\begin{align*}
		\cE(\nysest)^{1/2}
		&\leq  a\lambda ^{1 /2} \\
			&\quad + a \theta _1^2 \noprkhs{(\reXcov - \rXcov)\rXcov^{-1/2}} \\
			&\quad + \chalf^2  \noprkhs{(\YXcov-\eYXcov)\rXcov^{-1/2}}\\
			&\quad + a\chalf\chalfsw \cone \noprkhs{\px^\perp\rXcov^{1/2}}\\
			&\quad + \chalf^2  \noprkhs{\py^\perp \rXcov^{1/2}}
			.
	\end{align*}
\end{tproofof*}

\subsection{Excess risk rates for KRR}
\label{s:proof_krr_rates_a1}

In order to control the terms appearing in our decomposition, we recall that \Cref{a:spectral_decay} implies 
\begin{align}
	\deff &\leq  C_\beta  \lambda ^{-\beta} 
	\text{ where }
	C_\beta \de \lcba{\frac{c}{1-\beta }\quad, \beta <1\\ \supfmap^2 \quad,\beta =1 },
	\label{e:decay_deff}
\end{align}
where $c$ is the constant of \Cref{a:spectral_decay}, see \cite[Proposition 3 with $b\rightarrow 1/\beta$ and $\beta \rightarrow c$]{caponnetto07} and \cite[Lemma 11]{fischer2020SobolevNormLearning} which shows that the existence of a constant $C_\beta $ such that the first part of \eqref{e:decay_deff} holds implies in return $\lambda _i(\Xcov)\lesssim i^{-1/\beta }$.

\begin{tproofof*}{r:bound_ER_KRR_a1}{}
	By \Cref{r:bound_ER_det_KRR_a1} taking $\px=\py$, it holds almost surely
	\begin{align*}
		\cE(\nysest)^{1/2}
		&\leq  a\lambda ^{1 /2} 
			 + a \theta _1^2 \noprkhs{(\reXcov - \rXcov)\rXcov^{-1/2}} 
			 + \chalf^2  \noprkhs{(\YXcov-\eYXcov)\rXcov^{-1/2}}\\
			&\quad + a\chalf\chalfsw \cone \noprkhs{\px^\perp\rXcov^{1/2}}
			 + \chalf^2  \noprkhs{\py^\perp \rXcov^{1/2}}
	\end{align*}
	and we recall that
	$\chalf\de\n{\reXcov^{-1/2}\rXcov^{1/2}}$ and 
	$\cone\de\n{\reXcov^{-1}\rXcov}$.
	We bound separately the terms appearing in this expression.

	\paragraph{Bound of $\chalf$ and $\chalfsw$.} 
	We control these 
	term
	%two terms
	by bounding $\n{\rXcov^{-1/2}(\eXcov - \Xcov)\rXcov^{-1/2}}$.
	By \Cref{r:concentration_Xcov_sa_half} it holds
	for any $\delta '\in ]0,1[$ and any $\lambda \in ]0,\noprkhs{\Xcov}]$ with probability $1-\delta '$ 
	\begin{align}
		\n*{ \rXcov^{-1/2}(\eXcov - \Xcov)\rXcov^{-1/2} } \leq  
		\frac{4c_\tau  \beta }{3n\lambda ^{\tau }} + \sqrt{\frac{2c_\tau \beta }{n\lambda ^{\tau }}} 
		\quad\text{ where }\quad
		\beta  = \log\prt*{\tfrac{8 \supfmap^2 }{\delta '\lambda }} 
	\end{align}
	A sufficient condition to bound the right hand side of the previous expression by $1/4$ is to have
	%This is in particular ensured for $A=1/4$
	$n\lambda ^\tau  > 32 c_\tau  \beta $ (in which cases both terms are bounded by $1/8$).
	Assuming this holds, $I-\n{\rXcov^{-1/2}(\eXcov-\Xcov)\rXcov^{-1/2}}$ is invertible and we also have
	\begin{align*}
	\chalfsw^2 
		= \n{\reXcov^{1/2}\rXcov^{-1/2}}^2  
		&= \n{\rXcov^{-1/2}\reXcov\rXcov^{-1/2}}
		 = \n{I-\rXcov^{-1/2}(\Xcov-\eXcov)\rXcov^{-1/2}}\\
		&\leq 1+\n{\rXcov^{-1/2}(\Xcov-\eXcov)\rXcov^{-1/2}} \\
		&\leq 1.25 \\ % sqrt > 1.12
	\text{and thus}\quad
	\chalfsw 
		&\leq 1.12 \\ 
	\text{while}\quad
	\chalf^2 
		=\n{\reXcov^{-1/2}\rXcov^{1/2}}^2 
		&%\de\n{\reXcov^{-1/2}\rXcov^{1/2}}^2  
		 =\n{(\rXcov^{-1/2}\reXcov\rXcov^{-1/2})^{-1}} \\
		&\stackrel{(i)}{\leq }(1-\n{\rXcov^{-1/2}(\eXcov-\Xcov)\rXcov^{-1/2}})^{-1}\\
		&\leq  1.34\\ 	% sqrt -> 1.16
	\text{and thus}\quad
	\chalf
		&\leq  1.16 	
	\end{align*}
	where $(i)$ can be obtained by taking the Neumann expansion of $I-\n{\rXcov^{-1/2}(\eXcov-\Xcov)\rXcov^{-1/2}}$.

	\paragraph{Bound for $\cone$.}
	By \Cref{r:concentration_Xcov_inv_oneside} it holds with probability $1-\delta '$
	\begin{align}
	\nop{(\Xcov-\eXcov)\rXcov^{-1}} 
		&\leq  \frac{2\supfmap \sqrt{c_\tau } \log(\nicefrac{2}{\delta '})}{\lambda ^{(\tau +1)/2} n} + \sqrt{\frac{2 \supfmap^2  \Tr(\rXcov^{-2}\Xcov) \log(\nicefrac{2}{\delta '})}{n}} 
	\end{align}
	Both terms in the above rhs are bounded by $1/4$ provided
	\begin{align*}
	\lambda ^{(\tau +1)/2} n &\geq 
		8 \supfmap \sqrt{c_\tau } \log(\nicefrac{2}{\delta '})\\
	n &\geq 
		32 \supfmap^2  \lambda ^{-(1+\beta )} \log(\nicefrac{2}{\delta '})
	\end{align*}
	where we used $\Tr(\rXcov^{-2}\Xcov)=\sum \lambda_i(\Xcov)(\lambda_i(\Xcov)+\lambda )^{-2}\leq \lambda ^{-1}\Tr(\rXcov^{-1}\Xcov)\leq C_\beta \lambda ^{-(1+\beta )}$.
	When this is the case, we have $\nop{(\Xcov-\eXcov)\rXcov^{-1}}\leq 1/2<1$ and 
	the operator $I-(\reXcov-\rXcov)\rXcov^{-1}$ is invertible. 
	\begin{align*}
	\cone
		%&=\n{\rXcov\reXcov^{-1}}
		&=\n{(\reXcov\rXcov^{-1})^{-1}}
		 =\n{(I-(\reXcov-\rXcov)\rXcov^{-1})^{-1}}\\
		&\stackrel{(i)}{\leq } (1-\noprkhs{(\reXcov-\rXcov)\rXcov^{-1}})^{-1} \\
		&\leq  2.
	\end{align*}
	where $(i)$ can be obtained by considering the Neumann expansion of $I-(\reXcov-\rXcov)\rXcov^{-1}$.

	\paragraph{Bound for $\n{\px^\perp\rXcov^{1/2}}$.}
	By \Cref{r:uniform_nys_approx},
	provided $\lambda \in ]0,\noprkhs{\Xcov}]$ 
	it holds with probability $1-\delta '$
	\[ \noprkhs{\px^\perp \rXcov^{1/2}} \leq  \sqrt{3\lambda } \]
	provided $m\geq \max(67, 5\esssup_{x\sim \td} \n{\rXcov^{-1/2}\fmap{x}}^2 )\log\frac{4\supk}{\lambda \delta '}$, which by \Cref{r:bound_Ninf} is ensured if $m\geq \max(67, 5\frac{c_\tau }{\lambda ^{\tau }})\log\frac{4\supk}{\lambda \delta '}$.

	\paragraph{Bound for  $\n{(\Xcov-\eXcov)\rXcov^{-1/2}}$ and $\n{(\YXcov-\eYXcov)\rXcov^{-1/2}}$.}
	By \Cref{r:concentration_Xcov_onehalf}, 
	for any $\delta '\in ]0,1[$, each of the following events holds with probability $1-2\delta '$:
	\begin{align}
	\max(\n{(\Xcov-\eXcov)\rXcov^{-1/2}}, 
		\n{(\YXcov-\eYXcov)\rXcov^{-1/2}})
		&\leq  \frac{2\supfmap \sqrt{c_\tau } \log(\nicefrac{2}{\delta '})}{\lambda ^{\tau /2} n} + \sqrt{\frac{2 \supfmap^2  \deff \log(\nicefrac{2}{\delta '})}{n}}
	\end{align}
	By \Cref{e:decay_deff} we have $\deff \leq  C_\beta \lambda ^{-\beta }$. 
	
	Choosing $\delta '=\delta /5$, we get via a union bound with probability $1-\delta $ that $\theta _1 \theta _2 \theta _3 \leq 2.6$, $\theta _1 ^2 \leq 1.34$ and
	\begin{align*}
		\cE(\nysest)^{1/2}
		&\leq  a\lambda ^{1/2} 
			 + 1.34(a+1) \prt*{ 
			 	\frac{2\supfmap \sqrt{c_\tau } \log(\nicefrac{2}{\delta '})}{\lambda ^{\tau /2} n} + \sqrt{\frac{2 \supfmap^2  C_\beta  \log(\nicefrac{2}{\delta '})}{n\lambda ^{\beta }}} 
			}  
			+ (2.6a +1.34) \sqrt{3}\lambda ^{1/2}
			\\
		&\leq  c_1  \lambda ^{1/2} 
			 + c_2  \lambda ^{-\tau /2} n^{-1} 
			 + c_3  \lambda ^{-\beta /2} n^{-1/2}
			  \\
	\text{where:} \quad
		c_1  &\de (5.5a+2.33) \\
		c_2  &\de 1.34(a+1)2\supfmap \sqrt{c_\tau } \log(\nicefrac{2}{\delta '}) \\
		c_3  &\de 1.34(a+1)\sqrt{2 \supfmap^2  C_\beta  \log(\nicefrac{2}{\delta '})} 
	\end{align*}
	for any $\lambda $ and $m$ satisfying the constraints
	\begin{equation}
	\lcba{
	\lambda  &> n^{-1/\tau } (32 c_\tau )^{1/\tau } \log\prt*{\tfrac{8 \supfmap^2 }{\delta ' \lambda }}^{1/\tau } \\
	\lambda  &\geq  n^{-2/(\tau +1)}
		(8 \supfmap \sqrt{c_\tau } \log(\nicefrac{2}{\delta '}))^{2/(\tau +1)}\\
	\lambda  &\geq  n^{-1/(1+\beta )}
		 (32 \supfmap^2   \log(\nicefrac{2}{\delta '}))^{1/(1+\beta )} \\
	\lambda  &\in ]0,\supfmap^2 ]. \\
	m &\geq \max(67, 5\frac{c_\tau }{\lambda ^{\tau }})\log\frac{4\supk}{\lambda \delta '}
	\quad\text{(uniform sampling)}
	}
	\label{e:constraints_l_m}
	\end{equation}
	% Here for \alpha <1, it holds
	% 1/(1+\beta ) \leq  1 \leq  1/\tau 
	% 1/(1+\beta ) \leq  1 \leq  2/(\tau +1)  \\
	% So the saturating constraint (and term saturating the rates) is still the one in \beta 
	% > yields the largest \lambda 
	We pick
		$\boxed{ \lambda  \de c_\lambda  n^{-1/(1+\beta )} }$
	which is asymptotically the saturating constraint (given that $1/(1+\beta ) < 1 < 2/(\tau +1) \leq  1/\tau $), where $c_\lambda $ is a constant choosen to enforce the following equations (which are sufficient conditions for \cref{e:constraints_l_m} to hold):
	\begin{equation}
		\lcba{
		c_\lambda ^\tau  n^{1-\tau /(1+\beta )} 
			&> (32 c_\tau ) \log\prt*{\tfrac{8 \supfmap^2  n^{1/(1+\beta )}}{\delta ' c_\lambda }} \\
		c_\lambda ^{(\tau +1)/2} n^{1-(\tau +1)/(2(1+\beta ))} 
			&\geq  8 \supfmap \sqrt{c_\tau } \log(\nicefrac{2}{\delta '})\\
		c_\lambda  &\geq   (32 \supfmap^2   \log(\nicefrac{2}{\delta '}))^{1/(1+\beta )} \\
		c_\lambda  n^{-1/(1+\beta )} &\leq  \supfmap^2  
		}
		\label{e:constraints_l}
	\end{equation}
	As ${1-(\tau +1)/(2(1+\beta ))}>0$, a sufficient condition for the second equation is
	\[ c_\lambda  \geq  (8 \supfmap \sqrt{c_\tau } \log(\nicefrac{2}{\delta '}))^{2/(\tau +1)}.  \]

	%------ (Derivation constant if \lambda  has a log term) --------\\
	%Assuming $c_\tau \geq 8\supfmap^2 $, a sufficient condition to satisfy the first constraint is 
	%\begin{align*}
		%c_\lambda ^\tau  
			%&> (32 c_\tau ) \log\prt*{\tfrac{n^{1/(1+\beta )}}{\delta '}} \log(n)^{-\tau }  n^{-(1-\tau /(1+\beta ))}\\
		%c_\lambda ^\tau   
			%&> 64 c_\tau  \max((1+\beta )^{-1}\log\prt*{n}, \log\prt*{\nicefrac{1}{\delta '}}) \log(n)^{-\tau }n^{-(1-\tau /(1+\beta ))} \\
		%c_\lambda ^\tau   
			%&> 64 c_\tau  \max(
				%(\frac{\log\prt*{n}}{n^{((1+\beta -\tau )/((1+\beta )(1-\tau ))}} )^{1-\tau }, 
				%\log\prt*{\nicefrac{1}{\delta '}}\log(n)^{-\tau } n^{-(1-\tau /(1+\beta ))}) \\
		%c_\lambda ^\tau   
			%&> 64 c_\tau  \max\prt*{
				%\prt*{ e\frac{1+\beta -\tau }{1+\beta -\tau -\beta \tau } }^{-(1-\tau )}, 
				%\log\prt*{\nicefrac{1}{\delta '}}} \\
		%c_\lambda ^\tau   
			%&> 64 c_\tau  \max(
				%1, 
				%%( e^{-(1-\tau )}(\frac{1+\beta -\tau -\beta \tau }{1+\beta -\tau })^{1-\tau }, 
				%\log\prt*{\nicefrac{1}{\delta '}}) \\
		%c_\lambda   
			%&> (64 c_\tau  \log\prt*{\nicefrac{1}{\delta '}})^{1/\tau } \\
	%\end{align*}
	%-------------------------\\
	Assuming $c_\tau \geq 8\supfmap^2 $, a sufficient condition to satisfy the first constraint is 
	\begin{align*}
		c_\lambda ^\tau  n^{1-\tau /(1+\beta )} 
			&> (32 c_\tau ) 2\max(\log\prt*{n^{1/(1+\beta )}},  \log\prt*{(\delta ')^{-1}} ) 
	\end{align*}
	which is in particular ensured (noting that $\log(n)/n^\nu \leq 1/(\nu e)$ for any $n,\nu >0$) whenever
	\begin{align*}
		%&\lcba{
			%c_\lambda ^\tau  (n^{1/(1+\beta ))^{1+\beta -\tau } 
				%&> (32 c_\tau ) 2 \log\prt*{n^{1/(1+\beta )}} \\
		%c_\lambda ^\tau   
			%&> 64 c_\tau  (e(1+\beta -\tau ))^{-1} \\
		%c_\lambda ^\tau   
			%&> 64 c_\tau  \log\prt*{(\delta ')^{-1}}  \\
		%}
		c_\lambda   
			&> (64 c_\tau  \max((e(1+\beta -\tau ))^{-1},  \log\prt*{\nicefrac{1}{\delta '}}))^{1/\tau }  
	\end{align*}
	Noting that $1+\beta -\tau \leq 1$, we get that
	\begin{align*}
		c_\lambda   
			&> (64 c_\tau  (e(1+\beta -\tau ))^{-1} \log\prt*{\nicefrac{1}{\delta '}})^{1/\tau }  
	\end{align*}
	is also sufficient.
		We recall that $1/(1+\beta ) < 1 < 2/(\tau +1) \leq  1/\tau $, so that we can choose
	\begin{align*}
	\Aboxed{
		c_\lambda  
		&\de
		\log(\nicefrac{2}{\delta '})^{1/\tau } \max( (32 \supfmap^2   )^{1/(1+\beta )}, 
			(8 \supfmap \sqrt{c_\tau })^{2/(\tau +1)}, 
			(64 c_\tau  (e(1+\beta -\tau ))^{-1})^{1/\tau },
			8\supfmap^2 
		)
	}
	\end{align*}
	while the last constraint $n \geq  (c_\lambda /\supfmap^2 )^{1+\beta }$ is satisfied by assumption.

	\begin{align*}
		\cE(\nysest)^{1/2}
		&\leq  c_1  \lambda ^{1/2} 
			 + c_2  \lambda ^{-\tau /2} n^{-1} 
			 + c_3  \lambda ^{-\beta /2} n^{-1/2} \\
		&\leq  c_1  c_\lambda ^{1/2}n^{-1/(2(1+\beta ))} 
			 + c_2  c_\lambda ^{-\tau /2}n^{\tau /(2(1+\beta ))-1}
			 + c_3 c_\lambda ^{-\beta /2} n^{\beta /(2(1+\beta ))-1/2} \\
		&\leq  c_1  c_\lambda ^{1/2}n^{-1/(2(1+\beta ))} 
			 + c_2  c_\lambda ^{-\tau /2}n^{-(1+2\beta +(1-\tau ))/(2(1+\beta ))}
			 + c_3 c_\lambda ^{-\beta /2} n^{-1/(2(1+\beta ))} \\
		&\leq  (c_1  c_\lambda ^{1/2}
			 + c_2  c_\lambda ^{-\tau /2}
			 + c_3 c_\lambda ^{-\beta /2}) n^{-1/(2(1+\beta ))}.
	\end{align*}
	which gives the claimed result.
	The last constraint (on $m$) is satisfied by the assumptions of the lemma.
	%\begin{align*}
	%m &\geq \max(67, 5\frac{c_\tau }{\lambda ^{\tau }})\log\frac{4\supk}{\lambda \delta '}.\\
	%m &\geq \max(67, 5c_\tau c_\lambda ^{-\tau }n^{\tau /(1+\beta )})\log\frac{4\supk}{\lambda \delta '}.
	%\end{align*}
\end{tproofof*}
%%%%%%%%%%%%%%%%%%%%%%%%%%%%%%%%%%%%%%%%%%%%%%%%%%%%
%%%%%%%%%%%%%%%%%%%%%%%%%%%%%%%%%%%%%%%%%%%%%%%%%%%%
\section{Excess risk of the Nyström RRR estimator}

Recalling~\eqref{eq:RRR}, \nystrom RRR estimator is of the form $\rrrest = \irange{ \nB}_r (\rnXcov)^{-1/2}$, where $\nB := \nYXcov (\rnXcov)^{-1/2}$ for $\nYXcov := \py\eYXcov\px$ and $\rnXcov:= \px\eXcov\px + \lambda I$. While the population version is $\rrr:=\irange{ \tB}_r \rXcov^{-1/2}$ where $\tB := \YXcov (\rXcov)^{-1/2}$.  

In this section we follow the approach in \cite{kostic2023KoopmanOperatorLearning} and decompose the operator norm excess risk in the following way:
\begin{equation*}
		\cE(\rrrest)^{1/2} \!=\! \n{\mftYcX  - \krr\mftX}_{\cB(\lt,\rkhs)} + \n{(\krr-\rrr)\mftX }_{\cB(\lt, \rkhs)} + \n{(\rrr-\rrrest)\mftX}_{\cB(\lt,\rkhs)} 
\end{equation*}

Then, recalling that $\krr = \YXcov \rXcov^{-1}$ and $\nysest = \nB \rnXcov^{-1/2}$, we also have $\rrr = \ptB \krr$ and $\rrrest = \pnB\nysest$, where $\ptB$ and $\pnB$ are orthogonal projectors onto leading $r$ left singular vectors of $\tB$ and $\nB$, respectively.

Thus, 
\begin{align*}
		\cE(\rrrest)^{1/2}
        &\leq a\,\lambda^{1/2} + \sigma_{r+1}(\mftYcX) + \noprkhs{(\rrr-\rrrest)\mftX} \\
        &= a\,\lambda^{1/2} + \sigma_{r+1}(\mftYcX) + \noprkhs{(\ptB\krr-\pnB\nysest)\mftX} \\
        &\leq a\,\lambda^{1/2} + \sigma_{r+1}(\mftYcX) + \noprkhs{((\ptB-\pnB)\krr \mftX}+\noprkhs{\pnB(\krr-\nysest)\mftX} \\
        & \leq a\,\lambda^{1/2} + \sigma_{r+1}(\mftYcX) + K\,\frac{\noprkhs{\nB \nB^* - \tB\tB^*}}{\sigma_r^2(\tB)-\sigma_{r+1}^2(\tB)} +\noprkhs{(\krr-\nysest)\mftX}
		\label{e:error_decomposition_rrr}
	\end{align*}
where the last inequality is due to $\n{\krr}\leq a$ and \cite[Proposition 4]{kostic2023KoopmanOperatorLearning}.

Recalling \Cref{r:bound_ER_det_KRR_a1}, we observe that
\[
		\cE(\rrrest)^{1/2}
        \leq \sigma_{r+1}(\mftYcX) + K\,\frac{\n{\nB \nB^* - \tB\tB^*}}{\sigma_r^2(\tB)-\sigma_{r+1}^2(\tB)} + \underbrace{a\,\lambda^{1/2} + \n{(\krr-\nysest)\mftX}}_{\leq \text{r.h.s. of the bound in \Cref{r:bound_ER_det_KRR_a1}}}
		\label{e:error_decomposition_rrr}
\]

Therefore, to prove \Cref{r:bound_ER_RRR_a1} for the RRR estimator we just need to bound $\noprkhs{\nB \nB^* - \tB\tB^*}$. To that end, observe that, after some algebra, one obtains
\[
\nB \nB^* - \tB\tB^* \!=\! \krr (\nYXcov - \YXcov)^* + (\nYXcov - \YXcov) \krr^* - \krr (\rnXcov - \rXcov) \krr^*  +(\nkrr - \krr) \rnXcov (\nkrr-\krr)^*, 
\]
and, consequently,
\begin{align*}
    \noprkhs{\nB \nB^* - \tB\tB^*} \leq 2a & \noprkhs{\nYXcov - \YXcov} + a^2\noprkhs{\px\eXcov\px - \Xcov} \\
    & + \noprkhs{\rXcov^{-1/2}\rnXcov\rXcov^{-1/2}}\noprkhs{(\nkrr - \krr)\rXcov^{1/2}}^2,
\end{align*}
follows using that $\noprkhs{\krr}\leq a$.

On the other hand, 
\[
\noprkhs{\nYXcov - \YXcov}\leq \noprkhs{\py(\eYXcov - \YXcov)\px} + \noprkhs{\py^\perp\YXcov\px} + \noprkhs{\YXcov \px^\perp},
\]
which implies that
\[
\noprkhs{\nYXcov - \YXcov}\leq \noprkhs{\eYXcov - \YXcov} + 2\,a\,K\,\varepsilon_1,
\]
where $\varepsilon_1:=\max\{\noprkhs{\px^\perp\Xcov^{1/2}},\,\noprkhs{\py^\perp\Xcov^{1/2}}\}$. Similarly, we obtain 
\begin{equation}\label{eq:rncov}
\noprkhs{\rnXcov - \rXcov}\leq \noprkhs{\eXcov - \Xcov} + 2\,K\,\varepsilon_1.    
\end{equation}

But, $\varepsilon_1$ can be bounded by \Cref{r:uniform_nys_approx}. Indeed, provided $\lambda \in ]0,\noprkhs{\Xcov}]$,  
	it holds with probability $1-\delta '$
	\[ \varepsilon_1 \leq  \sqrt{3\lambda } \]
	provided $m\geq \max(67, 5\esssup_{x\sim \td} \n{\rXcov^{-1/2}\fmap{x}}^2 )\log\frac{4\supk}{\lambda \delta '}$, which by \Cref{r:bound_Ninf} is ensured if $m\geq \max(67, 5\frac{c_\tau }{\lambda ^{\tau }})\log\frac{4\supk}{\lambda \delta '}$.

Additionally,
\begin{align*}
\noprkhs{\rXcov^{-1/2}\rnXcov\rXcov^{-1/2}} & \leq \noprkhs{ \rXcov^{-1/2}\px\reXcov\px\rXcov^{-1/2}} + \lambda \noprkhs{ \rXcov^{-1/2}\px^\perp\rXcov^{-1/2}} \\
& \leq \theta_2^2 \, \noprkhs{ \reXcov^{-1/2}\px\reXcov\px\rXcov^{-1/2}} + 1 \\
& \leq \theta_2^2 \, \noprkhs{ \reXcov^{1/2}\px\reXcov^{-1}\px\rXcov^{1/2}} + 1 \\
& \leq \theta_2^2 \, \noprkhs{ \reXcov^{1/2}\px (\px\reXcov\px)^{\dagger}\px\rXcov^{1/2}} + 1, %\leq \theta_2^2 +1 \leq 2.25, 
\end{align*}
implies that 
\begin{equation}\label{eq:theta2_bound}
 \noprkhs{\rXcov^{-1/2}\rnXcov\rXcov^{-1/2}} \leq \theta_2^2 +1 \leq 2.25,   
\end{equation}
provided, as above, that $n\lambda ^\tau  > 32 c_\tau  \beta $.

Therefore, setting $\varepsilon_0 := \max\{a\noprkhs{\eYXcov - \YXcov}, a^2\noprkhs{\eXcov - \Xcov} \}$, for all $i\in[m]$ we have
\begin{equation}\label{eq:Bbound}
    \abs{\sigma_i^2(\nB)-\sigma_i^2(\tB)}\leq \n{\nB \nB^* - \tB\tB^*} \leq 3 \varepsilon_0+ 6.93\,K\,a^2\lambda^{1/2} + 2.25\,\varepsilon_2^2,
\end{equation}
where $\varepsilon_2:= \n{(\krr-\nysest)\rXcov^{1/2}}$ is the variance of \nystrom{} KRR estimator, and conclude that
\[
\cE(\rrrest)^{1/2} \leq \sigma_{r+1}(\mftYcX) + K\,\frac{3 \varepsilon_0 + 6.93\,K\,a^2\,\lambda^{1/2} + 2.25\,\varepsilon_2^2}{\sigma_r^2(\tB)-\sigma_{r+1}^2(\tB)} + a\,\lambda^{1/2} + \varepsilon_2.
\]
Therefore, the proof of \Cref{r:bound_ER_RRR_a1} for RRR estimator directly follows from the bound on $a\,\lambda^{1/2}+\varepsilon_2$ given in the proof of \Cref{r:bound_ER_KRR_a1}, and the fact that, see e.g.~\cite{kostic2022learning},  $\varepsilon_0 \lesssim n^{-1/2} \lesssim\lambda^{1/2}$.

\section{Excess risk of the Nyström PCR estimator}

Recalling~\Cref{{e:nys_pcr_app}}, \nystrom PCR estimator is of the form 
\[
\pcrest = \py\eYXcov \irange{\px\eXcov\px}_r^\dagger = \nYXcov \irange{\rnXcov}_r = \nysest \pnC,
\] 
for $\lambda=0$ and with $\pnC$ being the orthogonal projector onto leading $r$ eigenspace of $\rnXcov$. So, to prove \Cref{r:bound_ER_RRR_a1}  for PCR estimator, denote $\npcr:=\nysest \pnC$ for $\lambda\geq0$, and let us define the population version $\pcr = \krr \ptC$, where $\ptC$ being the orthogonal projector onto leading $r$ eigenspace of $\rXcov$.

As in the previous section we start with decomposition
\begin{align*}
		\cE(\pcrest)^{1/2} = & \n{\mftYcX  - \krr\mftX}_{\cB(\lt,\rkhs)} \,+\, \n{(\krr-\pcr)\mftX }_{\cB(\lt,\rkhs)}\, + \\
   & \n{(\pcr-\npcr)\mftX}_{\cB(\lt,\rkhs)} \,+\, \n{(\npcr-\pcrest)\mftX}_{\cB(\lt,\rkhs)}.
\end{align*}

The first and the second term are easily bounded by $\n{\mftYcX  - \krr\mftX}_{\cB(\lt,\rkhs)}\leq a\,\lambda^{1/2}$, and
\begin{align*}
\n{(\krr-\pcr)\mftX }_{\cB(\lt,\rkhs)} & = \n{\krr(I-\ptC)\mftX }_{\cB(\lt,\rkhs)} \\
& \leq a\, \noprkhs{(I-\ptC)\Xcov^{1/2} }\leq a \,\sigma_{r+1}(\mftX).
\end{align*}

For the third term, start by observing that 
\begin{align*}
\n{(\pcr-\npcr)\mftX}_{\cB(\lt,\rkhs)} & = \noprkhs{(\krr\ptC - \nysest\pnC)\Xcov^{1/2}} \\
& \leq \noprkhs{\krr(\ptC - \pnC)\Xcov^{1/2}} + \noprkhs{(\krr-\nysest)\pnC\Xcov^{1/2}}\\
&\leq a\,K\, \noprkhs{\ptC - \pnC} +  \noprkhs{(\krr-\nysest)\pnC\Xcov^{1/2}}\\
& \leq a\,K\, \frac{\noprkhs{\rnXcov - \rXcov}}{\sigma_r^2(\mftX) - \sigma_{r+1}^2(\mftX)}  + \noprkhs{(\krr-\nysest)\pnC\Xcov^{1/2}}\\
& \leq K\, \frac{\varepsilon_0 + 2\,a\,K\,\varepsilon_1}{\sigma_r^2(\mftX) - \sigma_{r+1}^2(\mftX)}  + \noprkhs{(\krr-\nysest)\pnC\Xcov^{1/2}}
\end{align*}
where the second last inequality is due to \cite[Proposition 4]{kostic2023KoopmanOperatorLearning} and the last one uses \Cref{eq:rncov}. Moreover, we have that 
\begin{align*}
 \noprkhs{(\krr-\nysest)\pnC\Xcov^{1/2}} &\leq  \noprkhs{(\krr-\nysest)\rXcov^{1/2}} \, \noprkhs{\rXcov^{-1/2}\rnXcov^{1/2}} \, \noprkhs{\rnXcov^{-1/2} \pnC\Xcov^{1/2}}\\
& \leq \varepsilon_2 \, \sqrt{1+\theta_2^2} \, \noprkhs{\pnC \rnXcov^{-1/2} \Xcov^{1/2}}\\
& \leq \varepsilon_2 \, \sqrt{1+\theta_2^2} \, \noprkhs{ \rXcov^{1/2} \rnXcov^{-1/2}},
\end{align*}
where we have used \Cref{eq:theta2_bound} and the fact that $\pnC$ is the spectral projector of $\rnXcov^{-1/2}$. Therefore, due to
\begin{align*}
 \noprkhs{ \rXcov^{1/2} \rnXcov^{-1/2}} & = \noprkhs{ \rXcov^{1/2}[\px^\perp + \px] \rnXcov^{-1/2}} \leq \noprkhs{ \rXcov^{1/2}\px \rnXcov^{-1/2}} + \noprkhs{ \rXcov^{1/2}\px^\perp\rnXcov^{-1/2}}  \\
 & \leq \theta_2\noprkhs{ \reXcov^{1/2}\px\rnXcov^{-1/2}} + \noprkhs{ \rXcov^{1/2}\px^\perp}\noprkhs{\rnXcov^{-1/2}} \\
 &\leq \theta_2\noprkhs{ \reXcov^{1/2}\px(\px\reXcov\px)^\dagger\px \reXcov^{1/2}}^{1/2} + \noprkhs{ \rXcov^{1/2}\px^\perp}\lambda^{-1/2} \\
 & \leq \theta_2 + \varepsilon_1\lambda^{-1/2} 
\end{align*}
we obtain
\[
\noprkhs{(\krr-\nysest)\pnC\Xcov^{1/2}} \leq \varepsilon_2 \left( 1.68 + 1.5\lambda^{-1/2}\,\varepsilon_1 \right),
\]
provided that $n\lambda ^\tau  > 32 c_\tau  \beta $.

Finally for the last term, observe that $\irange{\rnXcov}_r^\dagger$ and $\irange{\tilde{C}_{0}}_r^\dagger$ share the same eigenvectors, and hence $\irange{\tilde{C}_{0}}_r^\dagger - \irange{\rnXcov}_r^\dagger = \lambda \irange{\rnXcov  \tilde{C}_{0}}_r^\dagger $. Hence, it holds that
\begin{align*}
 \n{(\npcr-\pcrest)\mftX}_{\cB(\lt,\rkhs)} & = \noprkhs{\nYXcov(\irange{\rnXcov}_r^\dagger-\irange{\tilde{C}_{0}}_r^\dagger)\Xcov^{1/2}} = \lambda\noprkhs{\nYXcov \rnXcov^{-1} \irange{\tilde{C}_{0}}_r^\dagger \Xcov^{1/2}}\\
 &\leq \lambda\noprkhs{\nYXcov \rnXcov^{-1} \irange{\tilde{C}_{0}}_r^\dagger \rnXcov^{1/2}}\noprkhs{\rnXcov^{-1/2}\rXcov^{1/2}} \\
 & = \lambda\noprkhs{\nYXcov \rnXcov^{-1/2} \irange{\tilde{C}_{0}}_r^\dagger}\noprkhs{\rnXcov^{-1/2}\rXcov^{1/2}} \\
 & = \lambda\,\noprkhs{\irange{\tilde{C}_{0}}_r^\dagger}\, \noprkhs{\nB}\,\noprkhs{\rnXcov^{-1/2}\rXcov^{1/2}}.
\end{align*}

Now, recalling \Cref{eq:Bbound}, we can bound 
\begin{align*}
\noprkhs{\nB}^2 &\leq \noprkhs{\rXcov^{-1/2} \YXcov^*\YXcov \rXcov^{-1/2}} + \noprkhs{\tB\tB^* - \nB\nB^* } \\
& \leq a^2 K^2 + 3 \varepsilon_0+ 6.93\,K\,a^2\lambda^{1/2} + 2.25\,\varepsilon_2^2,
\end{align*}
and, 
\[
\lambda^{1/2} \noprkhs{\irange{\tilde{C}_{0}}_r^\dagger} = \frac{\lambda^{1/2}}{\lambda_r(\px\eXcov\px)} \leq \frac{\lambda^{1/2}}{\lambda_r(\Xcov) - \noprkhs{\rXcov - \rnXcov}} \leq \frac{\lambda^{1/2}}{\sigma_r^2(\mftX) -\noprkhs{\eXcov - \Xcov} - 2\,K\,\varepsilon_1}.
\]
Thus, consequently, we obtain 
\begin{align*}
 \n{(\npcr-\pcrest)\mftX}_{\cB(\lt,\rkhs)} \leq & \lambda^{1/2} \left(\theta_2 + \varepsilon_1\lambda^{-1/2} \right)\, \left( a^2 K^2 + 3 \varepsilon_0+ 6.93\,K\,a^2\lambda^{1/2} + 2.25\,\varepsilon_2^2 \right) \cdot\\
 & \hspace{0.5cm}\frac{\lambda^{1/2}}{\sigma_r^2(\mftX) -\noprkhs{\eXcov - \Xcov} - 2\,K\,\varepsilon_1}.
\end{align*}

To conclude, observe that $r> n^{\frac{1}{\beta(1+\beta)}}$ due to \Cref{a:spectral_decay} implies that $\sigma_{r+1}(\mftX)\lesssim n^{-\frac{1}{2(1+\beta)}}$.

Therefore, collecting all the terms, under the assumptions of \Cref{r:bound_ER_RRR_a1} we obtain
\[
\ERop(\pcrest)^{1/2} \lesssim c_{\rm PCR}\,n^{-\frac{1}{2(1+\beta)}},
\]
where $c_{\rm PCR} = (\sigma_{r}^2(\mftX) - \sigma_{r+1}^2(\mftX))^{-1}$ is the problem dependant constant.

\section{Auxiliary results}

\begin{tlemma}{\cite[Proposition 2]{kostic2023KoopmanOperatorLearning} with $\alpha=1$}{bound_rest}
	%Let $P_{\rkhs}:\lt\rightarrow \lt$ be the orthogonal projector onto $\cl(\Im(\amftX))$. \todo{$\rkhs$?}
	Under \Cref{a:universal_kernel} it holds
	\begin{align*}
		%\n{\kop \amftX - \amftX \rXcov^{-1}\XYcov}
		\n{\mftX\kop^*  - \YXcov\rXcov^{-1}\mftX }
		&\leq  a\lambda ^{1/2}.
		%&\leq  a\lambda ^{\alpha /2}.
			%+ \n{(I-P_{\rkhs})\kop \amftX}.
	\end{align*}
\end{tlemma}

\begin{tlemma}{}{bound_rest_opnorm_A_a1}
	Let $A$ be a bounded operator. 
	Under \Cref{a:regularity,a:bounded_fmap}, it holds
	\begin{align}
		\n*{\YXcov A} &\leq  a \n{\Xcov A}
	\end{align}
\end{tlemma}
\begin{tproofof*}{r:bound_rest_opnorm_A_a1}{}
	Note that under \Cref{a:regularity}, as $\XYcov\YXcov\preccurlyeq a^2 \Xcov^{2}$ it also holds 
	$A^*\XYcov\YXcov A \preccurlyeq a^2 A\Xcov^2 A$ 
	and thus: 
	\begin{align*}
	\n{\YXcov A}
		&= \n{A^* \YXcov\YXcov A}^{1/2}\\
		&\leq  a \n{A^* \Xcov^2 A}^{1/2}\\
		&= a \n{\Xcov A}.
	\end{align*}
\end{tproofof*}

%%%%%%%%%%%%%%%%%%%%%%%%%%%%%%%%%%%%%%%%%%%%%%%%%%%%%%%%%
%%%%%%%%%%%%%%%%%%%%%%%%%%%%%%%%%%%%%%%%%%%%%%%%%%%%%%%%%

The next lemma is a consequence of \Cref{a:embedding_property} and will be used in our concentration inequalities.
\begin{tlemma}{}{bound_Ninf}
	Under \Cref{a:embedding_property}, it holds $\td$-almost surely for any $\nu $:
	\begin{align*}
	\n*{\rXcov^{-(1-\nu )/2}\fmap{x}}^2 
		&\leq  c_\tau  \lambda ^{-[\tau -\nu ]_+}\supfmap^{2[\nu -\tau ]_+}. 
	\end{align*}
	The two following corollaries can be obtained picking $\nu =0$ and $\nu =-1$:
	\begin{align*}
	\n*{\rXcov^{-1/2}\fmap{x}}^2 
		&\leq  \frac{c_\tau }{\lambda ^{\tau }} 
	\quad\text{and}\quad
	\n*{\rXcov^{-1}\fmap{x}}^2 
		\leq  \frac{c_\tau }{\lambda ^{\tau +1}}.
\end{align*}
\end{tlemma}
~
\begin{tproofof*}{r:bound_Ninf}{}
	By \cite[Theorem 9]{fischer2020SobolevNormLearning}, it holds $c_\tau \de \n{k_{\td}^\tau }_\infty ^2 =\esssup_{x\sim\td} \sum_{i\in I} \mu _i^\tau  |e_i(x)|^2 $ (where $(e_i)$ is defined in \Cref{s:interpolating_spaces}, and we recall that $(\sqrt{\mu _i}e_i)_{i\in \bN}$ is an orthonormal basis of $\rkhs$. 

	Denoting $\mu_i\de \lambda_i(\Xcov)$, it holds
	\begin{align*}
	\n*{\rXcov^{-(1-\nu )/2}\fmap{x}}^2 
		&= \n*{\prt*{\sum_{i\in I} (\mu _i+\lambda )^{-(1-\nu )/2} (\sqrt{\mu _i}e_i) \kron (\sqrt{\mu _i}e_i)}\fmap{x}}^2  \\
		&= \prt*{\sum_{i\in I} \mu _ie_i(x)^2 (\mu _i+\lambda )^{-1+\nu } } \\
		&= \sum_{i\in I} \mu _i^{1-\tau }{(\mu _i+\lambda )}^{-1+\nu } \mu _i^\tau  e_i(x)^2   \\
		&= \sum_{i\in I} \prt*{\frac{\mu _i}{\mu _i+\lambda }}^{1-\tau }{(\mu _i+\lambda )}^{\nu -\tau } \mu _i^\tau  e_i(x)^2   \\
		&\leq  \sum_{i\in I} (\mu _i+\lambda )^{-(\tau -\nu )} \mu _i^\tau  e_i(x)^2   \\
		&\leq  c_\tau \lambda ^{-[\tau -\nu ]_+}\supfmap^{2[\nu -\tau ]_+}. 
	\end{align*}
	where we used $\sup |\mu _i|\leq \supfmap^2$. 
\end{tproofof*}

\color{black}
\section{Deterministic sketching results}

\begin{tlemma}{}{rkrrls_onehalf}
	Denoting $R \de I-\reXcov^{1/2} \regikrr \reXcov^{1/2}$, 
	it holds 
	\begin{align*}
		R \reXcov^{1/2} 
		&= R \reXcov^{1/2} \px^\perp .
	\end{align*}
\end{tlemma}
\begin{tproofof*}{r:rkrrls_onehalf}{}
	This is a direct consequence of the fact that $\regikrr \reXcov \px=\px$:
	\begin{align*}
		R \reXcov^{1/2} \px
		&= \reXcov^{1/2} \px -\reXcov^{1/2} \regikrr \reXcov \px
		= 0.
	\end{align*}
\end{tproofof*}

\section{Concentration results}

%%%%%%%%%%%%%%%%%%%%%%%%%%%%%%%%%%%%%%%%%%%%%%%%%%%%%%%%%
%%%%%%%%%%%%%%%%%%%%%%%%%%%%%%%%%%%%%%%%%%%%%%%%%%%%%%%%%
\subsection{Generic concentration lemmas}

All our concentration results derive from two versions of the Bernstein inequality.
We first state an inequality for sums of random variables in a Hilbert space based on \cite[Theorem 3.3.4]{yurinsky1995SumsGaussianVectors}, which itself derives from a result of \cite{pinelis1986RemarksInequalitiesLarge}.

\begin{tlemma}{}{concentration_yurinsky}
	Let $(A_i)_{1\leq i\leq n}$ be \tiid copies of a random variable $A$ in a separable Hilbert space $(H,\n{\cdot })$. Assume $\E A=\mu $ and $\exists \sigma >0,\exists L>0,\forall p\geq 2, \E \n{A-\mu }^p\leq \tfrac{1}{2} p!\sigma ^2 L^{p-2}$.
	Then for any $\delta \in ]0,1[$ it holds:
	\begin{align}
	P\brk*{ \n*{\frac{1}{n}\sum_{i=1}^n A_i - \mu  } \leq  
		\frac{2L \log(2/\delta )}{n} + \sqrt{\frac{2\sigma ^2 \log(2/\delta )}{n}} 
	} \geq  1-\delta 	
	\end{align}
	The assumption on the moments holds in particular when $\esssup \n{A}\leq L/2$ and $\E[\n{A}^2 ]\leq \sigma ^2 $. 
\end{tlemma}
\begin{tproofof*}{r:concentration_yurinsky}{}
	See proof of \cite[Lemma E.3]{chatalic2022NystromKernelMean} for a precise derivation based on \cite[Theorem 3.3.4]{yurinsky1995SumsGaussianVectors}.
\end{tproofof*}

We now state a version of the Bernstein concentration inequality for self-ajoint operators in operator norm, which is a restatement of \cite[Lemma 24]{lin2020OptimalConvergenceDistributed}.
In the following, we denote $\erk(A)\de \Tr(A)/\n{A}$ the effective rank of a nonnegative definite operator $A$.
\begin{tlemma}{Bernstein for self-ajoint operators acting on a Hilbert}{concentration_sa_HS_op_norm}
	Let $H$ be a separable Hilbert space and $A_i$ be \tiid copies of a random variable $A$ taking values in the space of self-adjoint Hilbert-Schmidt operators on $H$.
	Assume $\E A=0$, $\esssup \nop{A}\leq c$ for some $c>0$ (where $\nop{\cdot }$ denotes the operator norm) and that there exists a positive semi-definite trace class operator $V$ such that $\E[A^2 ]\preccurlyeq V$.
	Then for any $\delta \in ]0,1[$ and $n\geq 1$ it holds
	\begin{align}
	P\brk*{ \nop*{\frac{1}{n}\sum_{i=1}^n A_i } \geq  
		\frac{2c \beta }{3n} + \sqrt{\frac{2\n{V}\beta }{n}} }
	\leq  \delta 	
	\quad\text{ where }\quad
	\beta  = \log\prt*{\tfrac{4 \erk(V)}{\delta }} 
	\end{align}
\end{tlemma}
\begin{tproofof*}{r:concentration_sa_HS_op_norm}{}
	See \cite[Appendix B.7, Lemma 24]{lin2020OptimalConvergenceDistributed}.
\end{tproofof*}

\subsection{Applied concentration lemmas}

\begin{tlemma}{}{concentration_Xcov_sa_half}
	% NOTE: here apart from the the constant can be improved, there is a real gain using operator norm as we get ‖CC_\lambda ^{-1}‖\leq 1 otherwise we would get \sigma ^2 =E[‖A‖HS^2 ] -> effective dimension
	Let \Cref{a:bounded_fmap} hold.
	Let $\delta \in ]0,1[$. Then for \tiid samples $(x_i,y_i)_{1\leq i\leq n}$ and any $\lambda \in ]0,\noprkhs{\Xcov}]$ it holds
	\begin{align}
		P\brk*{ \noprkhs*{ \rXcov^{-1/2}(\eXcov - \Xcov)\rXcov^{-1/2} } \geq  
			%\frac{2(1+\supfmap^2/\lambda ) \beta }{3n} + \sqrt{\frac{2\supfmap^2 \beta }{\lambda n}} }
			\frac{4c_\tau  \beta }{3n\lambda ^{\tau }} + \sqrt{\frac{2c_\tau \beta }{n\lambda ^{\tau }}} }
		\leq  \delta 	
		\quad\text{ where }\quad
		\beta  = \log\prt*{\tfrac{8 \supfmap^2 }{\delta \lambda }} 
	\end{align}
	%\todo{We can get $(1+\supfmap^2/\lambda )$ instead of $2\supfmap^2 /\lambda $\ldots  but maybe it also helps to keep simple expressions?}
\end{tlemma}
\begin{tproofof*}{r:concentration_Xcov_sa_half}{}
	We apply \Cref{r:concentration_sa_HS_op_norm} on the random variables $A_i=\xi (X_i) \kron \xi (X_i) - \rXcov^{-1/2} \Xcov \rXcov^{-1/2}$ where $\xi (X_i)\de\rXcov^{-1/2} \fmap(X_i)$.
	It holds
	\begin{align*}
	\esssup \noprkhs{A_i}
		%&\leq  \esssup \nrkhs{\xi (X_i)}^2  + \noprkhs{\rXcov^{-1/2} \Xcov \rXcov^{-1/2}}\\
		%&\leq  \frac{c_\tau }{\lambda ^{\tau }} + 1. \quad \text{(by \Cref{r:bound_Ninf})}\\
		&\leq  2\esssup \nrkhs{\xi (X_i)}^2  \\
		&\leq  \frac{2c_\tau }{\lambda ^{\tau }}. \quad \text{(by \Cref{r:bound_Ninf})}\\
	\E[A_i^2 ]
		&= \E[ \n{\xi (X_i)}^2  \xi (X_i)\xi (X_i)^* ]- (\rXcov^{-1/2} \Xcov \rXcov^{-1/2})^2 \\
		&\preccurlyeq  \E[ \n{\xi (X_i)}^2  \xi (X_i)\xi (X_i)^* ]\\
		&\preccurlyeq  \frac{c_\tau }{\lambda ^{\tau }} \E [ \xi (X_i)\xi (X_i)^* ]\\
		&= \frac{c_\tau }{\lambda ^{\tau }} \Xcov \rXcov^{-1}
	\end{align*}
	Thus applying \Cref{r:concentration_sa_HS_op_norm} with
	%$c=1+\frac{c_\tau }{\lambda ^{\tau }}$
	$c=\frac{2c_\tau }{\lambda ^{\tau }}$
	and $V=\frac{c_\tau }{\lambda ^{\tau }}\Xcov \rXcov^{-1}$, we get
	\begin{align}
		P\brk*{ \n*{\frac{1}{n}\sum_{i=1}^n A_i } \geq  
			\frac{4c_\tau  \beta }{3n\lambda ^{\tau }} + \sqrt{\frac{2c_\tau \beta }{\lambda ^{\tau }n}} }
		\leq  \delta 	
		\quad\text{ where }\quad
		%\beta  = \log\prt*{\tfrac{4 \Tr(V)}{\noprkhs{V} \delta }} 
		\beta  = \log\prt*{\tfrac{8 \supfmap^2 }{\delta \lambda }} 
	\end{align}
	where we used the fact that $\noprkhs{\Xcov\rXcov^{-1}}\leq 1$ and controlled the effective rank using $\Tr({\Xcov\rXcov^{-1}})\leq \supfmap^2 /\lambda $ and $\noprkhs{\Xcov\rXcov^{-1}}=\noprkhs{\Xcov}/(\noprkhs{\Xcov}+1)\geq 1/2$ because $\lambda \leq \noprkhs{\Xcov}$ by assumption.
\end{tproofof*}

\begin{tlemma}{}{concentration_Xcov_onehalf}
	Let \Cref{a:bounded_fmap,a:embedding_property} hold.
	Let $\delta \in ]0,1[$. Then for \tiid samples $(x_i,y_i)_{1\leq i\leq n}$ we get
	\begin{align}
	P\brk*{\nop{(\Xcov-\eXcov)\rXcov^{-1/2}}
			\leq  \epsilon (\lambda ,\delta )
		} &\geq  1-\delta  
		\label{e:concentration_Xcov_onehalf}\\
	\quad\text{and}\quad 
	P\brk*{\nop{(\YXcov-\eYXcov)\rXcov^{-1/2}}
			\leq  \epsilon (\lambda ,\delta )
		} &\geq  1-\delta  
		\label{e:concentration_YXcov_onehalf}\\
	\text{where}\quad
	\epsilon (\lambda ,\delta ) 
		&\de \frac{2\supfmap \sqrt{c_\tau } \log(\nicefrac{2}{\delta })}{\lambda ^{\tau /2} n} + \sqrt{\frac{2 \supfmap^2  \deff \log(\nicefrac{2}{\delta })}{n}}
	\end{align}
\end{tlemma}
\begin{tproofof*}{r:concentration_Xcov_onehalf}{}
	We first write the proof for the \cref{e:concentration_Xcov_onehalf}.
	For this result, we use the fact that $\noprkhs{(\Xcov-\eXcov)\rXcov^{-1/2}}\leq \nhs{(\Xcov-\eXcov)\rXcov^{-1/2}}$ and bound the Hilbert-Schmidt norm.
	As $(\hsH,\nhsH{\cdot })$ is a Hilbert space, we apply \Cref{r:concentration_yurinsky} on the random variables $A_i=\fmap{x_i}\kron \xi (x_i)$ where $\xi (x)=\rXcov^{-1/2}\fmap{x}$. 
	\begin{align*}
	\esssup \nhs{A} 
		&= \esssup \nrkhs{\fmap{x}}\nrkhs{ \xi (x)} \\
		&\leq  \frac{\supfmap \sqrt{c_\tau }}{\lambda ^{\tau /2}}
			\quad\text{(by \cref{a:bounded_fmap,r:bound_Ninf})} \\
	\E[\nhs{A}^2 ]
		&= \E[\nrkhs{\fmap(x)}^2 \nrkhs{\xi (x)}^2 ] \\
		&\leq  \supfmap^2  \deff 
	\end{align*}
	Thus applying \Cref{r:concentration_yurinsky} with $L=\frac{\supfmap \sqrt{c_\tau }}{\lambda ^{\tau /2}}$ and $\sigma ^2 =\supfmap^2  \deff$ gives
	\begin{align*}
	%P\brk*{ \n*{\frac{1}{n}\sum_{i=1}^n A_i - \mu  } \leq  
		%\frac{4\supfmap^2  \log(2/\delta )}{n\sqrt{\lambda }} + \sqrt{\frac{2\supfmap^2  \deff\log(2/\delta )}{n}} 
	%} \geq  1-\delta 	
	%\quad\text{\cred{(Without embedding condition)}}
	%\\
	P\brk*{ \nhs*{\frac{1}{n}\sum_{i=1}^n A_i - \mu  } \leq  
		\frac{2\supfmap \sqrt{c_\tau } \log(\nicefrac{2}{\delta })}{\lambda ^{\tau /2} n} + \sqrt{\frac{2 \supfmap^2  \deff \log(\nicefrac{2}{\delta })}{n}} 
	} \geq  1-\delta .
	%\quad\text{\cred{(with embedding condition)}}
	\end{align*}
	This yields the desired result via the inequality between operator and Hilbert-Schmidt norms.

	For the bound \cref{e:concentration_YXcov_onehalf} on the cross-covariance, we take $A_i=\fmap{y_i}\kron \xi (x_i)$ but the rest of the proof is inchanged.
\end{tproofof*}

\begin{tlemma}{}{concentration_Xcov_inv_oneside}
	Let \Cref{a:bounded_fmap,a:embedding_property} hold.
	Let $\delta \in ]0,1[$. Then for \tiid samples $(x_i,y_i)_{1\leq i\leq n}$ we get
	\begin{align}
	P\brk*{ \nop{(\Xcov-\eXcov)\rXcov^{-1}} \leq  
		\frac{2\supfmap \sqrt{c_\tau } \log(\nicefrac{2}{\delta })}{\lambda ^{(\tau +1)/2} n} + \sqrt{\frac{2 \supfmap^2  \Tr(\rXcov^{-2}\Xcov) \log(\nicefrac{2}{\delta })}{n}} 
	} \geq  1-\delta .
	\end{align}
\end{tlemma}
\begin{tproofof*}{r:concentration_Xcov_inv_oneside}{}
	For this result, we use the fact that $\noprkhs{(\Xcov-\eXcov)\rXcov^{-1}}\leq \nhs{(\Xcov-\eXcov)\rXcov^{-1}}$ and bound the Hilbert-Schmidt norm.
	As $(\hsH,\nhsH{\cdot })$ is a Hilbert space, we apply \Cref{r:concentration_yurinsky} on the random variables $A_i=\fmap{x_i}\kron \omega (x_i)$ where $\omega (x)=\rXcov^{-1}\fmap{x}$. 
	\begin{align*}
	\esssup \nhs{A} 
		&= \esssup \nrkhs{\fmap{x}} \nrkhs{\omega (x)} \\
		&\leq  \frac{\supfmap \sqrt{c_\tau }}{\lambda ^{(\tau +1)/2}} 
			\quad\text{(by \cref{a:bounded_fmap,r:bound_Ninf})} \\
	\E[\nhs{A}^2 ] 
		&= \E[\nrkhs{\fmap(x)}^2 \nrkhs{\omega (x)}^2 ] \\
		&\leq  \supfmap^2  \E[\Tr\prt{\rXcov^{-2}\fmap{x}\fmap{x}^*}] \\
		&= \supfmap^2  \Tr\prt{\rXcov^{-2}\Xcov}
	\end{align*}
	Thus applying \Cref{r:concentration_yurinsky} with $L=\frac{\supfmap \sqrt{c_\tau }}{\lambda ^{(\tau +1)/2}}$ and $\sigma ^2 =\supfmap^2  \Tr(\rXcov^{-2}\Xcov)$ gives
	\begin{align*}
	%P\brk*{ \n*{\frac{1}{n}\sum_{i=1}^n A_i - \mu  } \leq  
		%\frac{4\supfmap^2  \log(2/\delta )}{n\sqrt{\lambda }} + \sqrt{\frac{2\supfmap^2  \deff\log(2/\delta )}{n}} 
	%} \geq  1-\delta 	
	%\quad\text{\cred{(Without embedding condition)}}
	%\\
	P\brk*{ \nhs*{\frac{1}{n}\sum_{i=1}^n A_i - \mu  } \leq  
		\frac{2\supfmap \sqrt{c_\tau } \log(\nicefrac{2}{\delta })}{\lambda ^{(\tau +1)/2} n} + \sqrt{\frac{2 \supfmap^2  \Tr(\rXcov^{-2}\Xcov) \log(\nicefrac{2}{\delta })}{n}} 
	} \geq  1-\delta .
	%\quad\text{\cred{(with embedding condition)}}
	\end{align*}
	This yields the desired result via the inequality between operator and Hilbert-Schmidt norms. 
	%For the bound \cref{e:concentration_YXcov_onehalf} on the cross-covariance, we take $A_i=\fmap{y_i}\kron \xi (x_i)$ but the rest of the proof is inchanged.
\end{tproofof*}

\color{black}
\subsection{Probabilistic inequalities}

\begin{tlemma}{Uniform Nyström approximation}{uniform_nys_approx}
	Let \Cref{a:bounded_fmap} hold.
	Let $P:\rkhs\rightarrow \rkhs$ denote the orthogonal projection on $\spa\Set{\fmap{\ldm{j}} | 1\leq j\leq m}$,
	where the landmarks $(\ldm{j})_{1\leq j\leq m}$ are drawn \tiid from the empirical data. Then for any $\lambda \in ]0,\noprkhs{\rXcov}]$ we have 
	\[ \noprkhs{P^\perp  \rXcov^{1/2}}^2 \leq  3\lambda  \]
	with probability at least $1-\delta $ provided 
	\[ m\geq \max(67, 5\esssup \n{\rXcov^{-1/2}\fmap{x}}^2)\log\frac{4\supk}{\lambda \delta }. \] 
\end{tlemma}

%\begin{tlemma}{ALS Nystöm approximation \cite[Lemma 7]{rudi2016LessMoreNystr}}{als_nys_approx}
%\todo{adapt notations $X/Y$, $P$, iid sampling\ldots }
%Let $\lambda >0$ and $\delta \in ]0,1[$.
%Let $(\hat{l}_i(t))_{1\leq i\leq n}$ be a collection of $(z,\lambda ₀,\delta /2)$-approximate leverage scores. Let $\lambda >0$, and $p_\lambda $ be a probability distribution on the set of indexes $\cb*{1,\ldots ,n}$ defined as $p_\lambda (i)\de \hat{l}_i(\lambda )/(\sum_{i=1}^n \hat{l}_i(\lambda ))$. Let $\cb*{i_1,\ldots ,i_m}$ be a collection of indices sampled independently with replacement from $p_\lambda $, and $P$ the orthogonal projection on $\spa\cb*{\fmap(x_{i_1 }),\ldots ,\fmap(x_{i_m})}$.
%We have with probability at least $1-\delta $ 
	%\[ \noprkhs{P^\perp  \rXcov^{1/2}}^2 \leq  3\lambda \]
	%provided that:
	%\begin{itemize}
		%\item there exists $z\geq 1$ and $\lambda ₀>0$ such that the $\hat{l}_i$ are $(z,\lambda ₀,\delta /2)$-approximate leverage scores ;
		%\item $n \geq  1655\supk + 233\supk \log(4\supk/\delta )$;
		%\item $\max\prt*{\lambda ₀, \frac{19\supk}{n}\log(\frac{4n}{\delta })} \leq  \lambda  \leq  \noprkhs{\Xcov}$;
		%\item $m \geq  \max\prt*{334, 78z^2\cN(\lambda )}\log\frac{16n}{\delta }$.
	%\end{itemize}
%\end{tlemma}

\color{black}
\subsection{Concentration lemmas for the sketched operators}

\begin{tlemma}{}{bound_proj_crosscov_onehalf}
	It holds almost surely
	\begin{align*}
		\n{ (\YXcov - \py \eYXcov )\rXcov^{-1/2}}
		&\leq  \n{\py^\perp \rXcov^{1/2}} 
			+ \n{(\YXcov-\eYXcov)\rXcov^{-1/2}} 
	\end{align*}
\end{tlemma}
\begin{tproofof*}{r:bound_proj_crosscov_onehalf}{}
	It holds
	\begin{align*}
		\YXcov - \py \eYXcov
		&= \YXcov 
		- \py \YXcov 
		+ \py \YXcov 
		- \py \eYXcov \\
		&= \py^\perp \YXcov + \py(\YXcov-\eYXcov) 
	\end{align*}
	Thus 
	\begin{align*}
		\n{ (\YXcov - \py \eYXcov )\rXcov^{-1/2}}
		&= \n{(\py^\perp \YXcov + \py(\YXcov-\eYXcov) )\rXcov^{-1/2} }\\
		&\leq  \n{\py^\perp \YXcov \rXcov^{-1/2}}
			+ \n{\py(\YXcov-\eYXcov)\rXcov^{-1/2}} \\
		&\leq  \n{\py^\perp \rXcov^{1/2}}\n{\rXcov^{-1/2} \YXcov \rXcov^{-1/2}} 
			+ \n{(\YXcov-\eYXcov)\rXcov^{-1/2}} 
	\end{align*}

	Eventually it holds $\n{\rXcov^{-1/2} \YXcov \rXcov^{-1/2}}\leq 1$. 
	Indeed, as $\td$ is invariant, it holds that
	\begin{align*}
		\n{\kop}
		%&= \sup_{f\in \ltsp: \nlt{f}\leq 1} \int_{x}|(\kop f)(x)|^2 \dif\td(x)\\
		&= \sup_{f\in \ltsp: \nlt{f}\leq 1} \int_{x}\absv*{\int f(y)p(x,\dif y)}^2 \dif\td(x)
		 \leq  1.
	\end{align*}
	and denoting $\mftX=\Xcov^{1/2} U$ the polar decomposition of $\mftX$ for some partial isometry $U:\lt \rightarrow  \rkhs$, and using $\amftYcX=\kop\amftX$, we get
	\begin{align*}
		\n{\rXcov^{-1/2} \YXcov \rXcov^{-1/2}}
		&= \n{\rXcov^{-1/2} \mftYcX \amftX \rXcov^{-1/2}} \\
		&\leq  \n{\rXcov^{-1/2} \Xcov^{1/2}}\n{ U \kop^* U^*}\n{\Xcov^{1/2} \rXcov^{-1/2}} \\
		&\leq  1.
	\end{align*}
\end{tproofof*}

\subsection{Concentration for mixing processes}
\label{s:concentration_mixing}

\begin{tlemma}{\citet[Lemma 1]{kostic2022learning}}{mixing}
	Let $X$ be strictly stationary with values in a normed space $(\dspace, \n{\cdot})$ and assume $n=2pk$ with $p,k\in \bN$. 
	% m> p, \tau > k
	Let $Z_1,\dots,Z_p$ be $p$ independent copies of $Z_1=\sum_{i=1}^k X_i$.
	Then for $s>0$:
	\begin{align*}
		P\brk[\Big]{\n[\Big]{\sum_{i=1}^n X_i} > s}
		&\leq  
		2 P\brk[\Big]{\n[\Big]{\sum_{j=1}^p Z_j} > s/2}
		+ 2(p-1) \beta_X(k).
	\end{align*}
\end{tlemma}

\end{document}